%% file: main.tex
\documentclass[journal]{IEEEtran} 

\input{preamble}

\newcommand{\rev}[1]{{\color{black}#1}}
\newcommand{\revtwo}[1]{{\color{black}#1}}
\definecolor{somegray}{rgb}{0.5, 0.5, 0.5}
\newcommand{\darkgrayed}[1]{\textcolor{somegray}{#1}}

\newcommand{\insertwatermarkabove}{%
  \vspace*{-3em}
  \begin{center}

    \large\darkgrayed{This paper has been accepted for publication at the \\
    IEEE Transactions on Robotics, 2026.
    \copyright~IEEE}
  \end{center}
  \vspace{1em}
}

\makeatletter
\pretocmd{\@maketitle}{\insertwatermarkabove}{}{}
\makeatother

\ifCLASSOPTIONcompsoc
\usepackage[nocompress]{cite}
\else
\usepackage{cite}
\fi

\newcommand{\imgwidth}{0.19\textwidth}
\newcommand{\spacewidth}{-1.2mm}

\begin{document}

\hyphenation{op-tical net-works semi-conduc-tor}
\def\MYTITLE{Event-Based De-Snowing for Autonomous Driving}
\title{\MYTITLE}
\makeatletter
\g@addto@macro\@maketitle{
  \captionsetup{type=figure}\setcounter{figure}{0}
      \begin{subfigure}[b]{\imgwidth}
        \begin{tikzpicture}
          \node[anchor=south west,inner sep=0] (image) at (0,0) {\includegraphics[width=\linewidth]{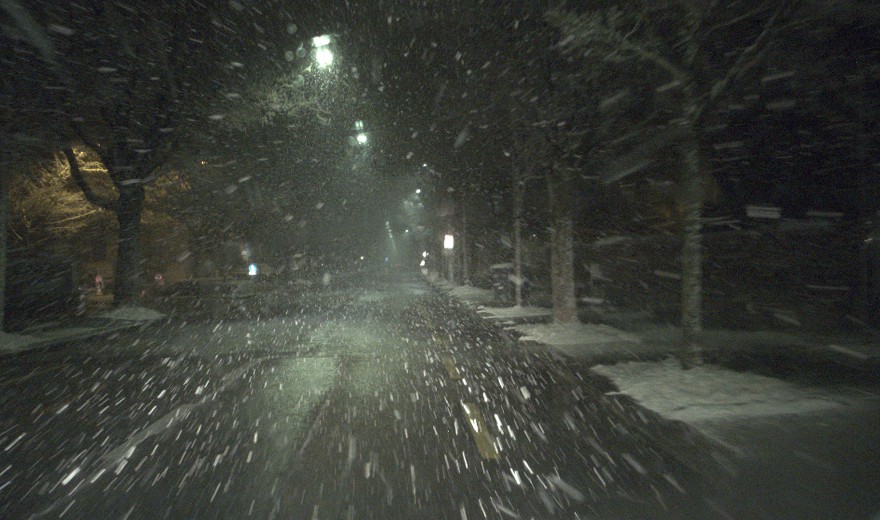}};
          \begin{scope}[on background layer]
            \draw[red, thick, rounded corners] (1.5,0.15) rectangle (2.1,0.75); %
          \end{scope}
         \end{tikzpicture}   %
      \end{subfigure}
      \hspace{\spacewidth}
      \begin{subfigure}[b]{\imgwidth}
        \frame{\includegraphics[trim={5cm, 3cm, 5.8cm, 2cm},clip,width=\linewidth]{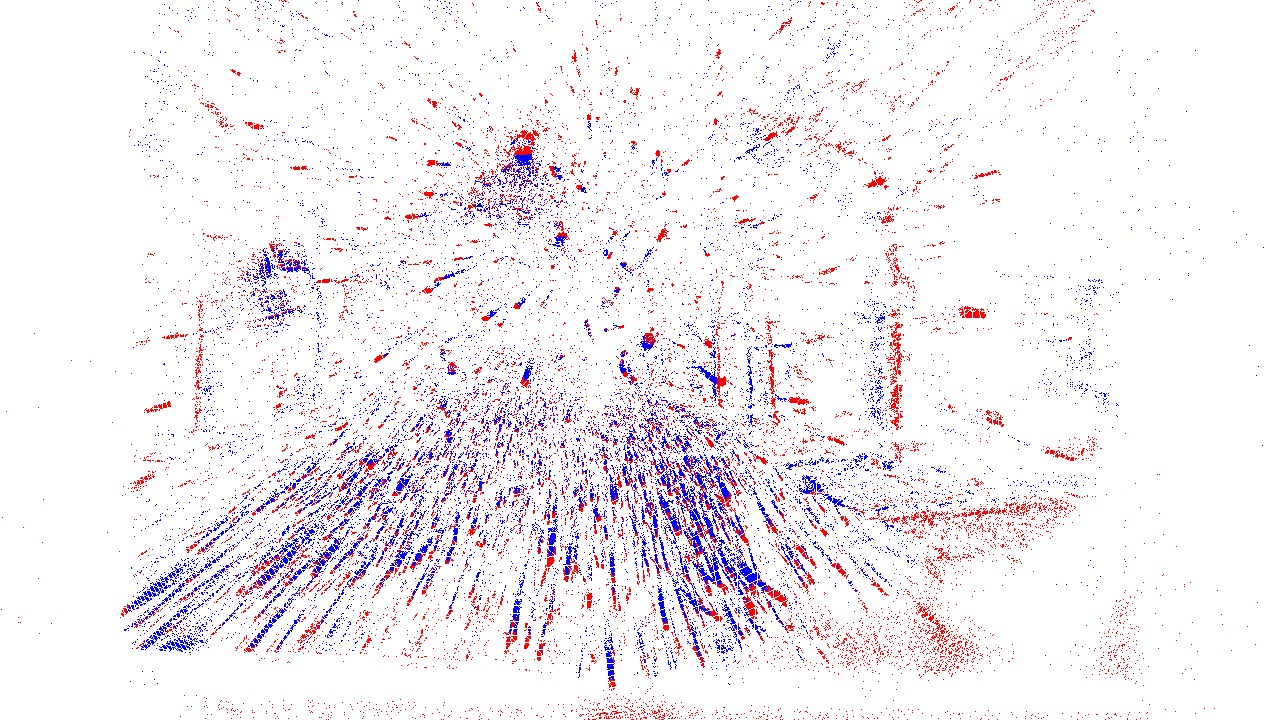}}
      \end{subfigure}
      \hspace{\spacewidth}
      \begin{subfigure}[b]{\imgwidth}
        \includegraphics[trim={5cm, 3cm, 5.8cm, 2cm},clip,width=\linewidth]{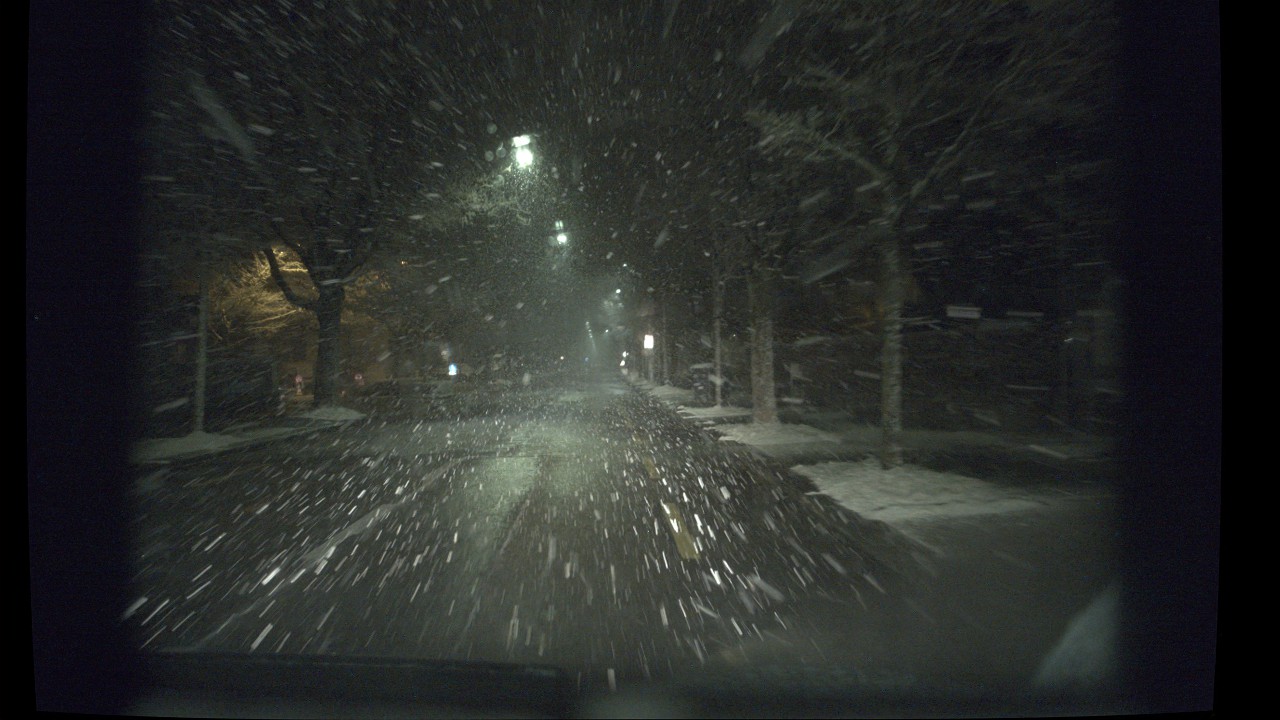}
      \end{subfigure}
      \hspace{\spacewidth}
      \begin{subfigure}[b]{\imgwidth}
        \includegraphics[width=\linewidth]{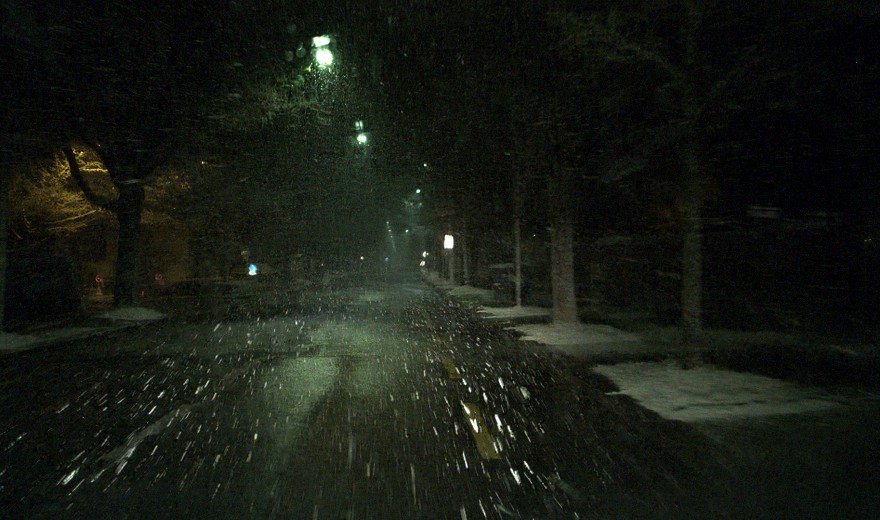}
      \end{subfigure}
      \hspace{\spacewidth}
      \begin{subfigure}[b]{\imgwidth}
        \includegraphics[trim={0cm, 3.5cm, 0cm, 3.5cm},clip,width=\linewidth]{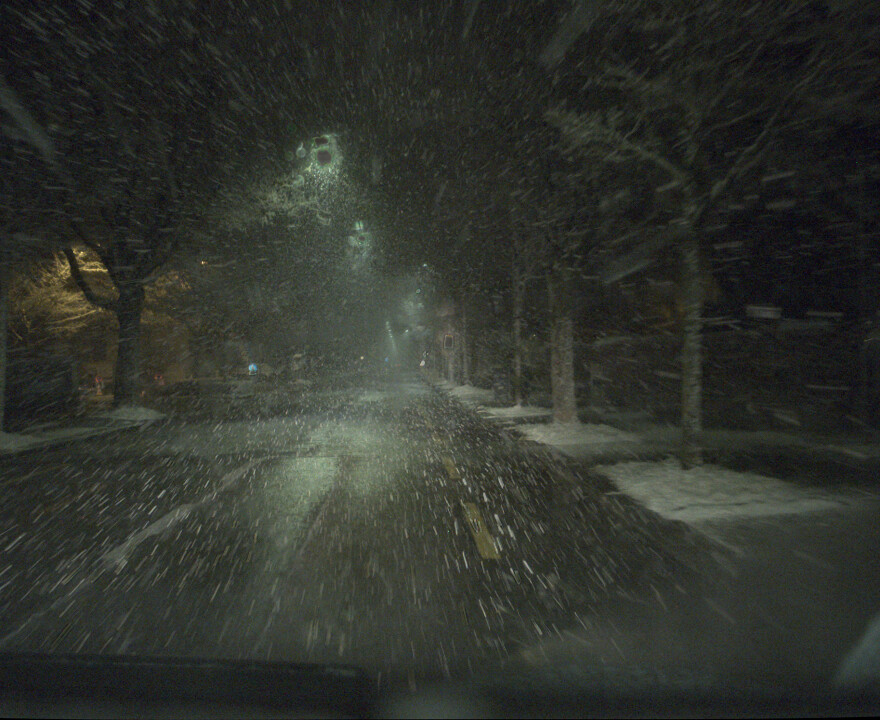}
      \end{subfigure}

      \vspace{0.2em}
      \begin{subfigure}[b]{\imgwidth}
        \includegraphics[trim={12cm, 1.5cm, 10.8cm, 12cm},clip,width=\linewidth]{floats/images/fig1_new/img_000040.jpg}
        \caption{Snowfall image}
      \end{subfigure}
      \hspace{\spacewidth}
      \begin{subfigure}[b]{\imgwidth}
        \frame{\includegraphics[trim={20cm, 5cm, 15.8cm, 15cm},clip,width=\linewidth]{floats/images/fig1_new/events000040.jpg}}
        \caption{Events}
      \end{subfigure}
      \hspace{\spacewidth}
      \begin{subfigure}[b]{\imgwidth}
        \includegraphics[trim={20cm, 5cm, 15.8cm, 15cm},clip,width=\linewidth]{floats/images/fig1_new/restormer_000040.jpg}
        \caption{Restormer}
      \end{subfigure}
      \hspace{\spacewidth}
      \begin{subfigure}[b]{\imgwidth}
        \includegraphics[trim={12cm, 1.5cm, 10.8cm, 12cm},clip,width=\linewidth]{floats/images/fig1_new/snowformer_original_000040.jpg}
        \caption{Snowformer}
      \end{subfigure}
      \hspace{\spacewidth}
      \begin{subfigure}[b]{\imgwidth}
        \includegraphics[trim={12cm, 5cm, 10.8cm, 15.5cm},clip,width=\linewidth]{floats/images/fig1_new/ours_000040.jpg}
        \caption{\textbf{Ours}}
      \end{subfigure}  

	\caption{
        \textbf{Event-based video snow removal in challenging nighttime scenes}:
    (a) Sample image from our dataset captured while driving in snowfall, and (b) corresponding event data highlighting the motion of snowflakes. 
    The presence of dense, dynamic snow and low visibility presents significant challenges for conventional image restoration methods, as illustrated by the results of (c) Restormer\cite{Zamir21cvpr} and (d) Snowformer\cite{Chen22arxiv}. 
    Our proposed event-based approach (e) utilizes event information to address these challenges and restore clearer scene content under adverse weather conditions.
  }
 \label{}
 \vspace{-1em}
}
\makeatother

\authif{
\author{Manasi Muglikar
\qquad
Nico Messikommer
\qquad
Marco Cannici
\qquad
Davide Scaramuzza\\
\,
\thanks{ 
    This work was supported by the European Research Council (ERC) under grant agreement No. 864042 (AGILEFLIGHT).
The authors are with the Robotics and Perception Group, Department of Informatics, University of Zurich, Switzerland.
}
}
}

\IEEEtitleabstractindextext{
}

\bstctlcite{BSTcontrol}
\maketitle

\input{sections/00_abstract}
\authif{\textbf{Code, dataset and video are available under:} \url{https://rpg.ifi.uzh.ch/evsnow.html}}
\anonif{}
\input{sections/01_introduction}
\input{sections/02_rel_work}
\input{sections/04a_method_model.tex}
\input{sections/04b_method_learning.tex}

\input{sections/04c_method_implementation.tex}
\input{sections/03_dataset}
\input{sections/05b_results.tex}
\input{sections/06_conclusion}

\ifCLASSOPTIONcaptionsoff
  \newpage
\fi

\clearpage
\section*{Supplementary Material}
\input{sections/supp/supplementary.tex}

\bibliographystyle{IEEEtran}
\bibliography{all}
\end{document}

%% file: preamble.tex
\usepackage[dvipsnames]{xcolor}

\usepackage{url}
\usepackage{verbatim}

\usepackage[nocompress]{cite}
\usepackage{subcaption}
\usepackage{graphicx}
\usepackage{array}
\usepackage{graphicx}
\usepackage{caption}
\newcolumntype{P}[1]{>{\centering\arraybackslash}p{#1}}
\newcolumntype{M}[1]{>{\centering\arraybackslash}m{#1}}

\usepackage{booktabs}
\usepackage{adjustbox}

\newcommand{\Sec}{Section~}
\newcommand{\Fig}{Fig.~}
\newcommand{\Tab}{Table~}
\newcommand{\Alg}{Algo.~}
\newcommand{\Eq}{Equation.~}

\usepackage{comment}
\usepackage{enumitem} %
\usepackage{hyperref}
\usepackage{booktabs}
\usepackage{multirow}
\usepackage{nicefrac}
\usepackage[normalem]{ulem}
\useunder{\uline}{\ul}{}
\usepackage{array}
\usepackage{tikz}
\usetikzlibrary{backgrounds}
\usepackage{pifont}%
\usepackage{makecell}
\usepackage{capt-of}
\usepackage{balance} %

\usepackage{tabularx}
\usepackage{siunitx}
\usepackage{graphicx}
\usepackage{pgfplots}

\usepackage{algorithm}
\usepackage{algpseudocode}
\algrenewcommand\algorithmicrequire{\textit{Input:}}
\algrenewcommand\algorithmicensure{\textit{Output:}}

\usepackage{colortbl} %
\usepackage{amssymb,amsmath,amsfonts}
\newcommand{\DAVISSnow}{Slider-Snow }
\newcommand{\DSECSnow}{DSEC-Snow }

\usepackage{color} 
\pgfplotsset{compat=1.18}

\newcommand{\ANONYMIZE}{0}  %

\newcommand{\authif}[1]{\ifnum\ANONYMIZE=0 #1\fi}
\newcommand{\anonif}[1]{\ifnum\ANONYMIZE=1 #1\fi}

%% file: sections/00_abstract.tex
\begin{abstract}
Adverse weather conditions, particularly heavy snowfall, pose significant challenges to both human drivers and autonomous vehicles. 
Traditional image-based desnowing methods often introduce hallucination artifacts as they rely solely on spatial information, while video-based approaches require high frame rates and suffer from alignment artifacts at lower frame rates.
Camera parameters, such as exposure time, also influence the appearance of snowflakes, making the problem difficult to solve and heavily dependent on network generalization. 
In this paper, we propose to address the challenge of desnowing by using event cameras, which offer compressed visual information with submillisecond latency, making them ideal for desnowing images, even in the presence of ego-motion. 
Our method leverages the fact that snowflake occlusions appear with a very distinctive streak signature in the spatiotemporal representation of event data. 
We design an attention-based module that focuses on events along these streaks to determine when a background point was occluded and use this information to recover its original intensity. 
We benchmark our method on DSEC-Snow, a new dataset created using a green-screen technique that overlays pre-recorded snowfall data onto the existing DSEC driving dataset, resulting in precise ground truth and synchronized image and event streams. 
Our approach outperforms state-of-the-art desnowing methods by $3$ dB in PSNR for image reconstruction. 
Moreover, we show that off-the-shelf computer vision algorithms can be applied to our reconstructions for tasks such as depth estimation and optical flow, achieving a $20\%$ performance improvement over other desnowing methods. 
Our work represents a crucial step towards enhancing the reliability and safety of vision systems in challenging winter conditions, paving the way for more robust, all-weather-capable applications.
\end{abstract}

%% file: sections/01_introduction.tex
\section{Introduction}
\label{sec:introduction}

Imagine driving through heavy snowfall. 
Bright, swirling flakes, windshield accumulation, and reduced visibility severely degrade scene perception, making driving not only unpleasant but also potentially dangerous.
Autonomous vehicles and assistive driving systems, designed to enhance road safety, also struggle in these conditions as snowflakes obscure both camera and LiDAR sensors.
To advance vehicle autonomy and automotive safety, addressing adverse weather challenges is crucial. 
Thus, effective desnowing techniques are necessary to ensure the reliability and safety of vision systems in snowy environments.

Existing solutions, such as training a network on specific desnowing datasets \cite{Bijelic20CVPR, Liu18TIP, Zhang21TIP, Chen20ECCV, Chen21ICCV, Chen23ICCV} or using Gated Cameras \cite{Bijelic20CVPR}, fall short—either failing to generalize across snow conditions or losing intensity information in low light. 
In comparison to a single image, a video provides richer temporal context about the dynamic features of the scene.
Building on this, existing works \cite{Garg05ICCV,Yang20ieee,Ren17cvpr,Li21tip, Li19arxiv,Kim15TIP,Bijelic20CVPR,Chen23ICCV} have studied the effect of incorporating this temporal information for image desnowing, showing promising results.
However, the performance of video desnowing relies on the framerate of the camera.
A high framerate ensures reliable alignment across individual frames, which can be used to align the background and remove snow occlusion.
For effective image desnowing, it is crucial to capture high-speed scene information.
We, therefore, propose to solve this problem by using event cameras, which provide extremely high temporal resolution (on the order of $1\,\mathrm{MHz}$), without a huge bandwidth demand.

Event cameras measure intensity changes with very low latency (up to $\SI{1}{\micro \second}$) and asynchronously \cite{Gallego20pami}.
This produces a stream of events that encode the time, location, and polarity of the brightness change.
The main advantages of event cameras include sub-millisecond latency, very high dynamic range ($>\SI{120}{dB}$), and strong robustness to motion blur.
This work leverages these unique properties to remove snow occlusions from images effectively.

Since event cameras have high temporal resolution and no exposure time, snowflakes always appear as streaks in the space-time representation of events (c.f. Fig.~\ref{fig:snow_effect}).
This is unlike conventional cameras, where the exposure time plays a major role in the appearance of the snowflake.
Thus, our method takes advantage of this unique signature in the space-time domain to track these streaks.
In the absence of ego-motion, the task of recovering background intensity is quite simple, as one only has to look at the intensity changes (or events) per pixel.
However, with ego-motion, the point corresponding to the pixel that is occluded at a certain time may not be the same at a future time.
Therefore, keeping track only along the pixel will result in significant errors.
Instead, we propose to look along the streak, as a point in 3D space will be occluded and de-occluded along this streak.
To recover the background intensity where the occlusion covers the background, it is necessary to look along the streak in time.
By focusing only on points along these streaks, we can accurately determine when a point in the scene was occluded. 
Using this information, we reconstruct the background intensity, enabling us to recover a clear image even under challenging weather conditions.
Our approach improves the performance of image-based and video-based approaches by over $3\,\mathrm{dB}$ in terms of PSNR of image reconstruction.
Not only that, we also outperform image-based approaches in further downstream tasks such as depth estimation, optical flow, and object detection by over 20\% in terms of accuracy.

We list our contributions as follows:
\begin{itemize}
    \item \textbf{A new event-based desnowing dataset generated by chroma composition method}: We develop a synthetic dataset (\DSECSnow) by overlaying recorded snowfall onto the DSEC \cite{MGehrig21ral} driving dataset using green-screen technology, providing precise ground truth and synchronized image-event streams without the need for complex snow rendering.
    \item \textbf{A novel approach for desnowing images using event cameras}: Our approach utilizes the rich temporal information of events to detect and remove snowflake occlusions by tracking their spatiotemporal streaks, enabling accurate background recovery. We also propose a simple learning framework to fuse the event and image information for improved desnowing performance.
    \item \textbf{A real-world driving dataset collected in snowy conditions}: We present a real-world event-camera dataset captured while driving in snowfall, offering valuable data for testing the robustness of event-based desnowing methods in practical scenarios.
\end{itemize}

%% file: sections/02_rel_work.tex
\section{Related work}
\input{floats/tables/tab_datasets}
\label{sec:rel_work}
We summarize the related works for image and video desnowing in \Sec \ref{sec:rel_work:image}.
There has been some progress in the context of image deraining with event cameras, which is summarized in \Sec \ref{sec:rel_work:event}.
Lastly, we also summarize the existing datasets published so far for image desnowing in \Sec \ref{sec:rel_work:data}.
\subsection{Image and video desnowing}
\label{sec:rel_work:image}
Estimating the background image in the presence of rain or snow is challenging as it requires hallucination of the background content.
Some earlier works focused on modeling the rain streaks and snowflakes and decomposed the image into background and foreground components \cite{Garg04CVPR, Kang12tip}.
However, even if the perfect model of the rain streaks or snowflakes is known, the background content is still challenging to estimate \cite{Yang21tpami}.
To address this issue, recent works \cite{Liu18TIP, Chen20ECCV} have proposed using deep learning methods to estimate the background content.
\cite{Zhang21TIP} proposed learning a representation for snow using the geometric and semantic properties of snow.
On the other hand, Snowformer \cite{Chen22arxiv} proposes using a vision transformer that fully combines local and global information and obtains state-of-the-art results.
Other image-based approaches focus more on image restoration \cite{Zamir21cvpr} or image deraining\cite{Zhang23ICCV}, both of which do not generalize well for image desnowing \cite{Chen23ICCV}.
Recent work~\cite{Zhang23ICCV} tackles the unique challenges of nighttime deraining—caused by non-uniform local illumination and complex rain-light interactions—by introducing a Rain Location Prior (RLP) learned via a recurrent residual model, along with a Rain Prior Injection Module (RPIM) to enhance feature representation and boost deraining performance at night.
Instead of handling each image restoration task separately, Restormer \cite{Zamir21cvpr} proposes a unified framework for image restoration tasks, including image deraining, deblurring, and denoising.
However, the performance of a generalized image restoration method is often limited for specialized tasks such as image desnowing.
Therefore, Snowformer \cite{Chen22arxiv} proposes a dedicated desnowing vision transformer with a scale-aware snow query and local-patch embedding, resulting in state-of-the-art results for image desnowing.
These methods, however, rely on the spatial information available in the image to estimate the background content.
In complex weather conditions, such as heavy snowfall, the spatial information is often not sufficient to accurately estimate the background content.
\rev{Recently, Sun et al. \cite{Sun24ECCV} proposed a histogram transformer network that leverages histogram self-attention and a dynamic-range convolution to efficiently captures long-range dependencies across similar-intensity regions, achieving superior restoration performance compared to existing methods.}

Increasing temporal resolution for this task gives rise to video-based desnowing methods.
Video-based desnowing methods \cite{Garg05ICCV} have explored the use of temporal information to improve the quality of desnowed images.
The seminal work of Garg et al. \cite{Garg05ICCV} proposed a model-based approach to remove occlusions such as rain by characterizing the photometric and temporal properties of rain streaks.
Subsequent works \cite{liu2009pixel, zhang2006rain,barnum2010analysis, bossu2011rain, santhaseelan2015utilizing} have extended this work by proposing various priors to model the rain streaks and snowflakes.
Another line of approach used matrix factorization to encode the correlation of background video along the temporal dimension \cite{chen2013generalized, Jiang2017, Kim15TIP, ren2017video}.

More recently, deep learning methods have also shown significant improvements in video desnowing \cite{li2018video, chen2018robust, liu2018erase, liu2018d3r, yang2019frame,Jin03vc, Yue21cvpr, Chen23ICCV}.
Li et al. \cite{li2018video} proposed a multi-scale convolutional neural network (CNN) for sparse coding to encode and remove the repetitive local patterns of rain streaks at different scales.
Liu et al. \cite{liu2018erase} proposed a recurrent neural network (RNN) to turn this problem into classification of rain pixels and then recover the background.
While these methods have shown promising results on synthetic data, they often struggle with complex weather conditions and fail to generalize to real-world scenarios.
To address the domain gap between synthetic and real rain data, recent work~\cite{Yue21cvpr} proposes a semi-supervised video deraining method that leverages a deep-learning-based dynamical rain generator and Monte Carlo EM optimization, jointly exploiting both labeled synthetic and unlabeled real videos for improved performance in real-world scenarios.
Recently, Chen et al. \cite{Chen23ICCV} addressed the challenging task of video snow removal by introducing a high-quality dataset that simulates realistic snow and haze through advanced rendering and augmentation techniques.

\subsection{Event-based image deraining and desnowing}
\label{sec:rel_work:event}
Event cameras, known for their high temporal resolution and robustness to motion blur, have been increasingly utilized to enhance traditional imaging systems \cite{Tulyakov22cvpr}. 
This high temporal resolution of events was used for de-occluding frames \cite{Zou23Arxiv}.
This technique utilizes the asynchronous nature of event cameras to provide additional temporal information, enabling more effective de-occlusion of images in real-time scenarios.
Similarly, \cite{Wang23ICCV} introduced an unsupervised video deraining method that combines event data with traditional video frames. 
By integrating these two data types, the method effectively removes rain streaks and other occlusions from video sequences, demonstrating significant improvements in visibility and clarity. 

\rev{Compared to image deraining, there has been limited work on event-based image desnowing.
To the best of our knowledge, the only existing work is \cite{Wolf25TPAMI}, which proposes a model-based approach for identifying and removing snow occlusions from events.
The method relies on statistical modeling of snowflake events to partition event streams into snowflake and background events.
While this method shows promising results, it is limited by the assumptions made in the model (e.g. thin structures are often misclassified as snowflakes instead of background) and does not leverage the complementary information available from intensity images, thereby limiting its performance in complex scenarios.
Additionally, for complex tasks such as image reconstruction, the lack of intensity information can lead to loss of fine details and color in the reconstructed images.
In this paper, we propose a learning-based approach that leverages both event and intensity information to effectively remove snow occlusions and reconstruct high-quality background images in color.
}
\subsection{Snow Datasets}
\label{sec:rel_work:data}
While there has been significant progress in the field of image and video desnowing, the availability of datasets for training and evaluating these methods remains limited.
Some existing datasets focus on image desnowing, such as Snow100K \cite{Liu18TIP}, which provides a large collection of synthetic images with snow occlusions generated using Photoshop rendering techniques \cite{photoshop}, providing ground truth images.
In SnowKITTI2012 and SnowCityScapes \cite{Zhang21TIP}, the authors used a similar approach to generate different densities of snow occlusions on images.
SRRS \cite{Chen20ECCV, Chen21ICCV} improved these models and included more realistic rendering of snowy scenes by introducing a veiling effect and then using Photoshop to render snowflakes.
Since these datasets rely on Photoshop, the realism of the snow occlusions is lacking, and therefore causes limited generalization to real-world desnowing of images.
Therefore, \cite{Chen23ICCV} proposed a new video desnowing dataset which is rendered using Unreal Engine, providing a more realistic simulation of snow occlusions.
These datasets, however, do not provide events.
\rev{Recently, \cite{Wolf25TPAMI} proposed a dataset for event-based image desnowing by recording snowfall using an event camera and manually annotating objects in the scene.
This dataset, however, does not provide images and ground truth background images, limiting its use for training learning-based desnowing methods.}
A comparison of these datasets is shown in \Tab \ref{tab:dataset}.

To the best of our knowledge, there currently exists no dataset that provides events and images for the task of desnowing.
We therefore propose two datasets for this task, namely the \DSECSnow dataset and \DAVISSnow Dataset.
The \DSECSnow dataset overlays foreground data of snowfall recorded with an event camera onto a background consisting of the driving dataset DSEC \cite{MGehrig21ral}.
This process removes the complexity of rendering snow particles that are both photo-realistic and physically accurate.
Another advantage this dataset provides is the availability of synchronized events, images, and ground truth images.
In addition to this, we also record real data of driving in snowfall using a color DAVIS event camera, called \DAVISSnow.

%% file: floats/tables/tab_datasets.tex
\begin{table}[ht]
    \centering
    \begin{adjustbox}{max width=\linewidth}
    \begin{tabular}{c|c|c|c}
        Dataset & Simulation & Sensors & GT  \\ \hline
        Snow100K\cite{Liu18TIP} & Yes & Image& Yes\\
        SnowCityScapes\cite{Zhang21TIP} & Yes& Image& Yes\\
        SnowKITTI2012\cite{Zhang21TIP} & Yes& Image& Yes\\
        SRRS\cite{Chen20ECCV} & Yes& Image& Yes\\
        CSD\cite{Chen21ICCV} & Yes& Image & Yes\\
        SnowVideo \cite{Chen23ICCV} & Yes& Video & Yes\\ \hline
        \rev{ESnowBR\cite{Wolf25TPAMI}} & \rev{No} & \rev{Events} & \rev{No}\\ \hline
        \textbf{DSEC-Snow (ours)} & Yes & Video + Events & Yes\\
        \textbf{SnowDriving (ours)} & No & Video + Events & No \\ \hline
    \end{tabular}        
    \end{adjustbox}

    \caption{\textbf{Comparison of snow datasets.}
    Existing snow datasets primarily focus on synthetic image-based data with ground truth (GT), while our DSEC-Snow and SnowDriving datasets provide real and simulated video sequences with event data, addressing the need for benchmarks suitable for event-based snow removal under more realistic conditions}
    \label{tab:dataset}
\end{table}

%% file: sections/04a_method_model.tex
\begin{figure}[t]
    \centering
    \includegraphics[width=\linewidth]{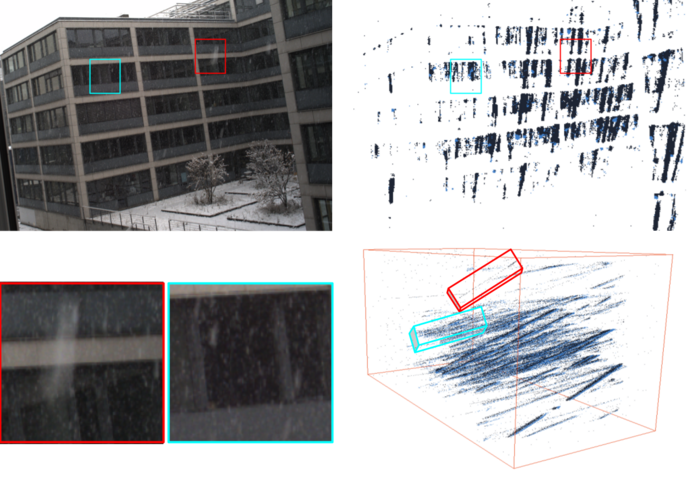}
    \caption{\textbf{Effect of snow on images (left) and events (right)}
    (Left) The appearance of snowfall in an image depends on multiple factors, such as snowflake size, density, ambient illumination, and camera exposure settings.
    Two such examples are shown within the red and blue regions.
    (Right) The same snow occlusions in event data are visualized on the image plane and x-y-t volume (bottom-right).
    Irrespective of the snowflake size, snow occlusions have a unique spatio-temporal pattern in the form of streaks (bottom-right).
    }
    \label{fig:snow_effect}
\end{figure}

\input{floats/alg_model_based}
\section{Understanding the effect of snow on images and events}
\label{sec:method:model}
The appearance of snow in images and events is quite different due to the nature of the sensors.
Garg et al. \cite{Garg05ICCV} analyzed snowflakes and showed that their appearance depends not only on the size, shape, and distance from the camera, but also on the ambient illumination and camera exposure settings.
This results in a wide variety of snowflake appearances in images, as shown in \Fig \ref{fig:snow_effect} (top).
For example, \cite{Garg05ICCV} provided a mathematical model of how snow can appear either as bright spots, streaks, or even haze depending on the distance of the snowflake from the camera, for the same exposure settings.
This makes the challenge of identifying snowflakes in images quite significant.

On the other hand, snow occlusions in event data have a unique spatio-temporal pattern in the form of streaks, as shown in \Fig \ref{fig:snow_effect} (bottom).
This is because of the high temporal resolution of the event camera, which captures the snow occlusions as streaks in the x-y-t volume.
This makes the task of identifying snow occlusions in events much easier compared to images.
The streaks in the x-y-t volume are independent of the distance and only depend on the relative difference between background intensity and snowflake intensity.
For example, in \Fig \ref{fig:snow_effect} (bottom, left), the snow streak corresponding to the same snowflake is visible in front of a darker background like the window, but not visible in front of a brighter background like the sky.
This makes the task of identifying snow occlusions in events much easier compared to images.

\subsection{De-occluding with events}
\label{sec:snow_inpainting}
We now describe a geometric way to solve for background extraction using events.
\rev{This algorithm is also summarized in \Alg \ref{alg:background_estimation_compact}}.
As a snowflake moves in front of the background, it causes an intensity change, resulting in events being triggered.
The intensity at a pixel $\textbf{X}$ at time $t$ caused by a snowflake of intensity $I_r$ occluding a background intensity of $I_b$ can be written as:
\begin{equation}
    \Delta I (t) = I_r  - I_b = pC
\end{equation}
In general, since the snowflake is usually brighter than the background, the intensity change is positive when the snowflake is occluding the background, triggering positive polarity events.
When the snowflake moves away from the background, the intensity change is negative, triggering negative polarity events.
Therefore, assuming we know the intensity of the snowflake, estimating the background intensity is simply a matter of integrating the events over time $\tau$, scaling it by the contrast threshold $C$, and subtracting it from the snowflake intensity.
Thus, the background intensity can be estimated as:
\begin{equation}
    I_b = I_r - \sum_{0}^{\tau}pC
    \label{eq:basic}
\end{equation}
Therefore, once we have identified the snow occlusions (using events), we can recover the background intensity from the equation above and attend to events that occur at the same pixel at different times.
In the presence of background motion, \Eq \ref{eq:basic} no longer holds true, as the pixel before and after the occlusion can belong to a different point.
This implies that the pixel at location $X_0$ being occluded at time $t_0$ could be at location $X_i$ when it is de-occluded.
Therefore, the search for the corresponding de-occluded pixel is no longer only along the time dimension but also along the spatial dimension.
What does remain true, however, is that the equation still holds in 3D space, and the above equation is modified as follows:
\begin{equation}
    I_b = I_r - \sum_{0}^{\tau}W(pC)
    \label{eq:motion}
\end{equation}
\rev{where $W$ represents the warping of events along the motion streak.}
However, estimating this warp is quite challenging, especially when the snow occlusions are dense and overlapping.
We follow the approach of \cite{Garg05ICCV, garg07IJCV} to identify the streaks using the rain-velocity prior and use this to recover the background intensity.
The main difference with respect to \cite{Garg05ICCV, garg07IJCV} is that we use events instead of a set of images as input to this method.
\rev{We take events in a short time window (e.g., $5$ ms) to ensure that the motion of the background is minimal, and therefore the warping can be approximated using a constant velocity model. Thus, the dominant motion is because of the snowflakes and can be estimated by approach proposed in \cite{Garg05ICCV, garg07IJCV}. An event pixel (which we assume is a snow event), will move by $v\times t$ where $v$ is velocity and $t$ is time. In the spatial neighborhood of this event, we search for other events that fit this motion model. This can be done by iterating over a set of velocity hypotheses and for each hypothesis, and selecting the hypothesis which results in the maximum number of events being aligned along the motion streak. However, it can be computationally expensive to search over all possible velocities. Therefore, we fix the speed and only change the direction. Instead of a dense velocity grid, pick a small number of directions ( $8$ uniformly spaced around a circle). For each direction, we project local events along that direction and measure alignment by calculating event density along the line.}

This forms our model-based approach to de-occlude the snow occlusions in images using events.
As this approach requires several assumptions about the snow occlusions, we also propose a data-driven approach to learn to de-occlude the snow occlusions using events, which we describe in the following sections.

%% file: floats/alg_model_based.tex
\begin{algorithm}[ht]
\caption{Background Intensity Estimation under Snow Occlusion Using Event Data}
\label{alg:background_estimation_compact}
\begin{algorithmic}[1]
\Require Polarity event stream $p(t)$ at pixel $X$; reference intensity $I_r$; contrast threshold $C$; time window $\tau$; [optional] warping function $W(\cdot)$
\Ensure Estimated background intensity $I_b$ at pixel $X$

\State $I_b \gets I_r$
\If{background is static}
    \State Accumulate events: $E \gets \int_0^\tau p(t) \, dt$ 
    \State Update background intensity: $I_b \gets I_r - C \times E$
\Else
    \State Identify warping $W(\cdot)$ along motion streaks (e.g., using velocity prior)
    \State Accumulate warped events: $E_w \gets \int_0^\tau W(p(t) \times C) \, dt$
    \State Update background intensity: $I_b \gets I_r - E_w$
\EndIf
\State \Return $I_b$

\end{algorithmic}
\end{algorithm}

%% file: sections/04b_method_learning.tex
\section{Data-driven Desnowing with Event Cameras}
\label{sec:method}
\begin{figure*}[!t]
    \centering
    \subfloat[Overview]{\label{fig:method_overview:full}\includegraphics[trim={3cm, 2cm, 0cm, 5cm},clip,width=0.5\linewidth]{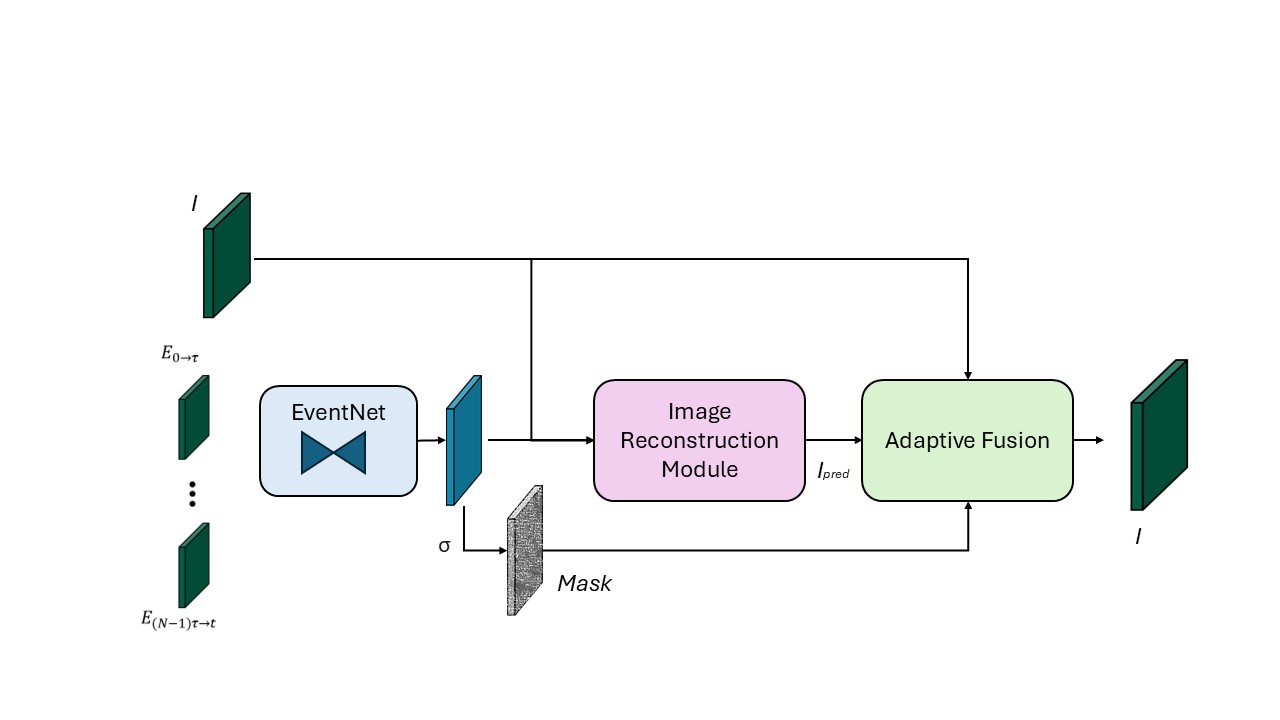}}
    \subfloat[EventNet]{\label{fig:method_overview:eventnet}\includegraphics[trim={0cm, 13cm, 16cm, 0cm},clip,width=0.5\linewidth]{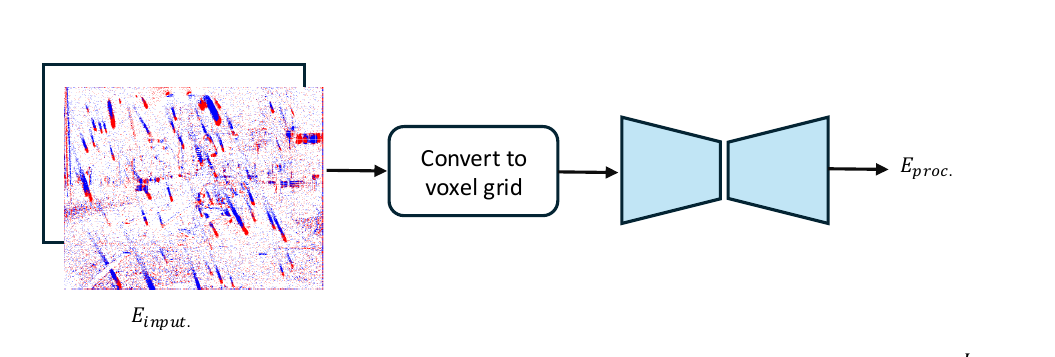}}\\
    \subfloat[Image Reconstruction]{\label{fig:method_overview:imgrec}\includegraphics[trim={0cm, 7cm, 20cm, 7cm},clip,width=0.5\linewidth]{floats/images/method_pdf.pdf}}
    \subfloat[Adaptive Fusion]{\label{fig:method_overview:adapt}\includegraphics[trim={15cm, 10cm, 8cm, 5cm},clip,width=0.5\linewidth]{floats/images/method_pdf.pdf}}\\
    
    \caption{
        \rev{\textbf{Overview of our data-driven method for reconstructing deoccluded images using event-camera data.}
        Events within the camera's exposure time are segmented into non-overlapping spatio-temporal windows, converted into voxel-grid representations, and processed alongside RGB images through modality-specific feature extraction. 
        Event features are extracted via EventNet, which also produces a spatial mask. 
        The reconstruction module fuses event and image features to reconstruct the image, and adaptive fusion uses a learned mask to blend this reconstruction with the original image.}
    }
    \label{fig:method_overview}
    \vspace{-1em}
\end{figure*}

In this section, we present our approach for desnowing images using event cameras.
\rev{Similar to the model-based approach, we model the observed image \(I_{input}(x)\) as a combination of the clean background image \(I_{clean}(x)\) and the occlusion caused by snowflakes \(I_{occl}(x)\).
However, instead of explicitly modeling the occlusion using geometric priors, we aim to learn a mapping from the input image and event data to the clean image using a neural network.
The design of each component is inspired by classical geometric approaches but replaces explicit modeling with learnable modules. 
An overview of our proposed method is shown in \Fig~\ref{fig:method_overview}.
}

\rev{
\textbf{Problem formulation}:
The event stream $E_{input} = \{e_i\}_{i=1}^N$ consists of asynchronous events $e_i = (x_i, y_i, t_i, p_i)$, where $(x_i, y_i)$ represents the spatial coordinates, $t_i$ is the timestamp, and $p_i \in \{-1, +1\}$ indicates the polarity of the brightness change. 
The events are accumulated over the exposure time of the camera to form a voxel grid representation $V \in \mathbb{R}^{H \times W \times B}$, where $H$ and $W$ are the height and width of the image, and $B$ is the number of temporal bins.
Given an input image $I_{input} \in \mathbb{R}^{H \times W \times 3}$ and the corresponding event voxel grid $V$, we aim to reconstruct the clean image $I_{pred} \in \mathbb{R}^{H \times W \times 3}$
where $I_{pred}$ is the reconstructed clean image.
Our approach consists of three main components: $(1)$\textbf{EventNet}, which processes the event stream to extract spatio-temporal patterns of snowflake streaks; $(2)$ \textbf{Image Reconstruction}, which fuses the image and event features using a transformer; and reconstructs the clean image by combining the network prediction and the input image guided by a learned mask.
} \\
\noindent \textbf{EventNet}:
The EventNet module is designed to extract spatio-temporal features from the raw event data, which encode the motion and geometry of snowflake occlusions. 
The events are first converted to a voxel grid representation.
Instead of relying on explicit warping or velocity priors as in classical geometric models, EventNet employs a convolutional LSTM (ConvLSTM) to capture temporal dependencies, followed by a U-Net architecture for hierarchical spatial feature extraction. 
\rev{This structure enables the network to implicitly learn the geometric properties of snow streaks from data.  
The output of EventNet is a feature map that encodes the spatio-temporal patterns of snowflake occlusions $E_{proc}$, as shown in \Fig \ref{fig:method_overview:eventnet}.
}

\noindent \textbf{Image Reconstruction}:
The image reconstruction module uses a hierarchical Transformer-based architecture proposed in Snowformer\cite{Chen22arxiv}.
This Transformer fuses features from both the intensity image and the processed event stream. 
It comprises a U-Net-style encoder-decoder structure built upon Transformer blocks, which are capable of capturing long-range dependencies and aggregating contextual information across multiple scales, as shown in \Fig \ref{fig:method_overview:imgrec}. 
Channel attention mechanisms and context interaction layers further enhance feature fusion at each scale.
\rev{The image and event features are concatenated and passed through the Transformer backbone to produce a de-snowed reconstruction $I_{rec}$
It replaces hand-crafted priors and explicit motion modeling with multi-head self-attention and hierarchical feature aggregation, allowing the network to learn complex, non-linear dependencies across both spatial and temporal domains. 
The Transformer's ability to capture global context serves as a learnable analog to geometric reasoning over motion and occlusion structure, integrating information from both local neighborhoods and the entire image.

The next stage of image reconstruction combines the Transformer's prediction with the original input image using the spatial mask produced by EventNet, as detailed below.
}
\rev{
\noindent \textbf{Mask Prediction and Adaptive Fusion}:
In the final stage, we perform an adaptive, pixel-wise fusion between the input image and the network’s de-snowed prediction. 
The fusion is controlled by the learned spatial mask generated by applying a sigmoid activation to EventNet's output feature map, which indicates the likelihood of occlusion at each pixel.
Since the mask is learned from data, it serves as a soft, adaptive counterpart to the explicit occlusion identification in geometry-based approaches.
\begin{equation}
    I_{pred} = mask \odot I_{rec} + (1 - mask) \odot I_{input},
\end{equation}
where \(I_{rec}\) is the output of the Transformer backbone, \(I_{input}\) is the original input image, and \(mask\) is the spatial mask produced from EventNet. The operator \(\odot\) denotes element-wise multiplication.
}
This formulation can be interpreted as a data-driven generalization of the subtraction operation in geometry-based approaches, where the mask adaptively determines the contribution of the predicted clean image and the original input. 
Rather than explicitly subtracting a warped occlusion estimate(as done in model-based approach), the network learns the optimal blending strategy for each pixel, guided by the inferred occlusion geometry from the event data.

\subsection{Loss Function}
For training supervision, we adopt the \textbf{L1 loss} as our primary reconstruction loss. The loss function is defined as:
\begin{equation}
L_{\text{L1}} = \| S(I(x)) - Y \|_1,
\end{equation}
where \(S(\cdot)\) denotes the proposed SnowFormer network, \(I(x)\) is the input snowy image, and \(Y\) is the corresponding ground truth image.

To further enhance the perceptual quality of the restored images, we also incorporate a perceptual loss. This loss is computed on feature maps extracted from specified layers of a pretrained VGG-19 network and is formulated as:
\begin{equation}
L_{\text{perceptual}} = \sum_{j=1}^2 \frac{1}{C_j H_j W_j} \left\| \phi_j(S(I(x))) - \phi_j(Y) \right\|_1,
\end{equation}
where \(\phi_j\) represents the activation of the \(j\)-th selected layer in VGG-19, and \(C_j\), \(H_j\), and \(W_j\) correspond to the number of channels, height, and width of the feature map at that layer, respectively.

The overall loss function is expressed as a weighted sum of the reconstruction and perceptual losses:
\begin{equation}
L = \lambda_1 L_{\text{L1}} + \lambda_2 L_{\text{perceptual}},
\end{equation}
where \(\lambda_1\) and \(\lambda_2\) are empirically set to 1 and 0.2, respectively.

%% file: sections/04c_method_implementation.tex
\section{Implementation Details}
\label{sec:implementation}

We implement our method using PyTorch \cite{paszke2019pytorch} on a single NVIDIA RTX 4090 GPU.
During training, we use a batch size of 4 and a learning rate of $10^{-4}$ with the Adam optimizer \cite{Kingma2014AdamAM}.
The images and events are cropped and resized to $256 \times 256$.
The events are accumulated over $10$ ms and converted to a voxel-grid representation with $10$ channels.

\section{Evaluation}
We now describe the evaluation of our proposed method on the DSEC-Snow dataset and the real snow dataset.
We compare our method with existing state-of-the-art methods for the task of snow occlusion removal.
We also perform ablation studies to understand the effect of events and images on the performance of our method.
Our method is evaluated using the following metrics:
\begin{itemize}
    \item Peak Signal-to-Noise Ratio (PSNR): Higher PSNR indicates better quality of the reconstructed image.
    \item Structural Similarity Index (SSIM): Higher SSIM indicates greater similarity between the reconstructed image and the ground-truth image.
\end{itemize}
In addition, we evaluate the performance of our method on downstream tasks such as object detection, depth estimation, and optical flow using their corresponding standard metrics.

\noindent \textbf{Baselines} 
We consider state-of-the-art single-image desnowing approaches proposed in \cite{Chen22arxiv, Zamir21cvpr}.
In addition, we consider the RLP model proposed in \cite{Zhang23ICCV}, which introduces a novel method for night-time deraining.
This is specifically considered as the rain in the night-time sequences has a similar appearance to snow occlusions.
We also consider the state-of-the-art publicly available video-based desnowing approach S2VD \cite{Yue21cvpr}.
Additionally, we include the E2VID \cite{Rebecq19pami} method, for event-based image reconstruction method.

%% file: sections/03_dataset.tex
\section{Dataset}
\label{sec:dataset}

\label{sec:data:dsecsnow}
\begin{figure*}[t]
    \includegraphics[width=\linewidth]{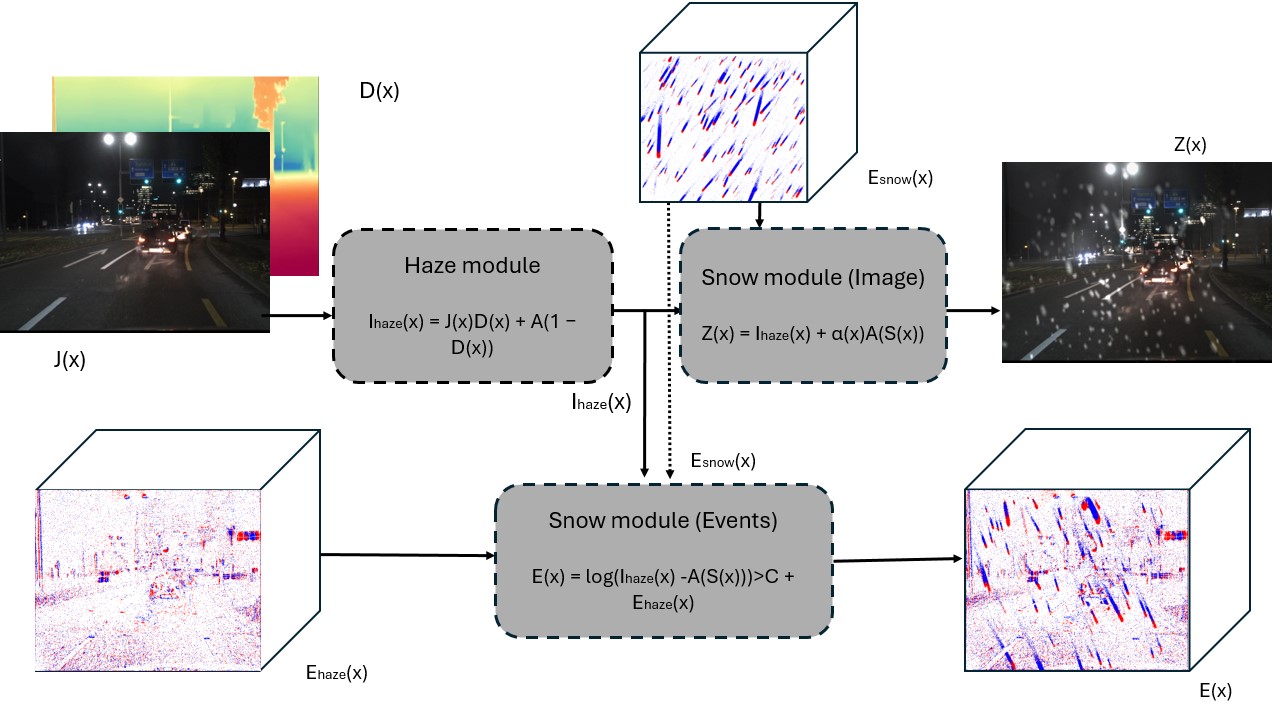}
    \caption{\textbf{Overview of the synthetic snow dataset generation process.}
    Given a clean background image $J(x)$ and its corresponding event stream $E_{\text{haze}}(x)$ and foreground snow event stream $E_{\text{snow}}(x)$, we generate a synthetic snow-occluded image $Z(x)$ and a synthetic event stream $E(x)$.
    The haze module generates a hazy image $I_{\text{haze}}(x)$ using an atmospheric scattering model \cite{li18tip} based on the background image $J(x)$ and its depth map.
    The snow module overlays snow onto the hazy image to produce the snow-occluded image $Z(x)$. 
    Simultaneously, background events $E_{\text{haze}}(x)$ and snow events $E_{\text{snow}}(x)$ are combined to generate the final event stream $E(x)$.}
    \label{fig:data:dsecsnow_datagen}
    \vspace{-1em}
\end{figure*}
To the best of our knowledge, there exists no dataset consisting of events for desnowing.
Moreover, obtaining accurate ground-truth for such a task is also quite challenging.
Previous image-based and video-based approaches relied on simulated datasets such as SnowCity \cite{Zhang21TIP}, SnowKITTI\cite{Zhang21TIP} to train and evaluate their models. However, due to the limited framerate of the cameras, simulating events from these videos is not possible.
We therefore propose to use real events recorded by a physical event camera for both the snow and background movement.
Our dataset uses a popular visual effect technique called ``Chroma key compositing''.
We record two independent sequences consisting of background motion and foreground motion using a real event camera and RGB camera.
The foreground motion is recorded in front of a black screen to have maximum contrast between the snow particles and the background.
These events are then overlaid on the background events as described in \Sec \ref{sec:data:dsecsnow}.
To show the performance of our approach with real snowfall, we also record driving sequences in snowfall using a real event camera.
We describe this in detail in \Sec \ref{sec:data:davissnow}.

\vspace{-1em}
\begin{algorithm}[t]
\caption{Synthetic Event Stream Generation Using Real Background and Snow Occlusion Events}
\label{alg:synth_event_compact}
\begin{algorithmic}[1]
\Require Clean background image $J(x)$, background events $E_{\mathrm{haze}}(x)$, snow events $E_{\mathrm{snow}}(x)$, haze parameters $A$.
\Ensure Snow-occluded image $Z(x)$, synthetic events $E(x)$.

\State Compute depth map $D(x)$ from $J(x)$ using a depth estimation model~\cite{Ranftl2021}
\State Generate haze image $I_{\mathrm{haze}}(x)$ using $D(x)$,, haze parameters $A$, and atmospheric scattering model~\cite{li18tip}
\State Overlay snow events on $I_{\mathrm{haze}}(x)$ to obtain snow-occluded image $Z(x)$

\State Initialize synthetic event stream: $E(x) \gets \emptyset$

\ForAll{events $e_{\mathrm{snow}} \in E_{\mathrm{snow}}(x)$}
  \State Let $(x, y)$ be the event location and $t$ the timestamp
  \If{$|I_{\mathrm{haze}}(x, y) - I_{\mathrm{snow}}(x, y)| > C$}
    \State Add $e_{\mathrm{snow}}$ to $E(x)$
  \EndIf
\EndFor

\ForAll{events $e_{\mathrm{haze}} \in E_{\mathrm{haze}}(x)$}
  \State Let $(x, y)$ be the event location and $t$ the timestamp
  \If{there is no overlapping snow event $e_{\mathrm{snow}}$ at $(x, y, t)$ in $E(x)$}
    \State Add $e_{\mathrm{haze}}$ to $E(x)$
  \EndIf
\EndFor

\State \Return $Z(x)$, $E(x)$

\end{algorithmic}
\end{algorithm}

\subsection{\DSECSnow dataset}
Accurate simulation of image degradation due to weather conditions has been a popular research topic as it provides a supervision signal in the form of ground-truth clean images, the ability to generate large-scale datasets, and a benchmark to evaluate methods.
Generating synthetic snow-occluded images and video has been proposed in the past \cite{Chen23ICCV, Zhang21TIP}, which overlay realistic snow occlusions on top of clean images, providing ground-truth for training and evaluation.
We adopt a similar approach for generating the snow-occluded images in our dataset.
The clean image ($J(x)$) is taken from the DSEC dataset \cite{MGehrig21ral}, which consists of driving sequences recorded using a Prophesee event camera and RGB camera mounted on a car while driving in different cities in Switzerland.
The snow occlusions are separately recorded using a real event camera during a snowfall.
These snow occlusions are overlaid on the background image to produce a snow-occluded image ($I_{snow}(x)$), similar to the approach used in \cite{Chen23ICCV}.
To add realistic snow image degradation, we also render haze on the background image based on the atmospheric scattering model \cite{li18tip}:
\begin{equation}
    I_{haze}(x) = J(x) \cdot t(x) + A \cdot (1 - t(x)),
\end{equation} 
where $t(x)$ is the transmission map, $A$ is the atmospheric light, and $J(x)$ is the background image.
The transmission map is computed using the depth map of the image computed by \cite{Ranftl2021}.
The snow rendering is done by overlaying the snow occlusion on the background image:
\begin{equation}
    Z(x) = I_{haze}(x) + \alpha \cdot Aug(E_{snow}(x)),
    \label{eq:snow_rendering}
\end{equation}
where $Z(x)$ is the final image, $\alpha$ models ambient illumination of the scene \cite{Chen23ICCV}, and $Aug(E_{snow}(x))$ is the augmentation function that simulates the occluded image from foreground snow events.

While such realistic rendering of snow occlusion is possible for images, for events, generating events that simulate the real world and the real sensor is quite challenging and still an open problem \cite{Delbruck20arxiv}.
We therefore propose to instead use a unique method to generate a synthetic dataset by combining multiple event streams recorded with a real event camera, as shown in \Fig \ref{fig:data:dsecsnow_datagen}.
It consists of recording two event streams, one corresponding to background motion resulting from camera movement ($E_{haze}$) and a second event stream corresponding to the motion of occlusion such as snow ($E_{snow}$).
These events are combined using the process described below.

Incorporating the ``chroma key compositing'' technique, we record the snow particles in front of a black screen.
These foreground events are merged with the background events using two main criteria:
\begin{itemize}
    \item Background intensity correction:
    Events are only generated if there exists a contrast between the snowflake ($I_\mathrm{snow}$) and the background. Since snowflakes are typically bright \cite{Garg05ICCV}, brighter backgrounds will not generate events even if a snowflake moves across a bright pixel.
    \item Background events overlap: In the case where background activity generates events at the same time as the occlusion, priority is given to the occlusion event as the occlusion is typically in front of the scene.
\end{itemize}
The algorithm described above is also summarized in \Alg \ref{alg:synth_event_compact}
Examples sequences are shown in \Fig \ref{fig:data:gtdataset}.
More details about our dataset can be found in the supplementary material.
\input{floats/fig_data_gtdataset.tex}
\vspace{-1em}
\subsection{\DAVISSnow dataset}
\label{sec:data:davissnow}
\input{floats/fig_poster_dataset}
To evaluate the model on real data, we propose to collect our own dataset.
Evaluating with real snowfall, however, is quite challenging as there is no ground-truth available.
We therefore propose to use a controlled setup to evaluate the performance in the real world, using a snow machine.
We use a linear slider to move the event camera at a fixed velocity while recording the snowfall using a snow machine.
This allows us to record sequences with snowfall and ground-truth data.
The details about the setup and data generation process are shown in \Fig \ref{fig:slider_setup} and elaborated in the supplementary material.
\input{floats/fig_slider_setup.tex}

\input{floats/fig_dataset_realdriving.tex}
\vspace{-1em}
\subsection{Real Snowfall Driving Sequences}
\label{sec:data:real_snow}
Finally, we also collect a new dataset consisting of real snowfall driving sequences.
We use the BeamSplitter setup of the Prophesee event camera  \cite{Finateu20isscc} and FLIR BlackFly S global shutter RGB camera mounted on the dashboard of the car while driving in the snowfall.
Of course, as there is no ground-truth available, we only use these sequences for qualitative evaluation.
See examples in \Fig \ref{fig:data:gtdataset}.

%% file: floats/fig_data_gtdataset.tex
\begin{figure}[!t]
    \centering
    \newcommand{\thisfigWidth}{0.3\linewidth}
    \newcommand{\basepath}{floats/images/dataset_images/dsec_v11}
    \newcommand{\basepathbsplit}{floats/images/dataset_images/poster_bsplit}
    \begin{tabular}{M{\thisfigWidth}M{\thisfigWidth}M{\thisfigWidth}}

        \includegraphics[width=\linewidth]{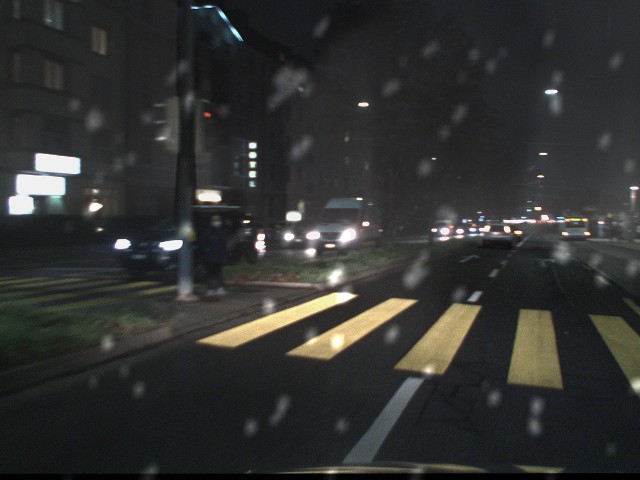}&
        \includegraphics[width=\linewidth]{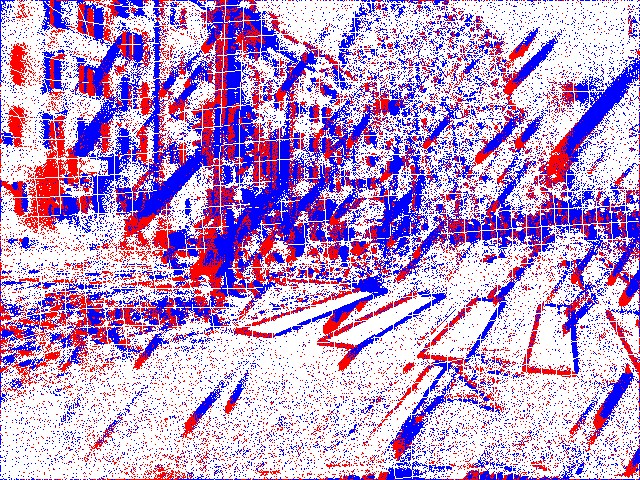}&
        \includegraphics[width=\linewidth]{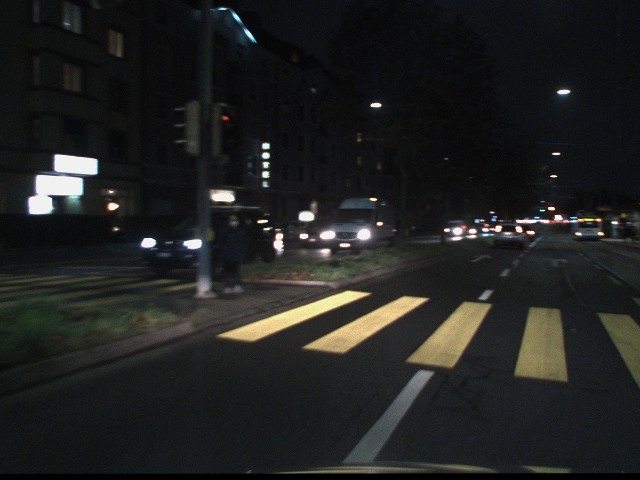}\\

        \includegraphics[width=\linewidth]{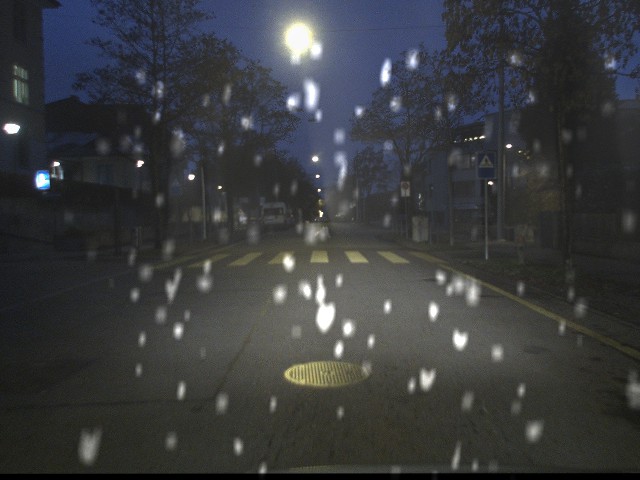}&
        \includegraphics[width=\linewidth]{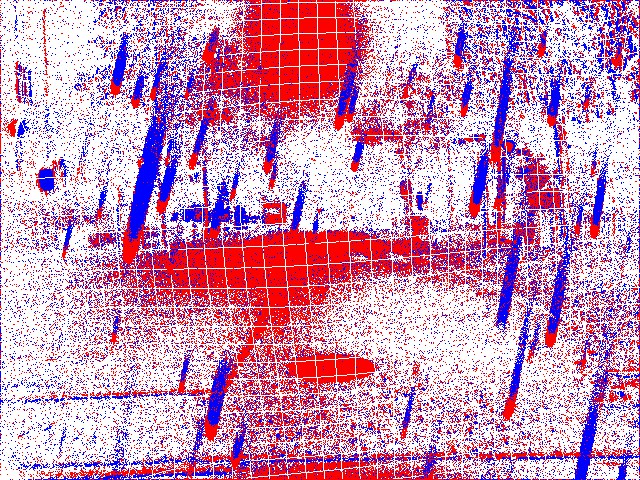}&
        \includegraphics[width=\linewidth]{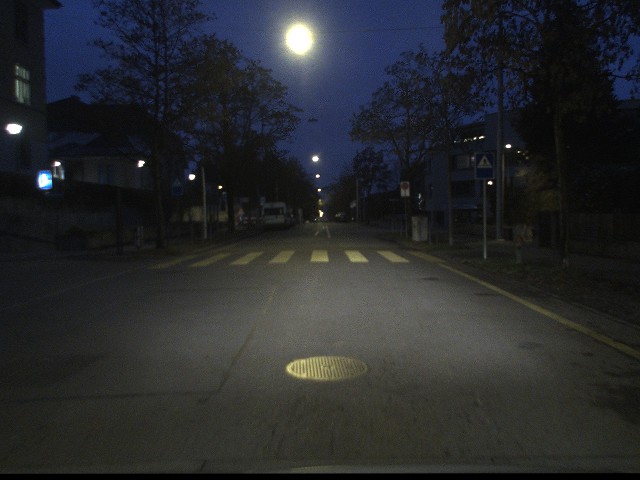}\\

        \includegraphics[width=\linewidth]{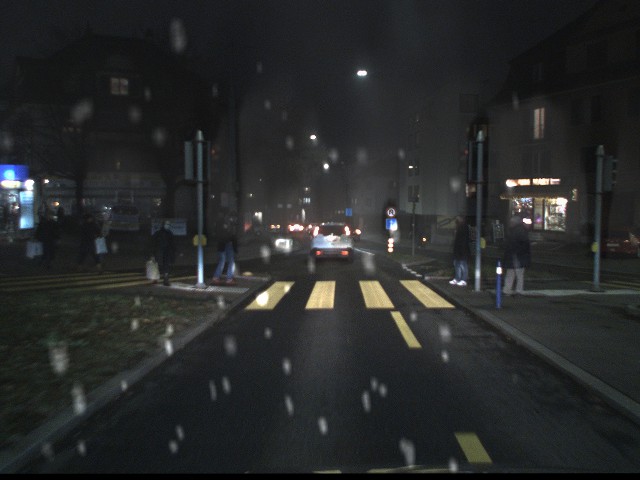}&
        \includegraphics[width=\linewidth]{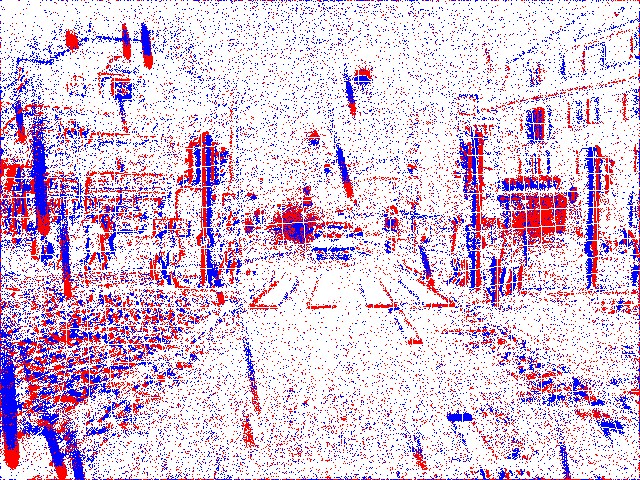}&
        \includegraphics[width=\linewidth]{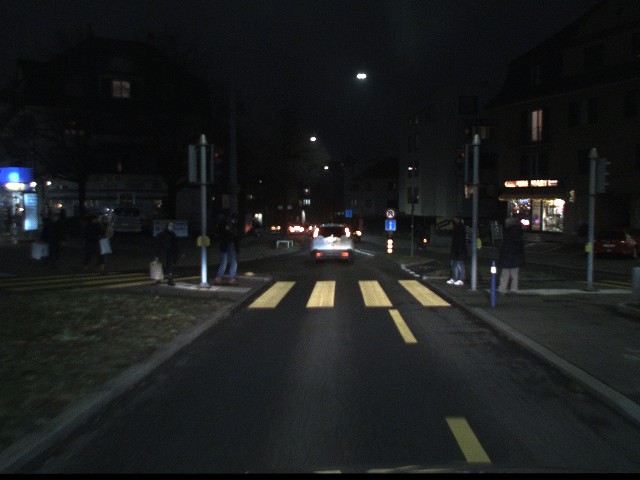}\\

    \end{tabular}
    \caption{\textbf{Example scenes from our  \DSECSnow dataset}. It consists of synchronized RGB frames(Left), Events (Middle) and Groundtruth (Right).}
    \label{fig:data:dsec_snow_example}
\end{figure}

%% file: floats/fig_poster_dataset.tex
\begin{figure}[!t]
    \centering
    \newcommand{\thisfigWidth}{0.3\linewidth}
    \newcommand{\basepath}{floats/images/dataset_images/dsec_v11}
    \newcommand{\basepathbsplit}{floats/images/dataset_images/poster_bsplit}
    \begin{tabular}{M{\thisfigWidth}M{\thisfigWidth}M{\thisfigWidth}}

        \includegraphics[width=\linewidth]{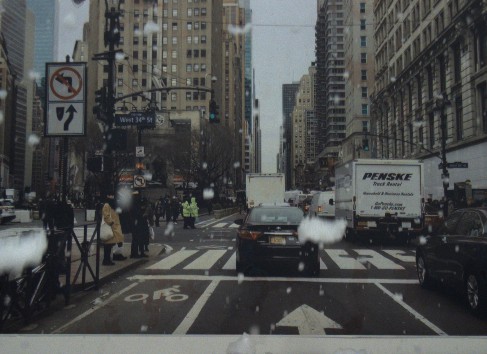}&
        \includegraphics[width=\linewidth]{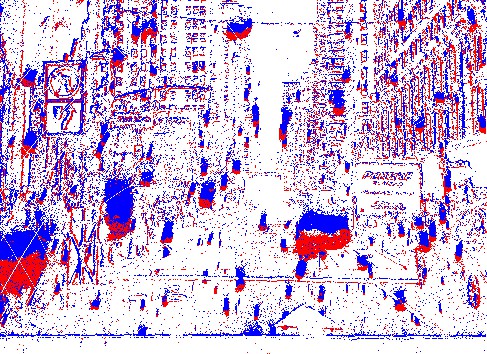}&
        \includegraphics[width=\linewidth]{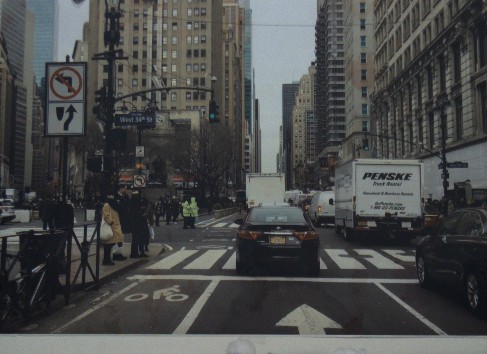}\\
        
        \includegraphics[width=\linewidth]{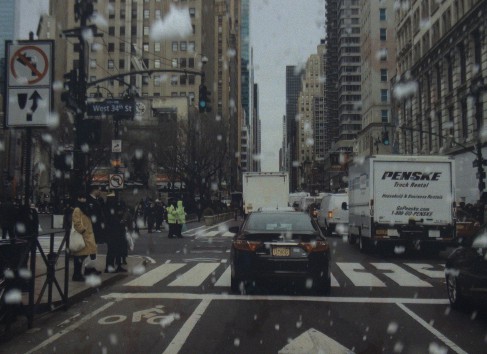}&
        \includegraphics[width=\linewidth]{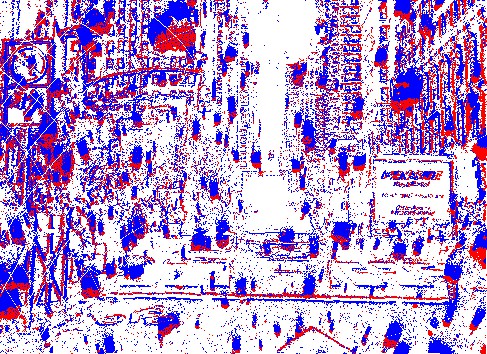}&
        \includegraphics[width=\linewidth]{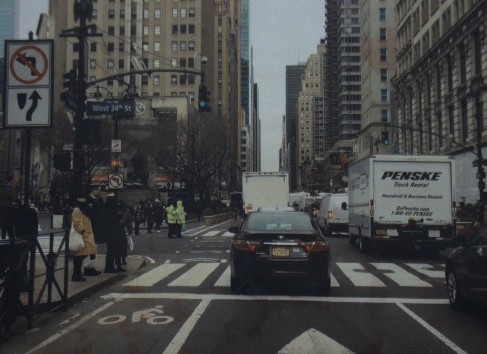}\\
        
        \includegraphics[width=\linewidth]{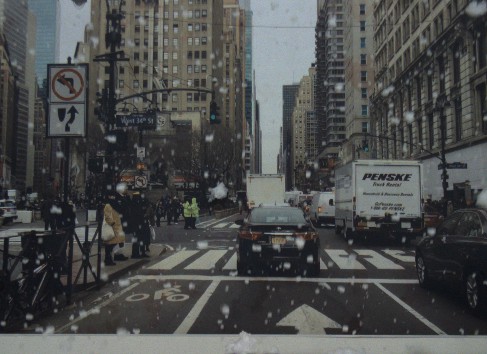}&
        \includegraphics[width=\linewidth]{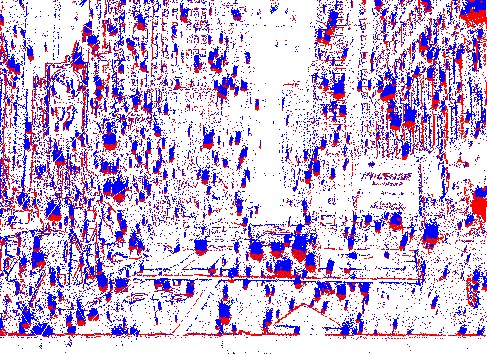}&
        \includegraphics[width=\linewidth]{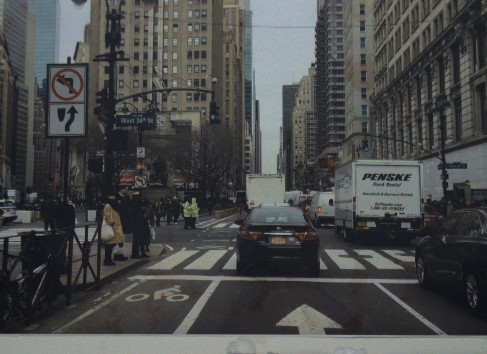}\\

    \end{tabular}
    \caption{\textbf{Examples scenes from our  \DAVISSnow dataset} It consists of synchronized RGB frames(Left), Events (Middle) and Groundtruth (Right).}
    \label{fig:data:gtdataset}
    \vspace{-2em}
\end{figure}

%% file: floats/fig_slider_setup.tex
\begin{figure}[!ht]
    \centering
    \begin{subfigure}[b]{0.45\linewidth}
        \includegraphics[trim={0cm, 4cm, 0cm, 0cm},clip,width=\linewidth]{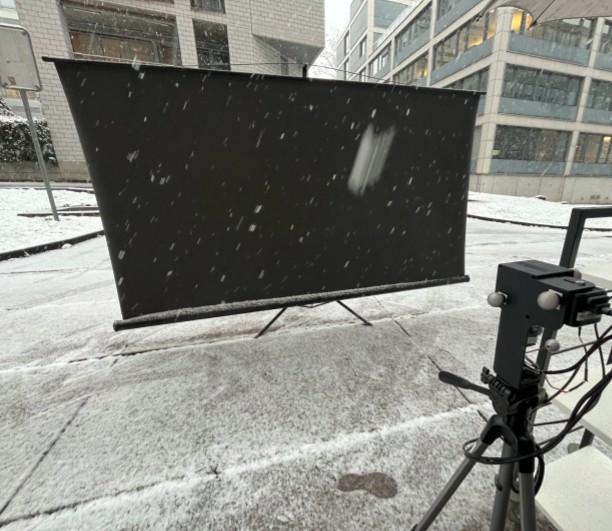}
        \caption{}
        \label{fig:greenscreen-setup}
    \end{subfigure}
    \hfill %
    \begin{subfigure}[b]{0.506\linewidth}
        \includegraphics[trim={0cm, 0cm, 5cm, 0cm},clip,width=\linewidth]{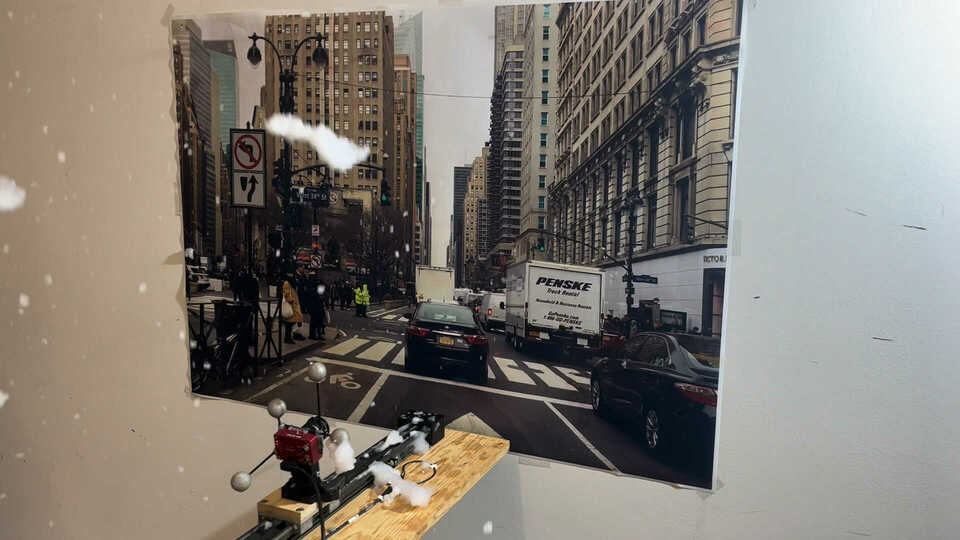}
        \caption{}
        \label{fig:slider-setup}
    \end{subfigure}
    \caption{\textbf{Overview of the experimental setups} (a) Our setup for recording the foreground occlusion events used for generating \DSECSnow sequence. (b) Experimental setup for our controlled real-world dataset. The scene was printed on a poster and the camera was placed on a linear slider with a \rev{fixed velocity}. The snowfall was simulated using a snow machine.}
    \label{fig:slider_setup}
    \vspace{-1em}
\end{figure}

%% file: floats/fig_dataset_realdriving.tex
\begin{figure*}[ht]
    \centering
    \newcommand{\thisfigWidth}{0.22\linewidth}
    \newcommand{\basepath}{floats/images/dataset_images/drive_snowfall_bsplit/}
    \begin{tabular}{M{\thisfigWidth}M{\thisfigWidth}M{\thisfigWidth}M{\thisfigWidth}}
        \includegraphics[trim={5cm, 2cm, 5cm, 0cm},clip,width=\linewidth]{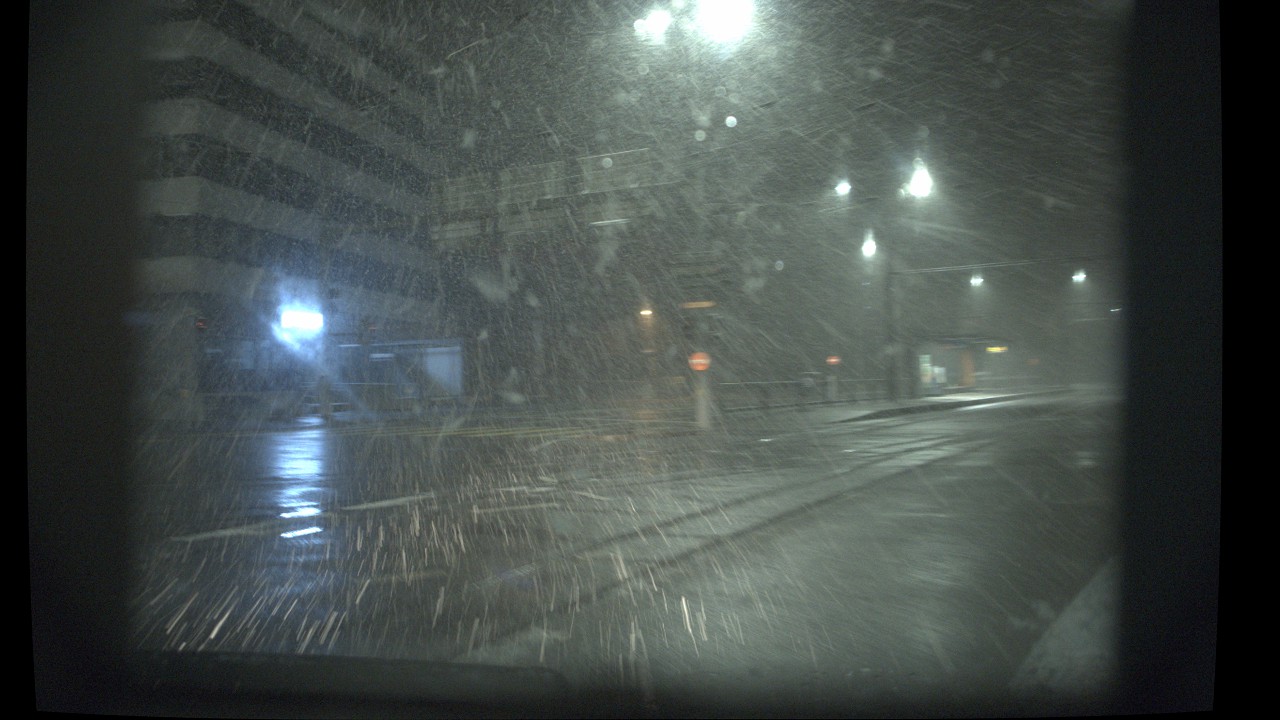} &

        \includegraphics[trim={5cm, 2cm, 5cm, 0cm},clip,width=\linewidth]{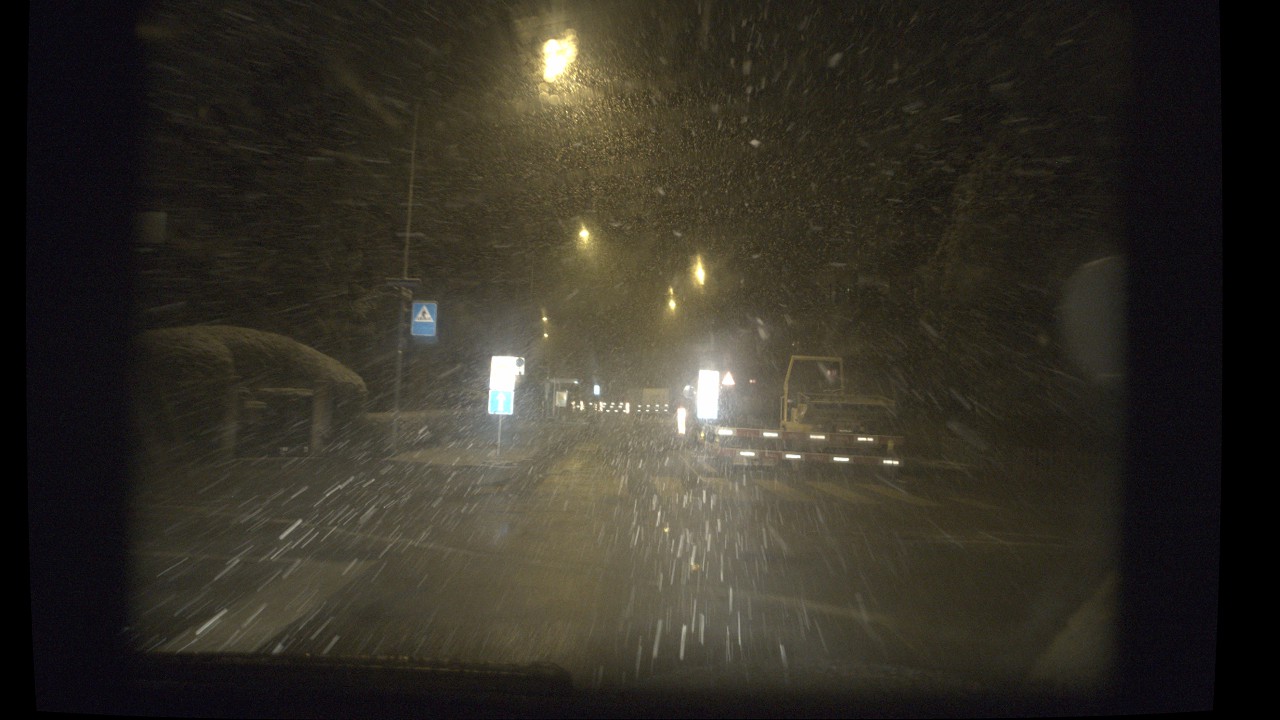} &

        \includegraphics[trim={5cm, 2cm, 5cm, 0cm},clip,width=\linewidth]{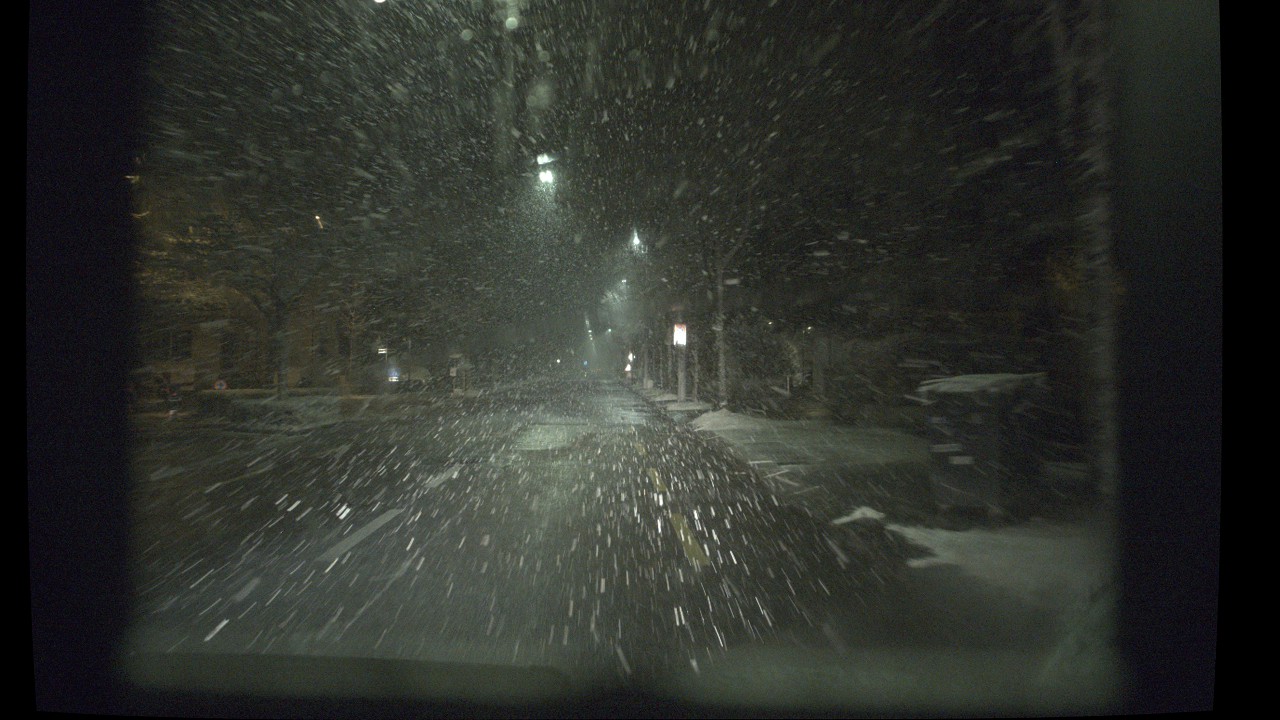} &
        \includegraphics[trim={5cm, 2cm, 5cm, 0cm},clip,width=\linewidth]{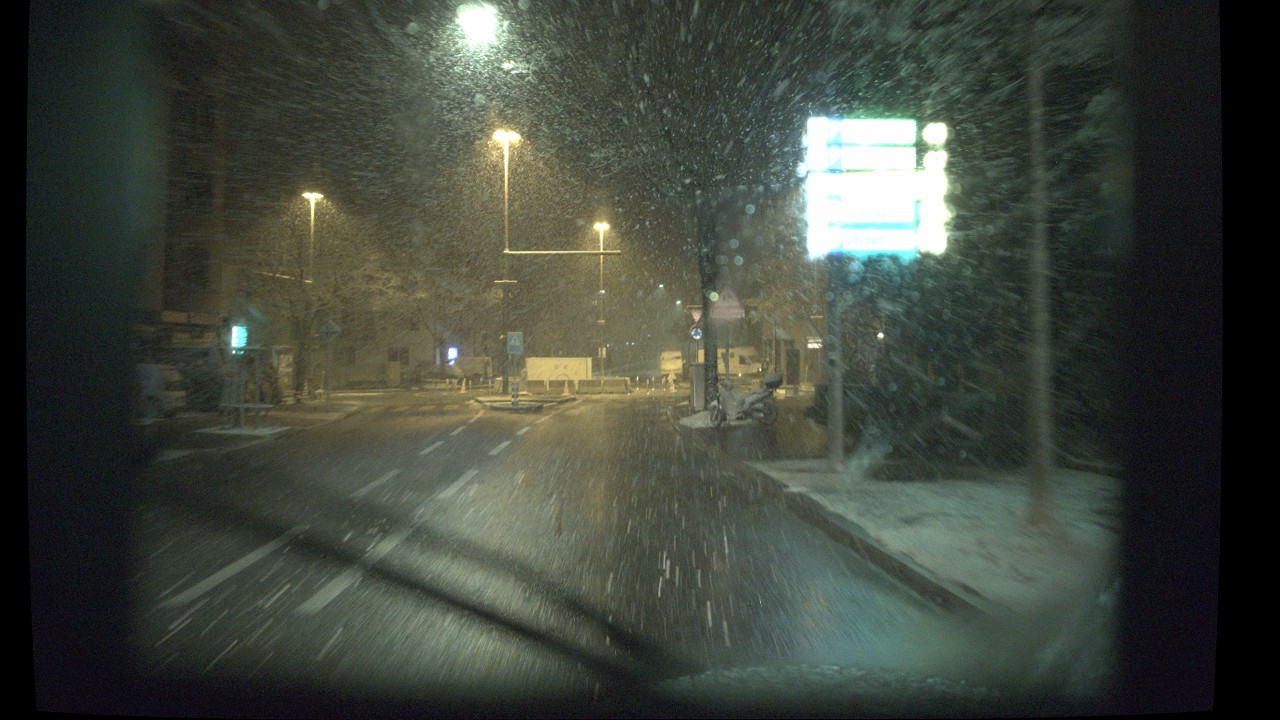} \\

        \includegraphics[trim={2.5cm, 1cm, 2.5cm, 0cm},clip,width=\linewidth]{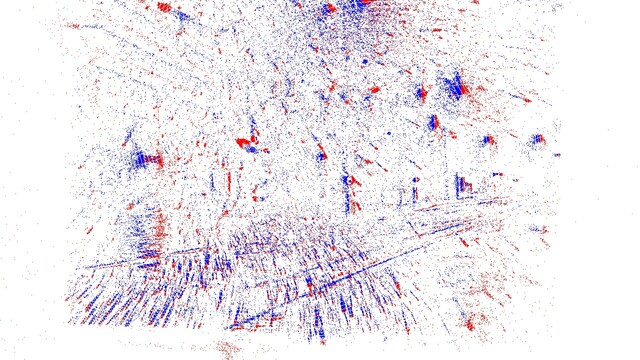}&
        
        \includegraphics[trim={2.5cm, 1cm, 2.5cm, 0cm},clip,width=\linewidth]{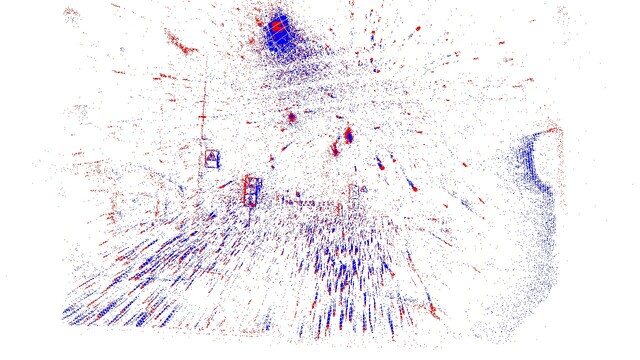}&
        
        \includegraphics[trim={2.5cm, 1cm, 2.5cm, 0cm},clip,width=\linewidth]{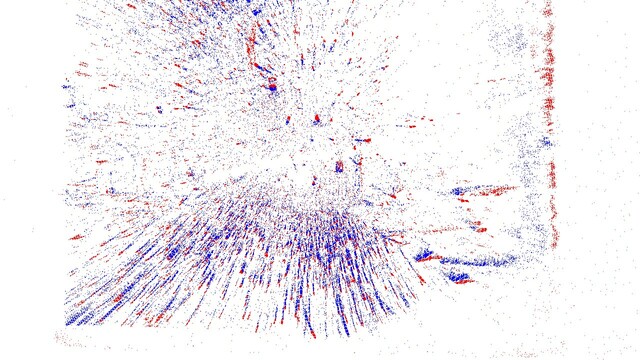}&
        
        \includegraphics[trim={2.5cm, 1cm, 2.5cm, 0cm},clip,width=\linewidth]{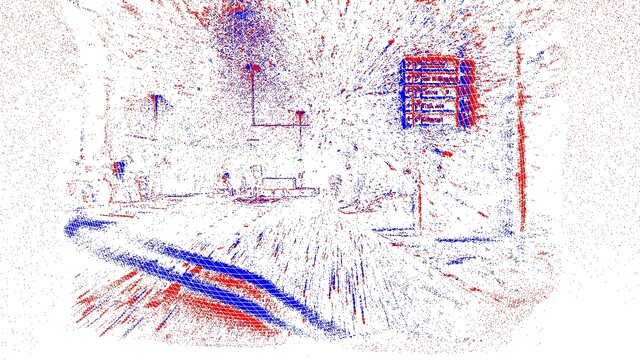}\\

    \end{tabular}
    \caption{\textbf{Samples from our real-world snowfall driving dataset} The images are recorded using the synchronized and aligned setup of RGB camera (top) and event camera (bottom) and  mounted on the dashboard of the car while driving in the snowfall.}
    \label{fig:result_realsnow_driving}
\end{figure*}

%% file: sections/05b_results.tex
\section{Results}
\label{sec:results}
\rev{We evaluate our method on three datasets: the \DSECSnow dataset,  the \DAVISSnow dataset and real-world driving sequences.
We benchmark against existing state-of-the-art methods on the DSEC-Snow dataset in \Sec \ref{sec:result:dsec_snow} and on the \DAVISSnow dataset in \Sec \ref{sec:results:davissnow}.
Furthermore, we provide qualitative results and comparisons on real-world driving datasets in Section \ref{sec:results:realsnow_driving}.
The supplementary material includes more results, ablation studies on network architecture, and evaluations on downstream tasks such as optical flow.
}
\subsection{\DSECSnow Dataset}
\label{sec:result:dsec_snow}
\input{floats/fig_result_dsecsnow.tex}
Table~\ref{tab:results:dsecsnow} presents a quantitative comparison of various image desnowing methods evaluated on the \DSECSnow and \DAVISSnow datasets, reporting PSNR and SSIM for each method. 
The image-based approaches (Restormer, SnowFormer, and RLP) which relying solely on intensity images (I) show moderate performance.
Since RLP was trained for night-time deraining, it struggled to generalize to day-time occlusions which is evident from the low PSNR and SSIM scores.
Video-based methods (S2VD) leverage temporal information from video sequences, achieving better results than single-image methods, but still fall short of the performance of our proposed method, as events provide high resolution temporal information that videos of $20$ fps cannot capture.
Notably, event-only methods (E2VID) perform significantly worse, particularly in terms of SSIM, indicating that events alone are insufficient for effective desnowing in highly occluded scenes.
Model-based fusion of events and images (E+I) shows improved PSNR over single-modality approaches however, the overall image quality is low (indicated by low SSIM scores), suggesting that naive fusion of modalities does not capture the complex spatio-temporal patterns of snow occlusions effectively.
In comparison to all, our method, which fuses both event and intensity information (E+I), achieves the highest PSNR and SSIM across both datasets, outperforming all baselines by a significant margin.
Qualitative results are shown in \Fig \ref{fig:result_dsec_snow}, where our method effectively removes snow occlusions while preserving fine details of the background scene.

\input{floats/tables/tab_dsec_snow.tex}

\noindent \textbf{Ablation on occlusion density}
\input{floats/fig_occlusion_density.tex}
We study the robustness of desnowing methods under varying occlusion conditions in \Fig \ref{fig:result_occlusion_density}. 
Each row in the figure represents a distinct occlusion density, increasing from top to bottom. 
As the occlusion density increases, we observe a substantial degradation in scene visibility and reconstruction quality for the image-only and event-only baselines. 
SnowFormer and RLP exhibit noticeable artifacts and loss of detail under severe occlusions, leaving substantial snow artifacts in the reconstructed outputs. 
In contrast, our method demonstrates consistent robustness across all occlusion densities, effectively removing snow particles and preserving fine scene details. 
These results highlight the benefit of jointly leveraging event and image data, enabling our approach to maintain high-quality reconstructions even in highly adverse weather conditions with dense occlusions.
\vspace{-1em}
\subsection{Results on \DAVISSnow Dataset}
\label{sec:results:davissnow}
\input{floats/fig_result_slidersnow.tex}
We further evaluate our method on a real-world snowfall dataset to assess its generalization ability beyond the synthetic domain. Specifically, we deploy the model trained on our synthetic DSEC-Snow dataset and apply it directly to the \DAVISSnow dataset, which is captured under real snowfall conditions. 
\Fig~\ref{fig:result_davissnow} compares our method with state-of-the-art image-based baselines such as Restormer~\cite{Zamir21cvpr} and SnowFormer~\cite{Chen22arxiv}. 
Our approach (column d) consistently produces better reconstructions compared to the original occluded image (column a), Restormer (column b), and SnowFormer (column c). 
Notably, in the first and third rows, where dense snowflakes heavily obscure vehicle details, our method is able to restore the underlying vehicle contours and even fine-grained elements such as traffic signs, which are either blurred or completely lost in the baseline. 
\Tab \ref{tab:results:dsecsnow} presents a quantitative comparison of all baselines on \DAVISSnow dataset.
Although certain image-based baselines, such as Restormer\cite{Zamir21cvpr} achieve slightly higher PSNR, qualitative comparisons indicate that they fail to effectively remove occlusions, as illustrated in \Fig~\ref{fig:result_davissnow}.
These qualitative improvements highlight the strength of using event data for temporally consistent occlusion-aware image reconstruction. By attending to motion patterns in the event stream, our model infers occluded content more robustly than purely image-based approaches. Importantly, the model is not fine-tuned on real-world snow data, indicating strong cross-domain generalization. This result validates the effectiveness of our synthetic dataset generation pipeline and supports its applicability to training models deployable in real-world conditions.
\vspace{-1em}
\subsection{Results on real snowfall driving sequences}
\label{sec:results:realsnow_driving}

We also evaluate our method on real-world snowfall driving sequences acquired using a beamsplitter-based sensor rig comprising a Prophesee Gen4 event camera ($1280\times 720$) and a FLIR BlackFly S global shutter RGB camera ($1440 \times 1080$). This setup enables temporally aligned acquisition of high-dynamic-range (HDR) event data and intensity frames under challenging low-light and high-motion conditions.
Qualitative results are presented in Fig. \ref{fig:result_realdrive}.

Our method demonstrates a clear advantage in removing snow-induced occlusions and recovering underlying scene structure. The baseline RGB-only methods, Restormer \cite{Zamir21cvpr} and Snowformer \cite{Chen22arxiv}, struggle with motion blur and fail to distinguish snowflakes from meaningful scene content, often resulting in over-smoothed or distorted reconstructions. In contrast, our method preserves high-frequency details and recovers semantically relevant features such as traffic signs, barriers, and road markings, even under heavy snowfall.

In \Fig \ref{fig:result_realdrive}, where oncoming headlights and snowflakes dominate the visual field, the image-only baselines suffer from halo artifacts and flare-induced saturation. Our approach effectively suppresses such artifacts, allowing visibility of distant road features, including lane markers and background vehicles. 
This is largely attributable to the asynchronous, high-temporal-resolution nature of the event data, which captures scene dynamics without the integration-based blur inherent to conventional RGB sensors.

The qualitative comparisons in Fig. \ref{fig:result_realdrive} validate the effectiveness of our approach in real-world adverse weather conditions. By leveraging the complementary sensing modalities of events and images, our method achieves robust occlusion removal and scene restoration beyond the capability of state-of-the-art RGB-only models.
\rev{Please see the supplementary material for additional results on real snowfall driving sequences.}
\input{floats/fig_result_driving.tex}
\\
\noindent \textbf{Model Complexity and Parameters Comparison}
We evaluated the computational efficiency of our method by measuring the average runtime on NVIDIA GeForce RTX 4090 GPU.  
Our approach processes a single image with resolution $512 \times 512$ in approximately 0.06 seconds, which is comparable to recent state-of-the-art methods while providing improved desnowing performance. 
The model contains $10.3$ million parameters, which is on par with Snowformer \cite{Chen22arxiv} ($8.38M$) and significantly lower than Restormer \cite{Zamir21cvpr} ($26M$). 
\vspace{-1em}

%% file: floats/fig_result_dsecsnow.tex
\begin{figure*}[!t]
    \centering
    \newcommand{\thisfigWidth}{0.18\linewidth}
    \newcommand{\basepath}{floats/images/dataset_results/new_v11_test/}
    \newcommand{\basepathdavis}{floats/images/dataset_results/new_slider_test/}
    \begin{tabular}{M{\thisfigWidth}M{\thisfigWidth}M{\thisfigWidth}M{\thisfigWidth}M{\thisfigWidth}}
        \includegraphics[width=\linewidth]{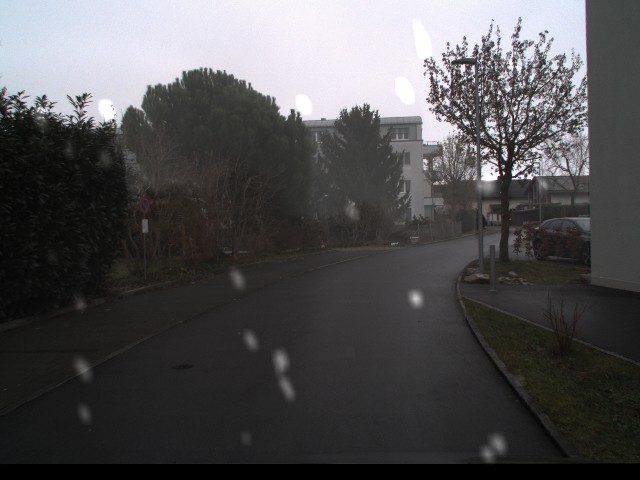}&
        \includegraphics[width=\linewidth]{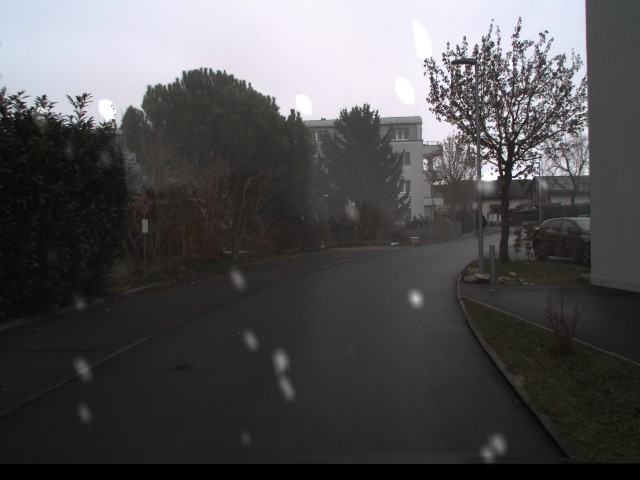}&
        \includegraphics[width=\linewidth]{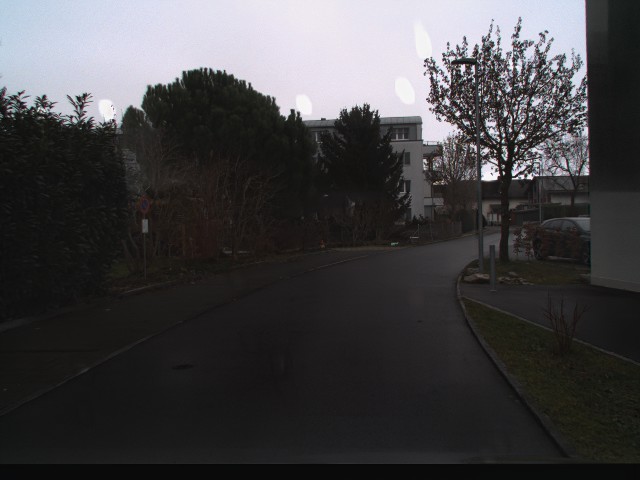}&
        \includegraphics[width=\linewidth]{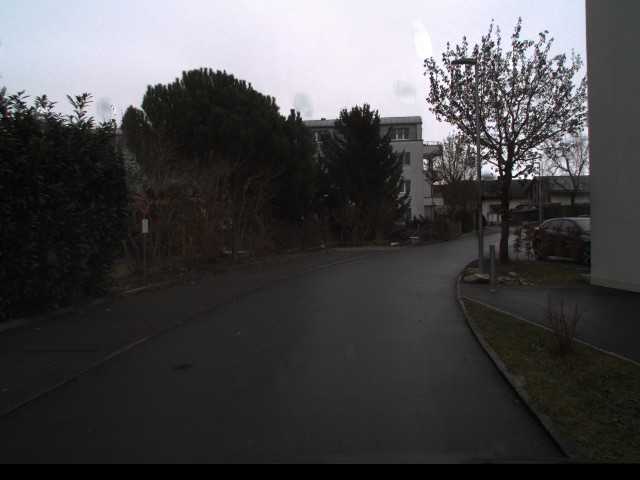}&
        \includegraphics[width=\linewidth]{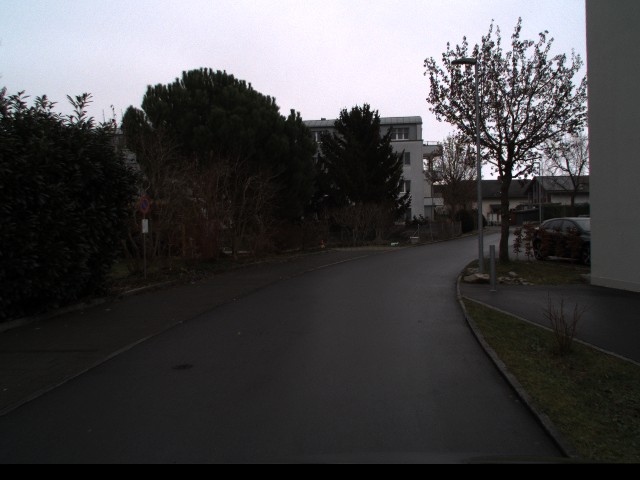}\\

        \includegraphics[width=\linewidth]{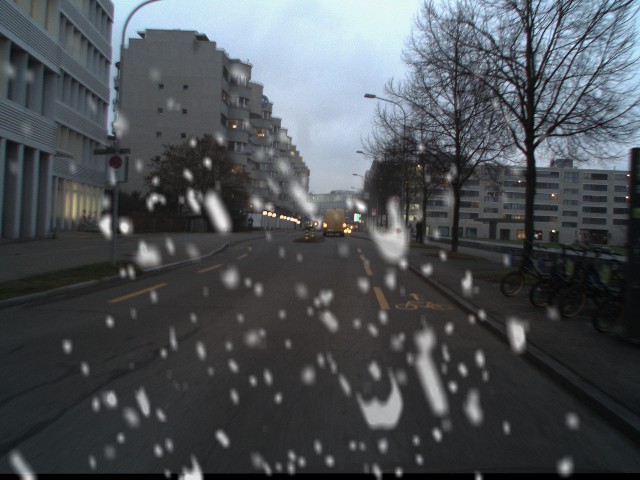}&
        \includegraphics[width=\linewidth]{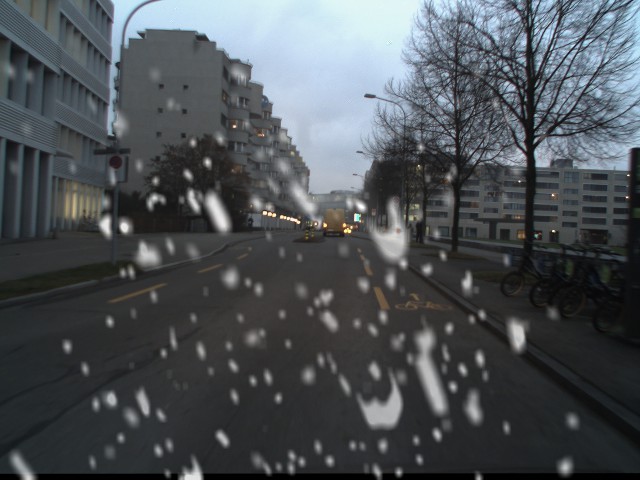}&
        \includegraphics[width=\linewidth]{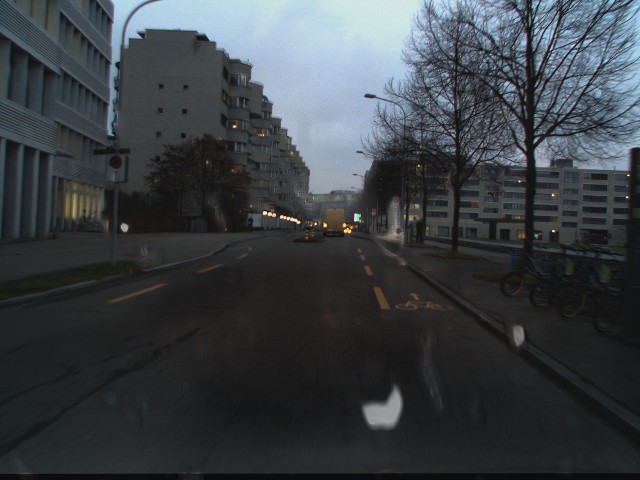}&
        \includegraphics[width=\linewidth]{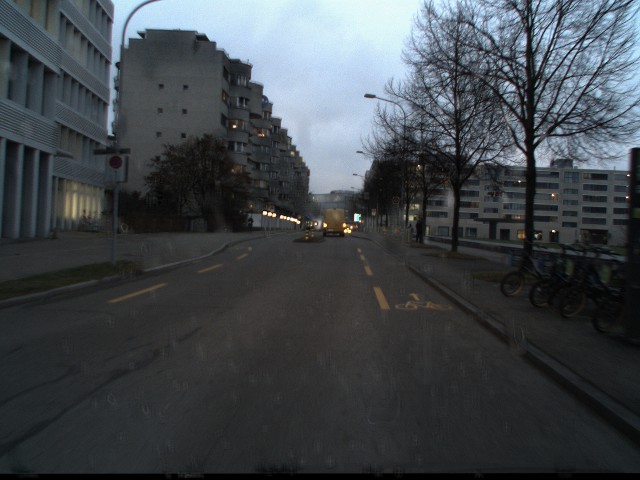}&
        \includegraphics[width=\linewidth]{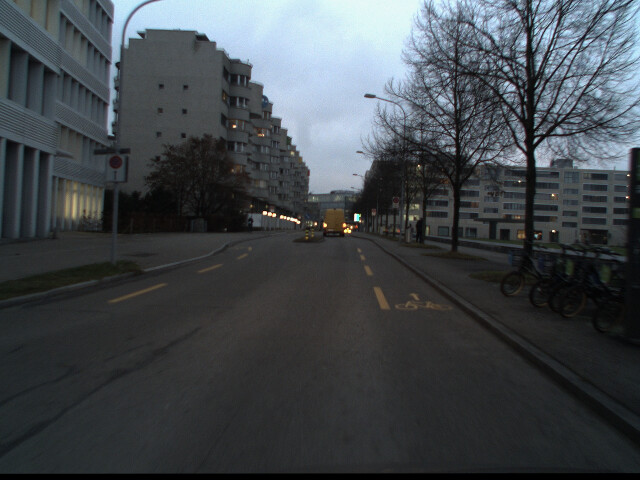}\\

        (a) Image & (b) Restormer \cite{Zamir21cvpr} & (c) Snowformer \cite{Chen22arxiv} & (d) \revtwo{\textbf{Ours}} &(e) GT\\
    \end{tabular}
    \caption{ \textbf{Qualitative comparison on \DSECSnow dataset.}
    (a) Input images with synthetic snow,(b)-(c) image restoration baselines , (d) our result, and (e) ground truth (GT) images. 
    On our synthetic \DSECSnow dataset, our fusion of event and image data enables more effective snow removal and better recovery of occluded details.
    The proposed approach recovers clearer scene structures and more faithfully restores the underlying content compared to prior methods.}
    \label{fig:result_dsec_snow}
    \vspace{-1em}
\end{figure*}

%% file: floats/tables/tab_dsec_snow.tex
\begin{table}[t]
\centering
        \begin{tabular}{lccc|cc}
            \toprule
                    &   &  \multicolumn{2}{c}{\DSECSnow} &  \multicolumn{2}{c}{\DAVISSnow} \\ \hline
            \textbf{Method}        & \textbf{Input}  & \textbf{PSNR} & \textbf{SSIM}   & \textbf{PSNR} & \textbf{SSIM} \\ \hline
            Restormer \cite{Zamir21cvpr}          & I                &  \revtwo{23.12}             & \revtwo{0.8909}          &  \textbf{25.16}  & \textbf{0.8939} \\ 
            SnowFormer \cite{Chen22arxiv}         & I                &  \revtwo{25.12}             & \revtwo{0.9240}          &  17.63   & 0.5593 \\ 
            RLP \cite{Zhang23ICCV}                & I                &  \revtwo{11.20}             & \revtwo{0.5383}          &  10.10   & 0.6062 \\ 
            S2VD \cite{Yue21cvpr}                 & V                &  \revtwo{23.37}             & \revtwo{0.8758}          &  \textbf{25.16}   & 0.8706 \\ \hline
            E2VID \cite{Rebecq19pami}             & E                &  \revtwo{11.84}             & \revtwo{0.3369}          &  15.77   & 0.3416     \\ 
            Model-based                           & E+I              &  \revtwo{21.24}             & \revtwo{0.5656}          &  19.81   & 0.6254\\
            \revtwo{Ours}                         & E + I            & \revtwo{\textbf{31.76}}     & \revtwo{\textbf{0.9686}} &  \textit{24.18}   & \textit{0.8467}\\ 
        \bottomrule
        \end{tabular}
    \caption{\textbf{Quantitative comparison of image desnowing methods on our \DSECSnow and \DAVISSnow datasets}
    We report PSNR and SSIM for each method using different input modalities: intensity images (I), video (V), events (E), and both (E+I). 
    The proposed approach, leveraging both event and image data, demonstrates higher image reconstruction quality compared existing image-based and event-based methods.}
    \label{tab:results:dsecsnow}
    \vspace{-1em}
\end{table}

%% file: floats/fig_occlusion_density.tex
\begin{figure*}[!t]
    \centering
    \newcommand{\thisfigWidth}{0.18\linewidth}

    \begin{tabular}{
	M{\thisfigWidth}M{\thisfigWidth}M{\thisfigWidth}M{\thisfigWidth}M{\thisfigWidth}}
    \includegraphics[width=\linewidth]{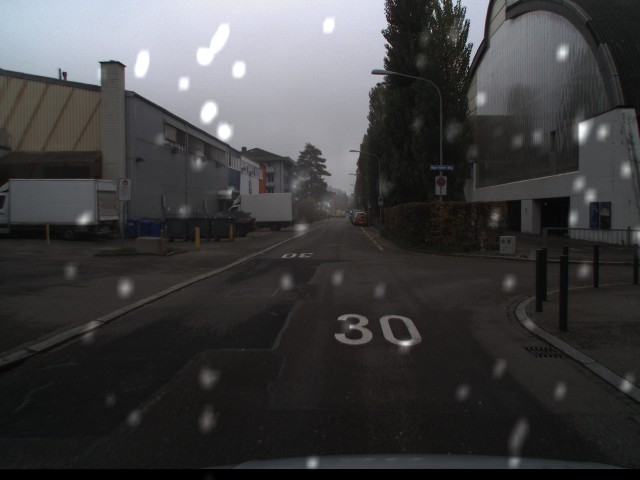}&
    \includegraphics[width=\linewidth]{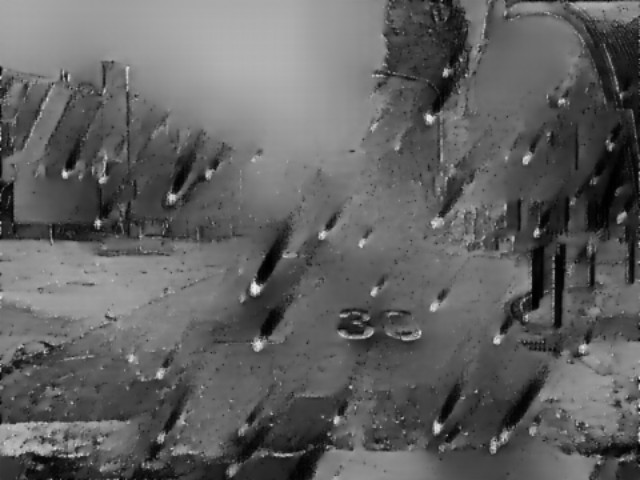}&
    \includegraphics[width=\linewidth]{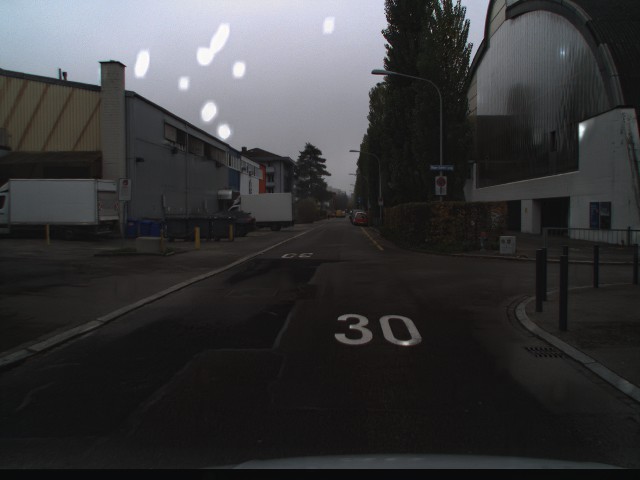}&
    \includegraphics[width=\linewidth]{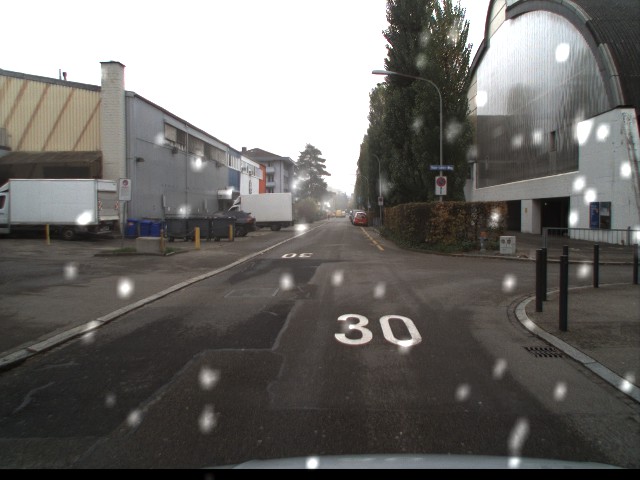}&
    \includegraphics[width=\linewidth]{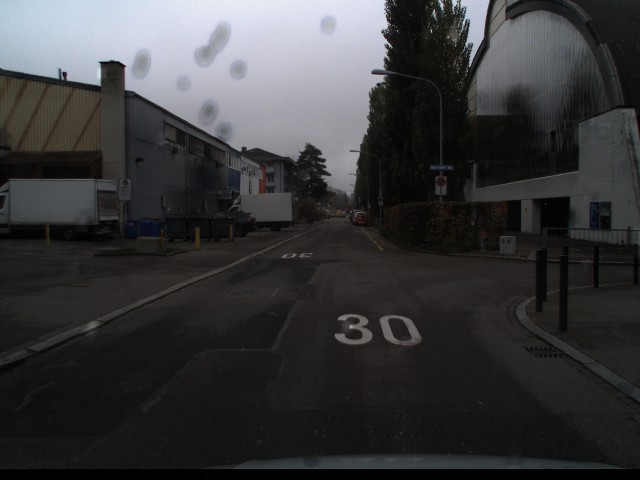}\\

    \includegraphics[width=\linewidth]{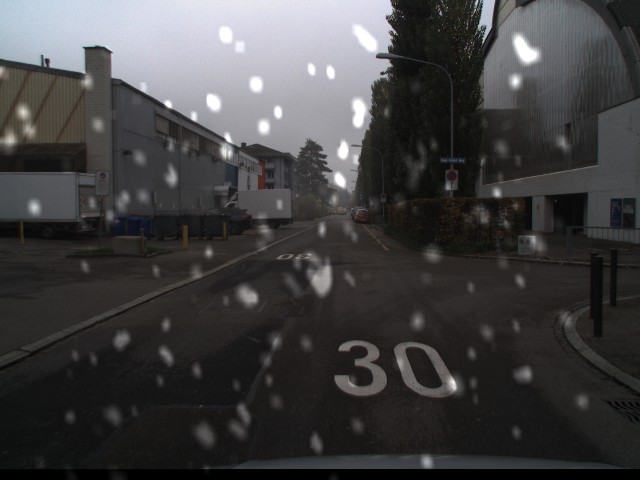}&
    \includegraphics[width=\linewidth]{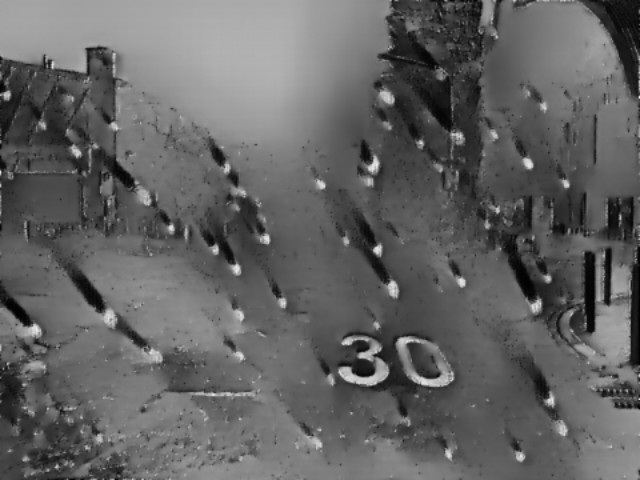}&
    \includegraphics[width=\linewidth]{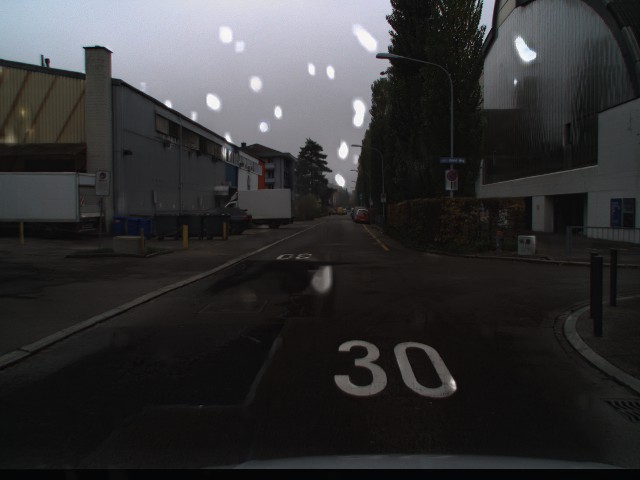}&
    \includegraphics[width=\linewidth]{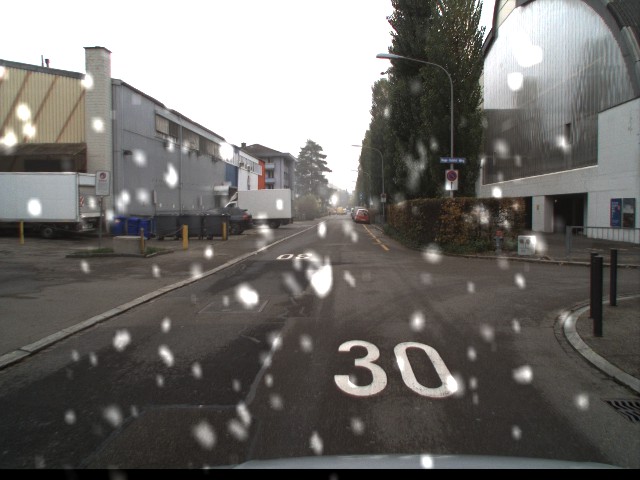}&
    \includegraphics[width=\linewidth]{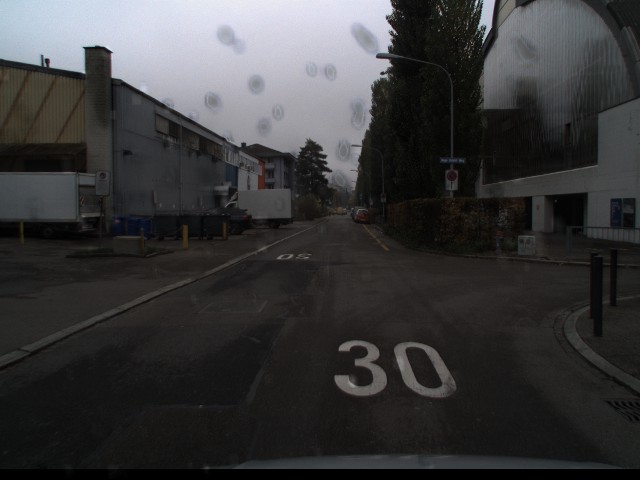}\\

    \includegraphics[width=\linewidth]{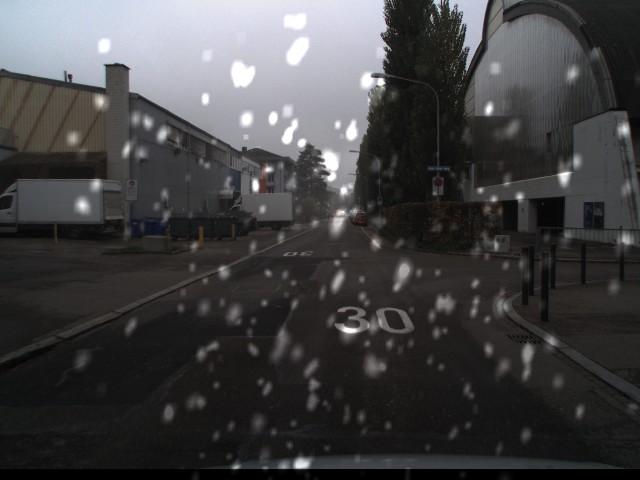}&
    \includegraphics[width=\linewidth]{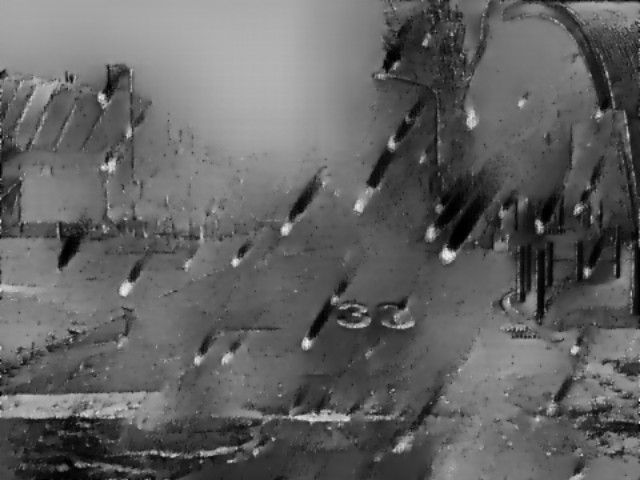}&
    \includegraphics[width=\linewidth]{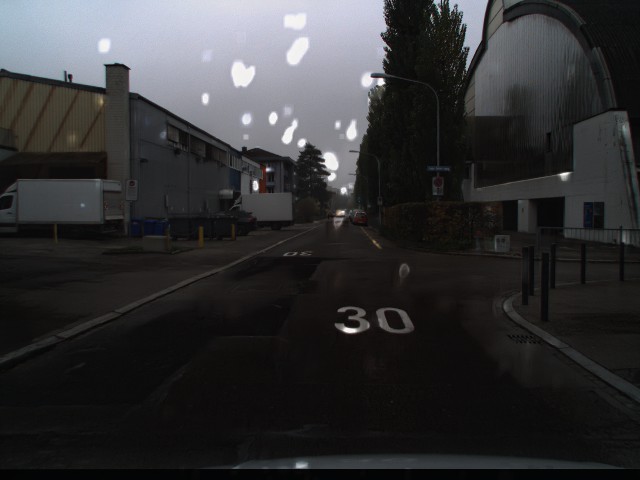}&
    \includegraphics[width=\linewidth]{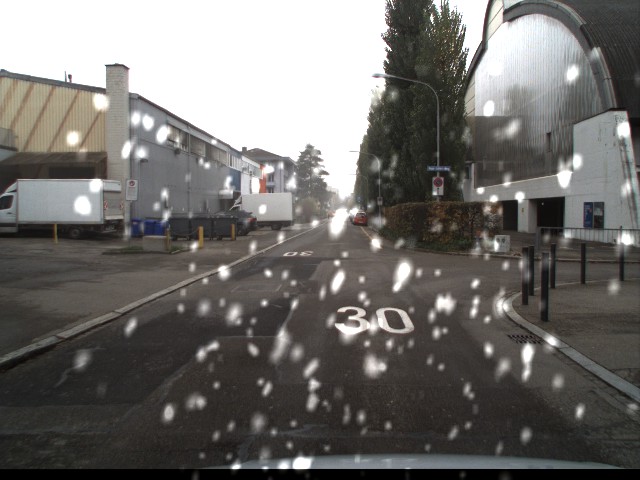}&
    \includegraphics[width=\linewidth]{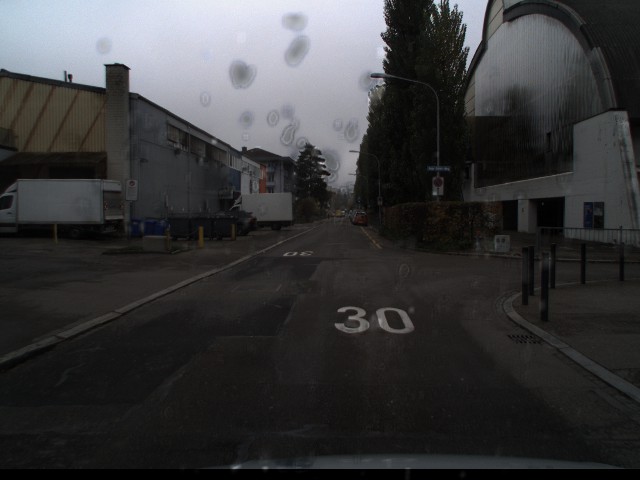}\\
    
    (a) Image & (b) E2VID \cite{Rebecq19pami} & (c) Snowformer \cite{Chen22arxiv} & (d) RLP \cite{Zhang23ICCV} & (e) \revtwo{Ours} \\
    \end{tabular}
     \caption{\textbf{Effect of occlusion density on image reconstruction quality}
      Each row corresponds to a different level of occlusion density, increasing from top to bottom.
      As occlusion density increases, visibility of scene details and robustness of the image-only algorithm degrade significantly.
      Our method performs consistently well across all occlusion densities, effectively removing snow occlusions and preserving scene details.
      }
    \label{fig:result_occlusion_density}
\vspace{-1em}
\end{figure*}

%% file: floats/fig_result_slidersnow.tex
\begin{figure*}[!t]
    \centering
    \newcommand{\thisfigWidth}{0.18\linewidth}
    \newcommand{\basepath}{floats/images/dataset_results/new_v11_test/}
    \newcommand{\basepathdavis}{floats/images/dataset_results/new_slider_test/}
    \begin{tabular}{M{\thisfigWidth}M{\thisfigWidth}M{\thisfigWidth}M{\thisfigWidth}M{\thisfigWidth}}

        \includegraphics[width=\linewidth]{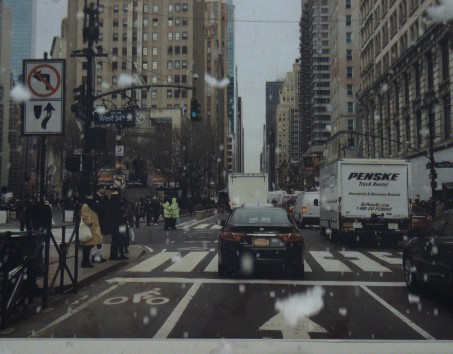}&
        \includegraphics[trim={0.5cm 0cm 0.5cm 0cm},clip,width=\linewidth]{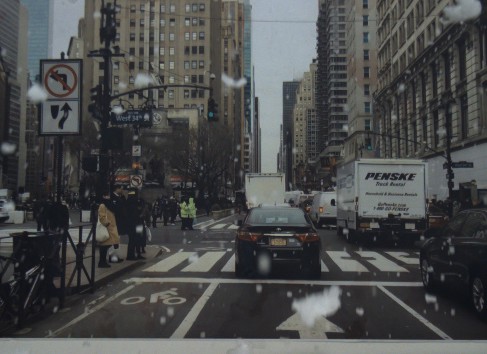}&
        \includegraphics[trim={0.5cm 0cm 0.5cm 0cm},clip,width=\linewidth]{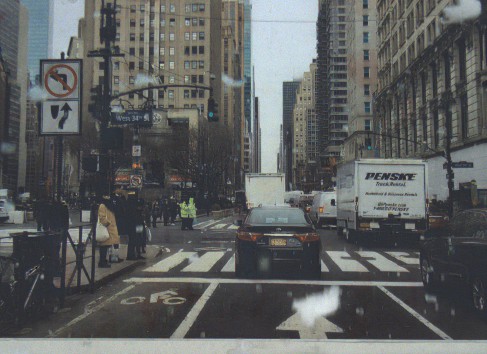}&
        \includegraphics[trim={0.5cm 0cm 0.5cm 0cm},clip,width=\linewidth]{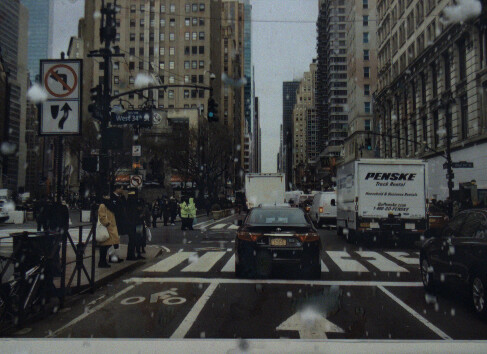}&
        \includegraphics[trim={0cm 0cm 0cm 0cm},clip,width=\linewidth]{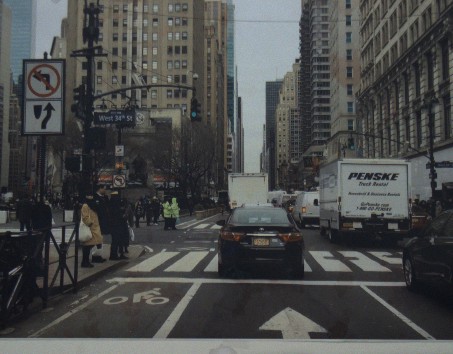}\\

        \includegraphics[width=\linewidth]{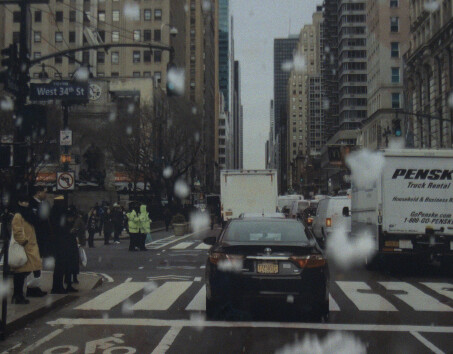}&
        \includegraphics[trim={0.5cm 0cm 0.5cm 0cm},clip,width=\linewidth]{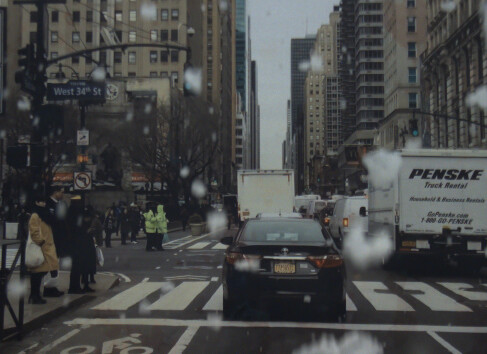}&
        \includegraphics[trim={0.5cm 0cm 0.5cm 0cm},clip,width=\linewidth]{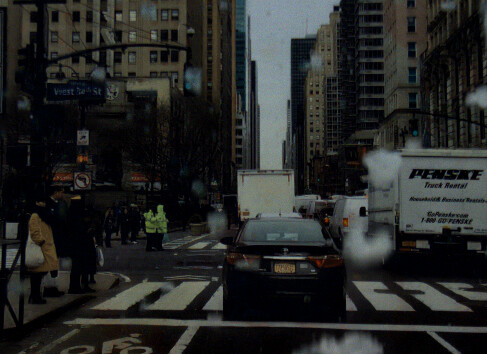}&
        \includegraphics[width=\linewidth]{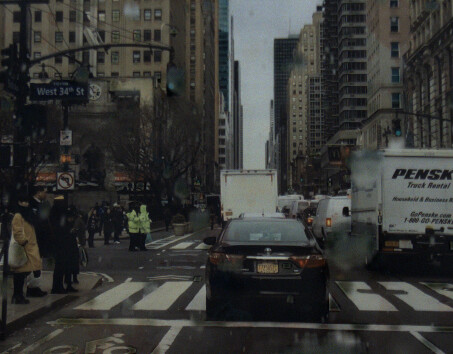}&
        \includegraphics[trim={0.5cm 0cm 0.6cm 0cm},clip,width=\linewidth]{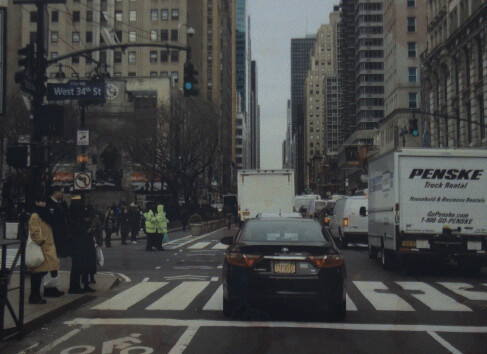}\\
        (a) Image & (b) Restormer \cite{Zamir21cvpr} & (c) Snowformer \cite{Chen22arxiv} & (d) \revtwo{\textbf{Ours}} &(e) GT\\
    \end{tabular}
    \caption{ \textbf{Qualitative comparison of image desnowing results on \DAVISSnow data.}
    (a) Input images with synthetic snow,(b)-(c) image restoration baselines , (d) our result, and (e) ground truth (GT) images. 
    On real-world \DAVISSnow data, we show similar improvements in snow removal and detail recovery, demonstrating the strong generalization of our method to real snow conditions.
    The proposed approach recovers clearer scene structures and more faithfully restores the underlying content compared to prior methods.}
    \label{fig:result_davissnow}
    \vspace{-1em}
\end{figure*}

%% file: floats/fig_result_driving.tex
\begin{figure*}[!t]
    \centering
    \newcommand{\thisfigWidth}{0.23\linewidth}
    \newcommand{\labelWidth}{0.005\linewidth}
    \newcommand{\basepath}{floats/images/dataset_results/results_driving_bsplit/}
    \begin{tabular}{M{\thisfigWidth}M{\thisfigWidth}M{\thisfigWidth}M{\thisfigWidth}}

        \includegraphics[trim={12cm, 2.2cm, 6.5cm, 13.5cm},clip,width=\linewidth]{\basepath/14_02_05/original_000068.jpg} &
        \includegraphics[trim={19cm, 2.2cm, 13.5cm, 13.5cm},clip, width=\linewidth]{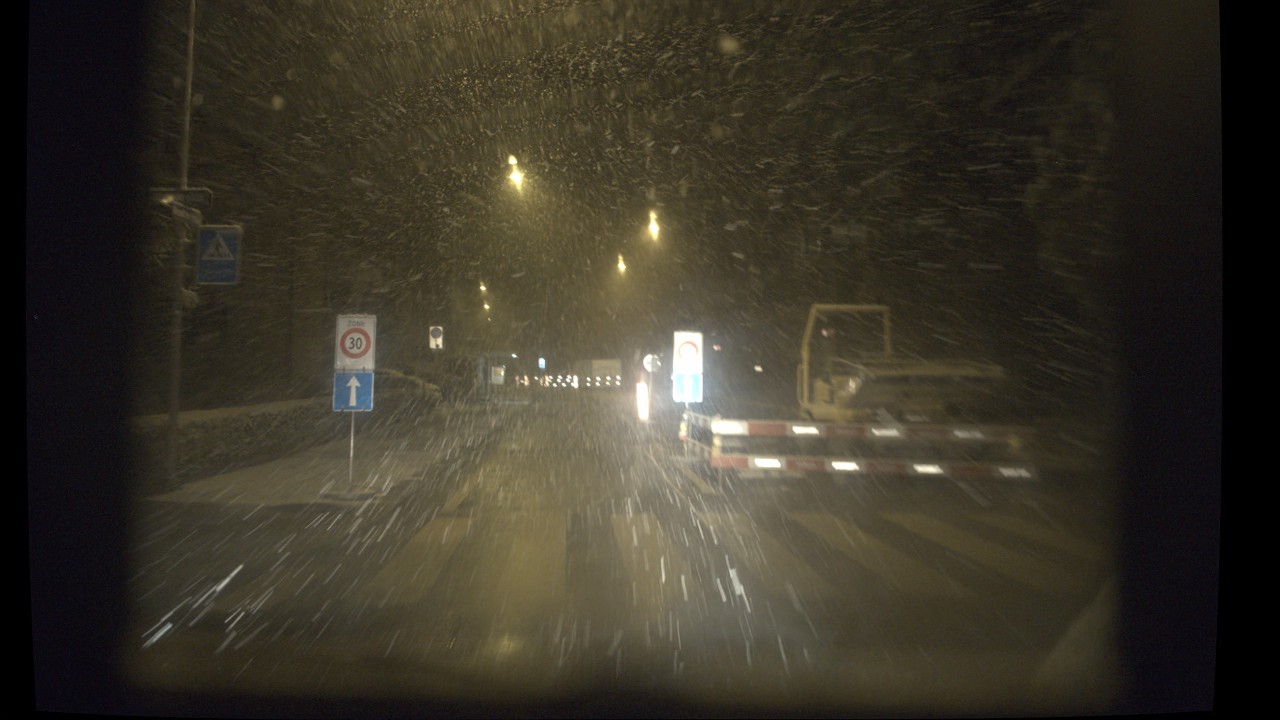} &
        \includegraphics[trim={19cm, 4cm, 13.5cm, 13.5cm},clip,width=\linewidth]{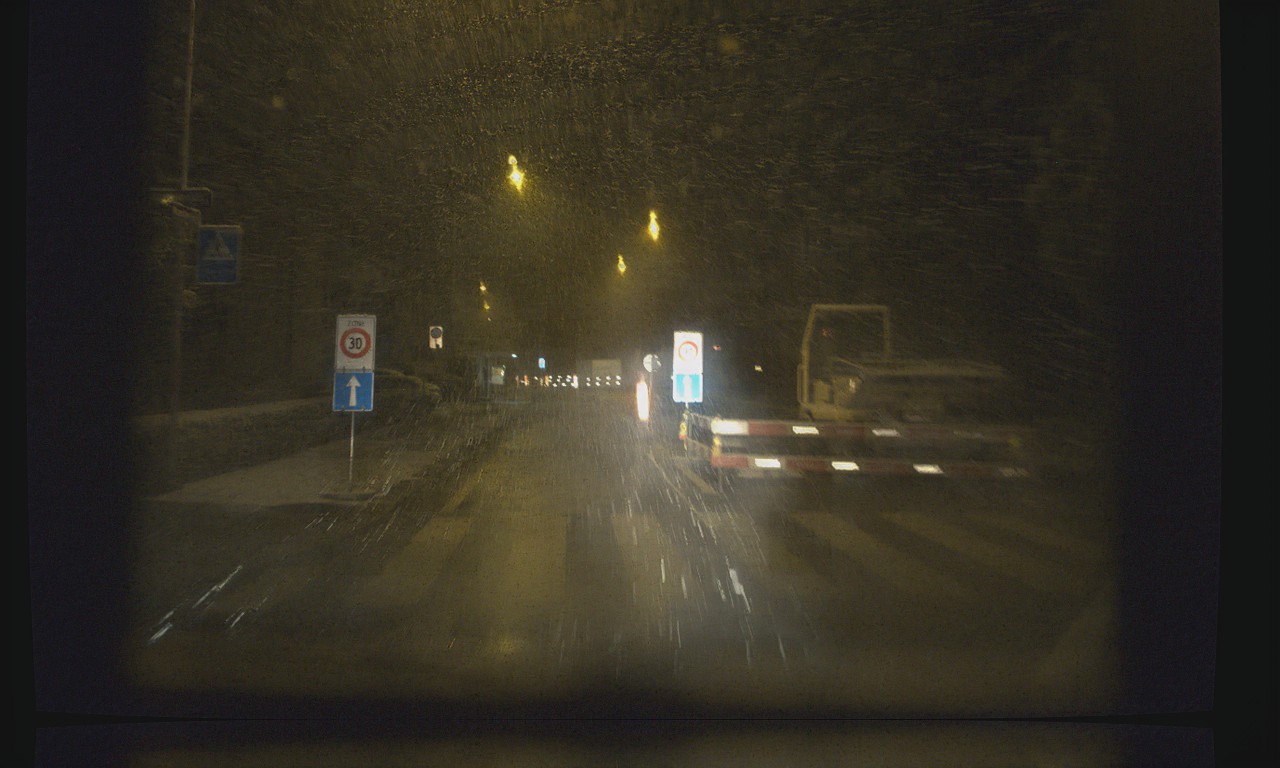} &
        \includegraphics[trim={12cm, 2.5cm, 6.5cm, 13.5cm},clip,width=\linewidth]{\basepath/14_02_05/ours_000068.jpg} \\

        \includegraphics[trim={5cm, 2.2cm, 6.5cm, 10.5cm},clip,width=\linewidth]{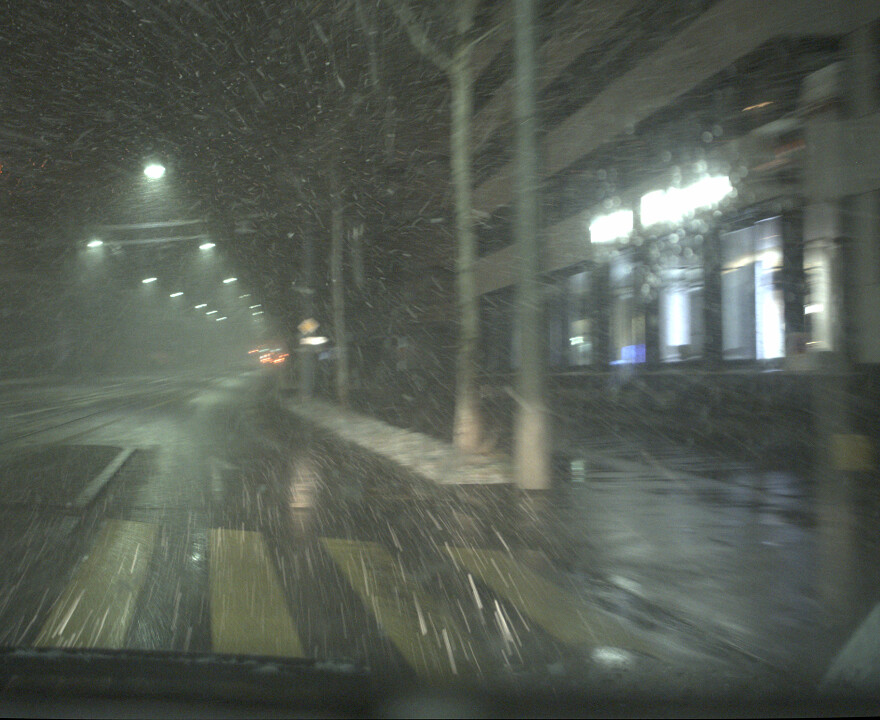} &
        \includegraphics[trim={12cm, 2.2cm, 13.5cm, 10.5cm},clip, width=\linewidth]{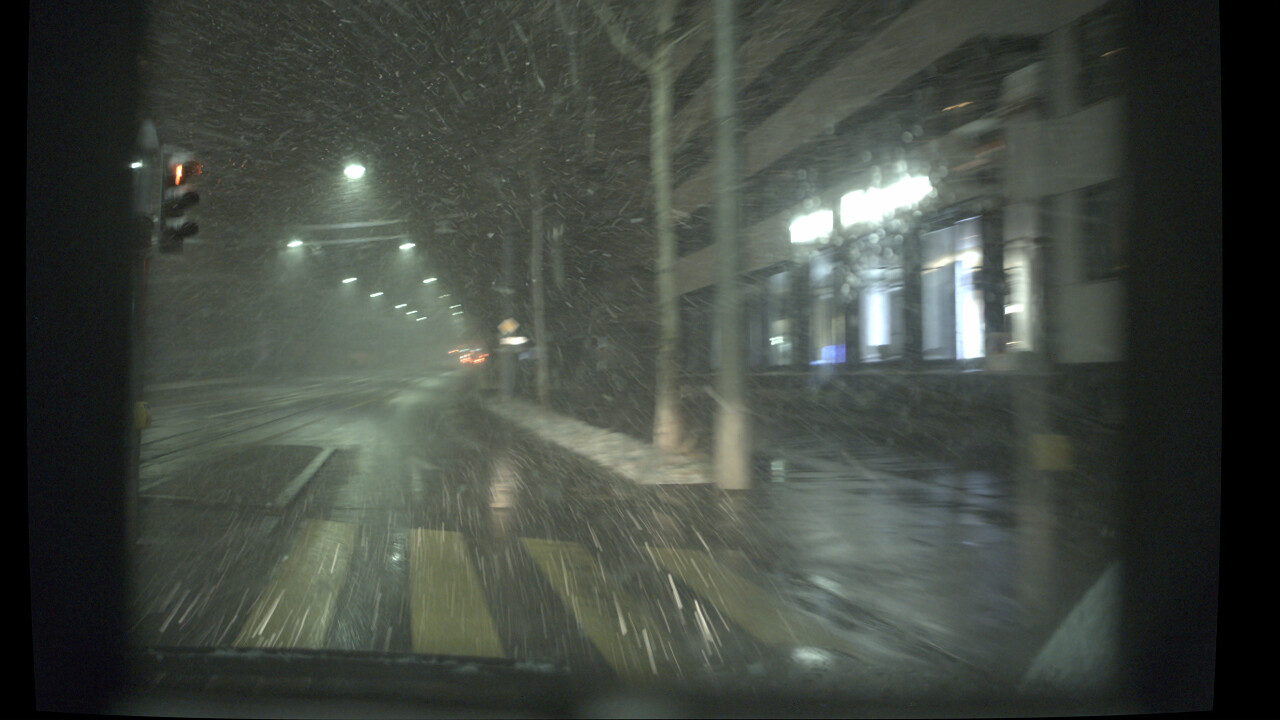} &
        \includegraphics[trim={12cm, 4cm, 13.5cm, 10.5cm},clip,width=\linewidth]{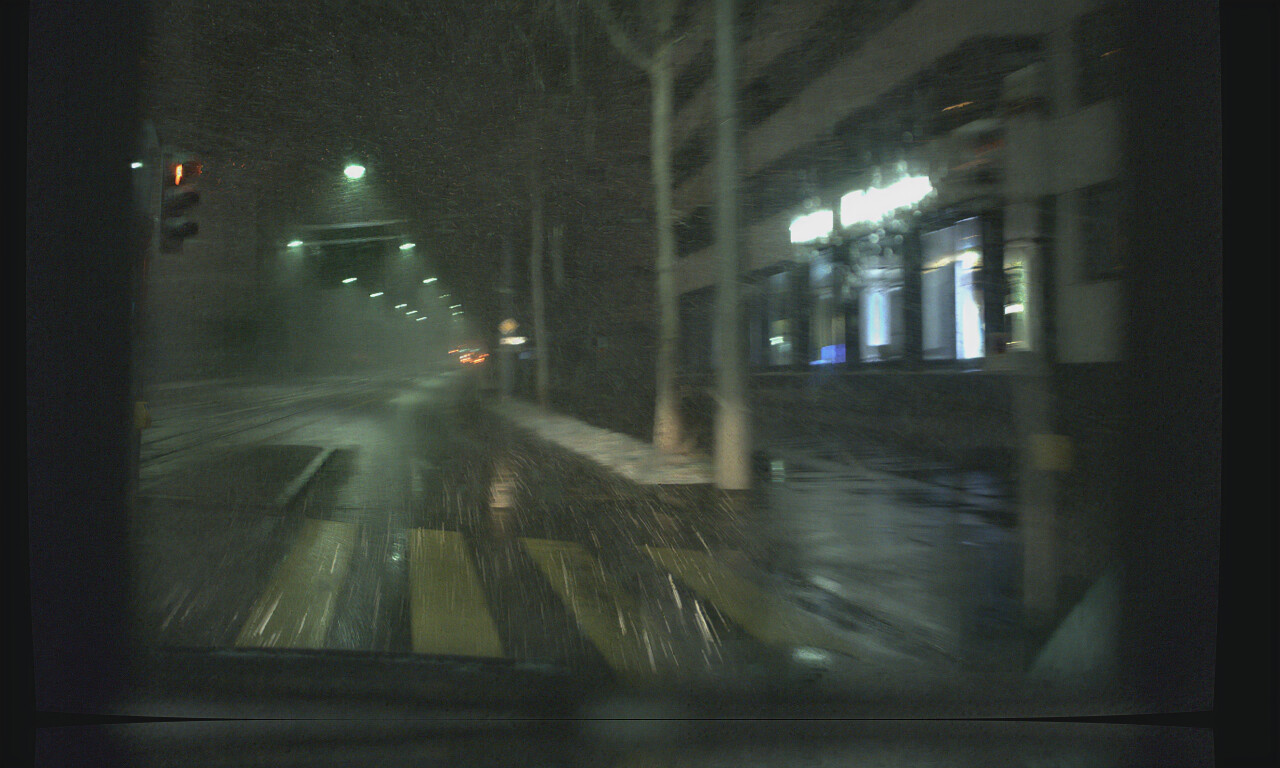} &
        \includegraphics[trim={5cm, 2.5cm, 6.5cm, 10.5cm},clip,width=\linewidth]{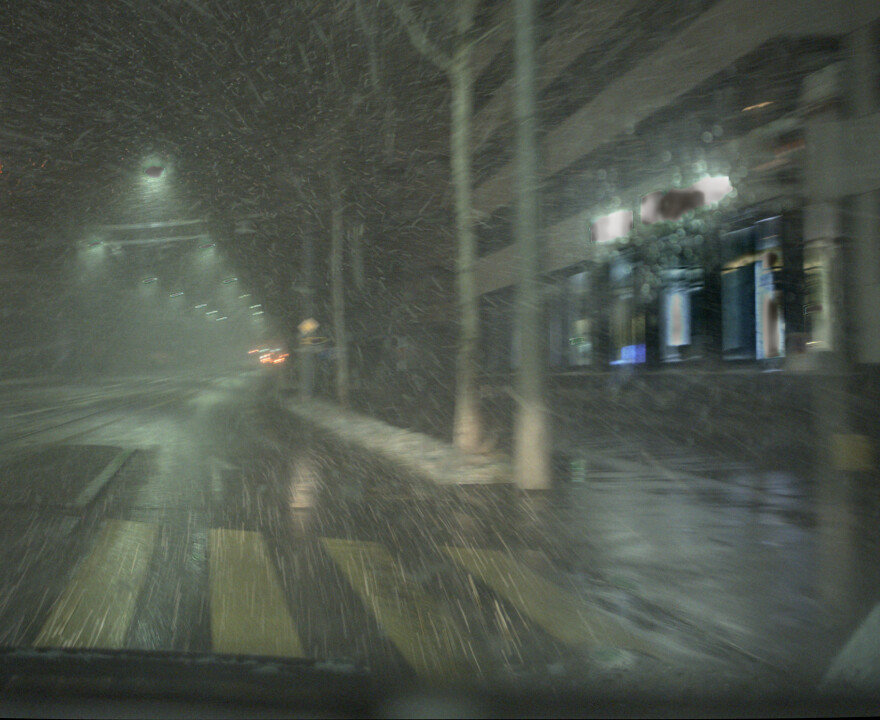} \\

        \includegraphics[trim={13cm, 2.2cm, 6.5cm, 14.5cm},clip,width=\linewidth]{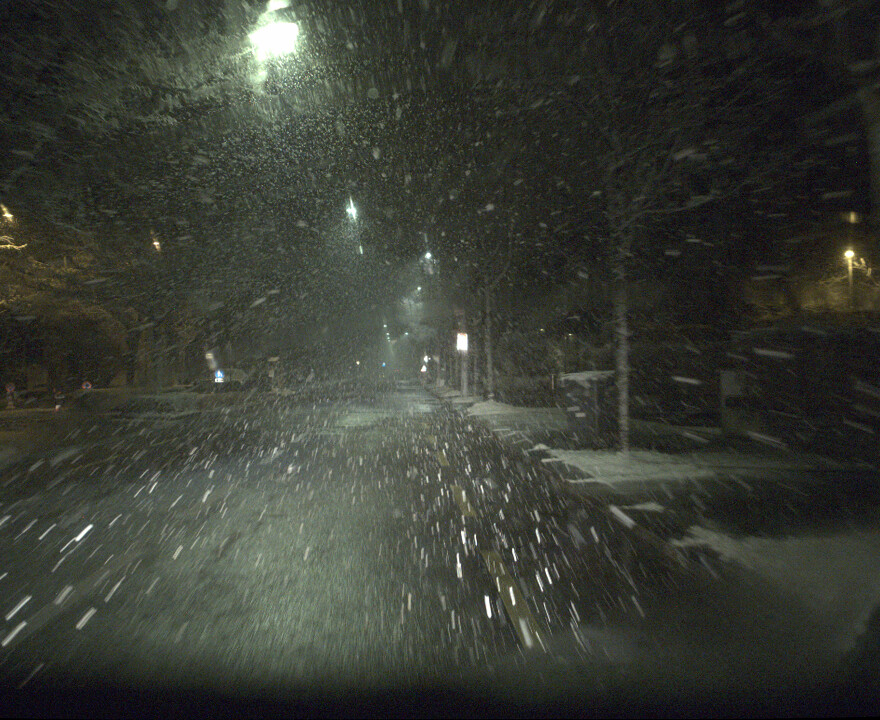} &
        \includegraphics[trim={20cm, 2.2cm, 13.5cm, 14.5cm},clip, width=\linewidth]{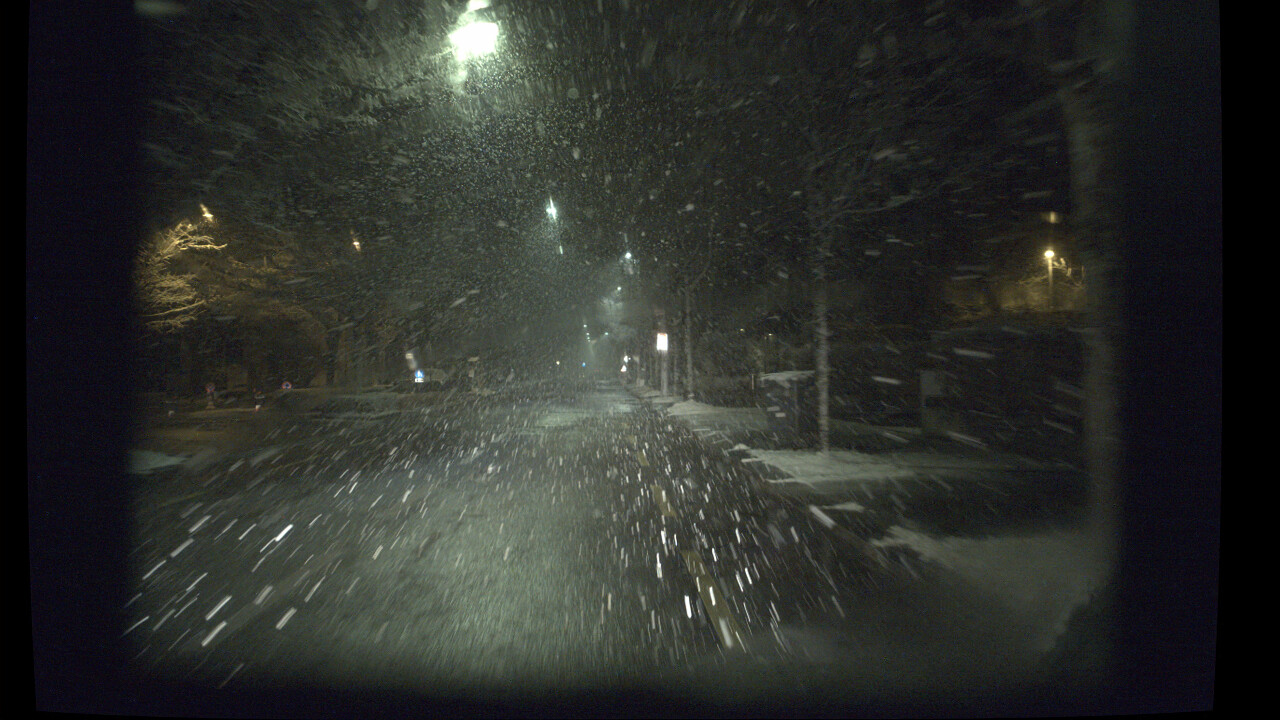} &
        \includegraphics[trim={20cm, 4cm, 13.5cm, 14.5cm},clip,width=\linewidth]{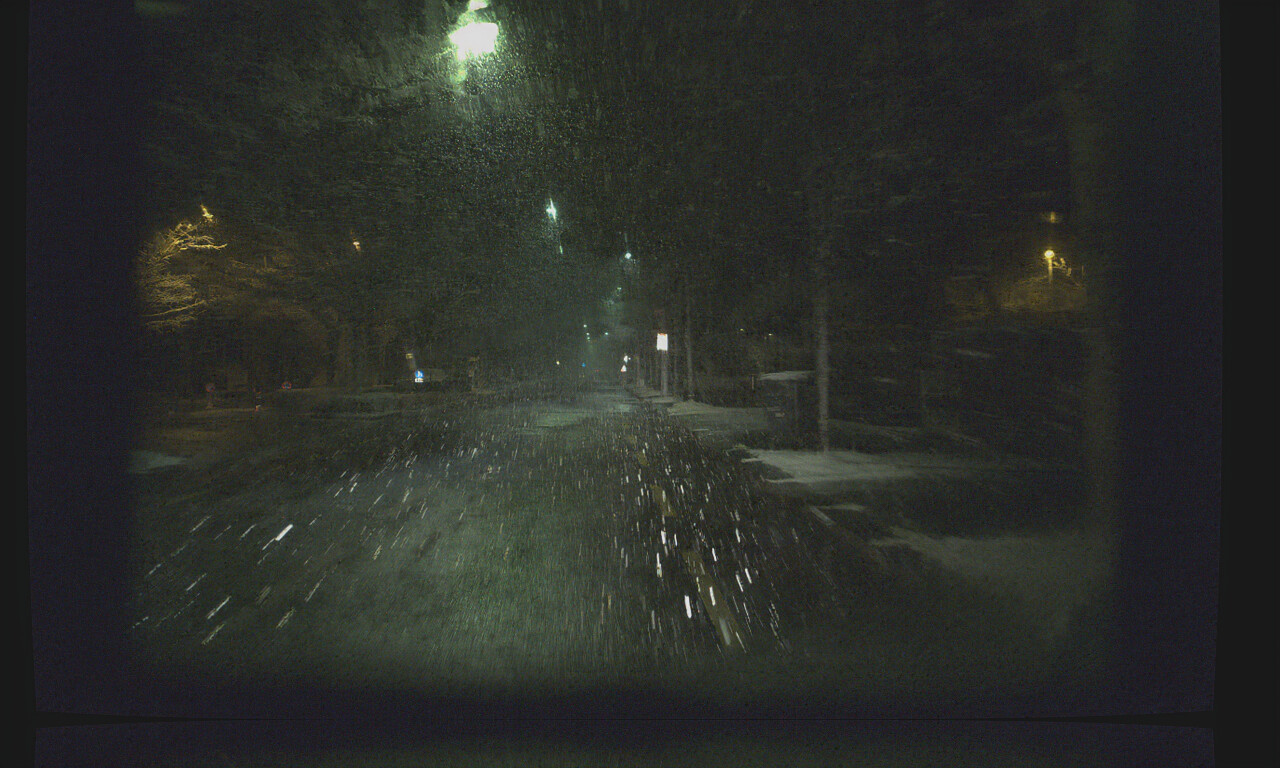} &
        \includegraphics[trim={13cm, 2.5cm, 6.5cm, 14.5cm},clip,width=\linewidth]{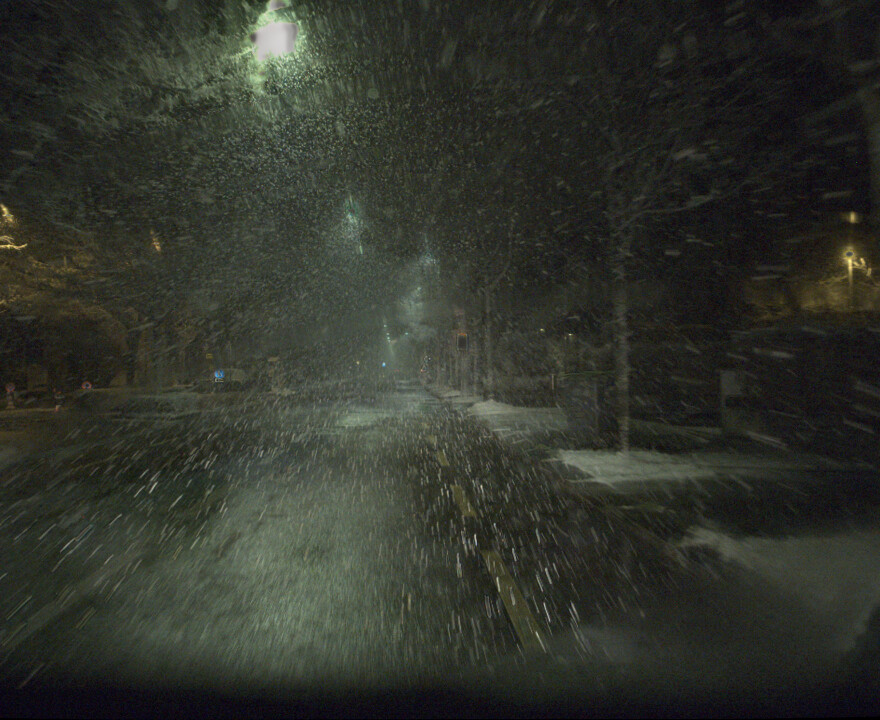} \\

        (a) Image & (b) Restormer \cite{Zamir21cvpr} & (c) Snowformer \cite{Chen22arxiv} & (d) \revtwo{\textbf{Ours}} \\
    \end{tabular}
    \caption{\revtwo{\textbf{Qualitative comparison on real driving scenes.}
    (a) Input images, (b)-(c) image restoration baselines, and (d) our results. 
    This shows a magnified region with dense snow and reflective surfaces, highlighting the ability of our method to reduce snow occlusion and preserve scene details under adverse snowfall conditions.}}
    \label{fig:result_realdrive}
    \vspace{-1em}
\end{figure*}

%% file: sections/06_conclusion.tex
\section{Discussion and limitations}
\label{sec:limitation}
\begin{figure}
    \centering
    \begin{tabular}{ccc}
        \includegraphics[width=0.3\linewidth]{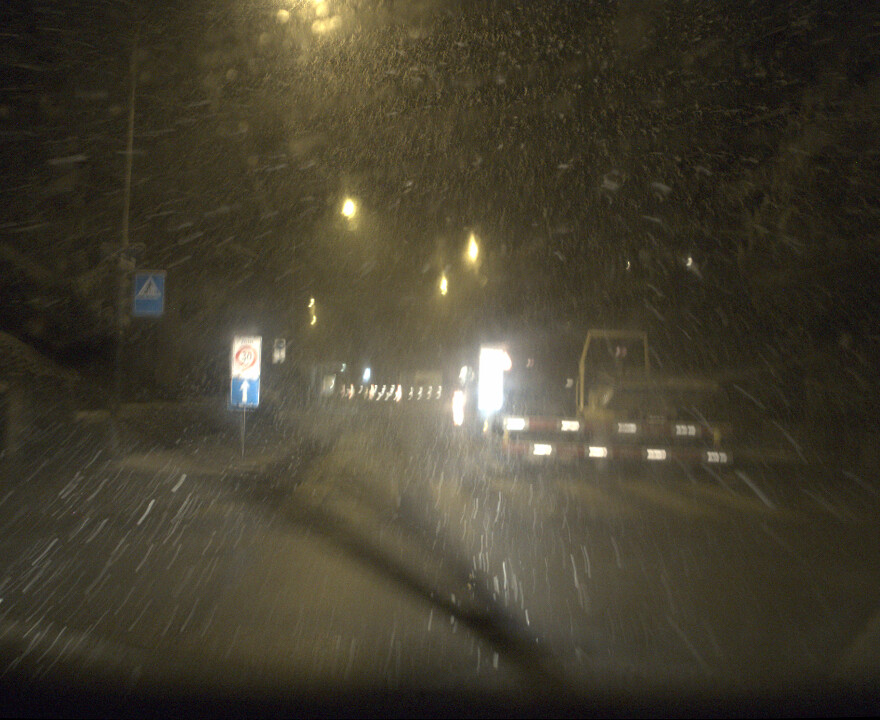} &
        \includegraphics[trim={4cm, 0cm, 4cm, 0cm},clip,width=0.3\linewidth]{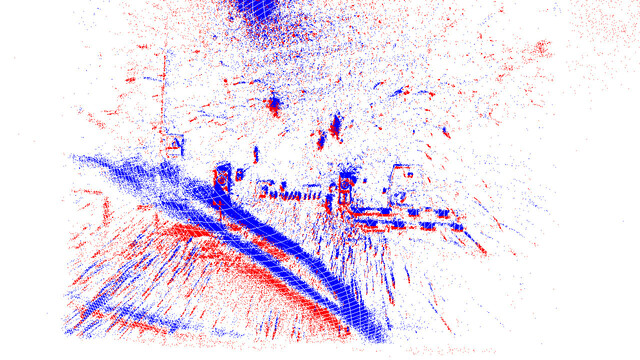} &
        \includegraphics[width=0.3\linewidth]{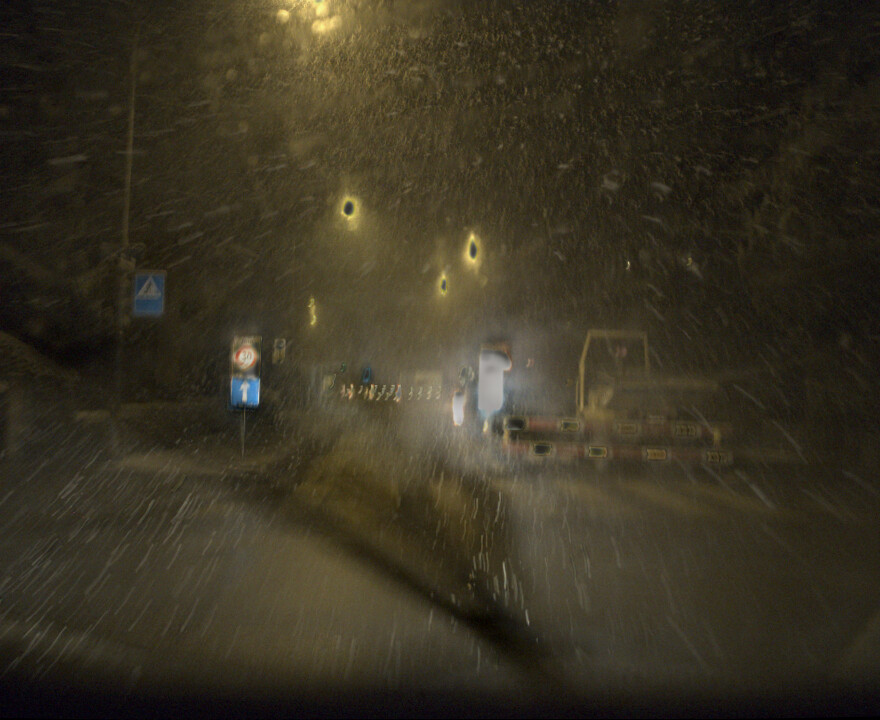} \\
        (a) Image & (b) Events & (c) Ours
    \end{tabular}
    \caption{\revtwo{\textbf{Limitations of the compositing approach}. The synthetic dataset generation method relies on the assumption that independently captured motions (e.g., background and occluder events) can be linearly combined without introducing artifacts. However, in real snowfall scenes, there are complex interactions such as flickering light sources from the street lamps, and car wipers which create dynamic occlusions that are not captured by our compositing approach.}}
    \label{fig:limitation}
    \vspace{-2em}
\end{figure}

While our proposed synthetic dataset provides a practical and physically grounded approach to simulating event-based occlusions, it relies on the assumption that independently captured motions, such as background and occluder (e.g., snowflake) events, can be linearly combined without introducing artifacts. 
This simplification does not account for inter-dependencies such as lighting interactions, occlusion ordering, or nonlinear sensor responses, which may be present in real-world scenes. 
\revtwo{For example, in \Fig~\ref{fig:limitation}, we show a real snowfall scene where the street lamps create flickering light sources and the car wipers create dynamic occlusions that are not captured by our compositing approach and lead to artifacts in the reconstructed images.
}

Another inherent limitation arises from the characteristics of event cameras themselves. 
Events are triggered only when a local brightness change exceeds a sensor-defined contrast threshold (typically around 15\%). 
As a result, regions with flat textures, uniform lighting, or minimal motion fail to produce meaningful event activity. 
This restricts the utility of our approach in low-texture environments or scenes where fine-grained intensity variations fall below the contrast threshold. 
Consequently, the reconstruction quality is strongly tied to the presence of high-frequency spatial and temporal information in the scene.

Despite these limitations, our approach offers a scalable and flexible alternative to traditional physics-based simulators, which often require complex modeling of event camera characteristics. 
Notably, the dataset generation methodology can be extended beyond weather simulation: any independently recordable motion (e.g., pedestrians, vehicles, or dynamic objects) can be composited over different backgrounds to augment data diversity in a controlled manner. 
This makes the method valuable for a wide range of applications such as robotics, autonomous driving, and dynamic scene understanding, where controlled yet realistic event-based data are scarce.
\vspace{-2em}
\section{Conclusion}
We introduce a novel event-based approach for background image reconstruction in the presence of dynamic occlusions.
It leverages the complementary nature of event cameras and frames to reconstruct true scene information instead of hallucinating occluded areas as done by image inpainting approaches.
Specifically, our proposed data-driven approach reconstructs the background image using only one occluded image and events.
The high temporal resolution of the events provides our method with additional information on the relative intensity changes between the foreground and background, making it robust to dense occlusions. %
To evaluate our approach, we present the first large-scale dataset recorded in the real world containing challenging scenes with synchronized events, occluded images, and ground-truth images.
Our method achieves an improvement of 3 dB in PSNR over state-of-the-art frame-based and event-based methods on both synthetic and real datasets.
We believe that our proposed method and dataset lay the foundation for future research.

%% file: sections/supp/supplementary.tex
\title{\MYTITLE\\---Supplementary Material---}
\maketitle

\input{sections/supp/datasets.tex}

\input{sections/supp/all_comparison.tex}

\revtwo{
\section{Artifacts and limitation analysis}
\label{sec:supp:artifacts}

In some cases, the reconstructed images may exhibit a degree of blurriness or checkerboard artifacts. 
Checkerboard artifacts arise from the inherent checkerboard pattern in the event stream, which is a consequence of the alignment from the event camera to the RGB camera. 
As shown in \Fig \ref{fig:artifact_checkerboard}, this checkerboard pattern is visible in the event stream.

In some cases, slight blurriness or artifacts can also appear in the reconstructed images. 
This effect is largely caused by depth-dependent alignment limitations in the DSEC dataset. 
In DSEC, events and frames are aligned using a global 2D homography. 
During camera motion in non-planar scenes, objects at different depths cannot be perfectly aligned simultaneously (see \Fig \ref{fig:depth_artifact}). 
As a result, certain scene elements may appear slightly misaligned between events and frames, which can lead to localized smoothing artifacts during reconstruction. 
For example, the degradation around the road marking “30” in is influenced by this depth-dependent misalignment.

Image reconstruction metrics such as PSNR often favor globally smooth reconstructions that reduce high-frequency differences with the reference image. 
Consequently, methods that slightly smooth textures can sometimes obtain higher PSNR values even when local sharpness is reduced. 
To better evaluate the impact of reconstruction quality on downstream perception tasks, we therefore also include motion-based metrics based on optical flow estimation. 
These metrics provide an indirect measure of how well the reconstructed images preserve motion-consistent scene structures.

As shown in \Tab \ref{tab:results:downstreamdepth}, our method achieves improved endpoint error (EPE) compared to competing approaches, indicating that the reconstructed images preserve motion information useful for downstream perception tasks. 
At the same time, the average pixel error (APE) remains within a very small margin ($<1\%$), suggesting that the remaining blur artifacts mainly affect fine pixel-level alignment rather than the global structural consistency of the scene.

These observations highlight an important direction for future work. 
In particular, improving event-frame alignment through depth-aware calibration or incorporating more physically grounded snowfall simulation could further reduce these artifacts and improve local texture preservation while maintaining robust snow suppression.

\begin{figure}[htp]
    \centering
    \vspace{-3mm}
    \begin{tabular}{cc}
        \includegraphics[width=0.45\linewidth]{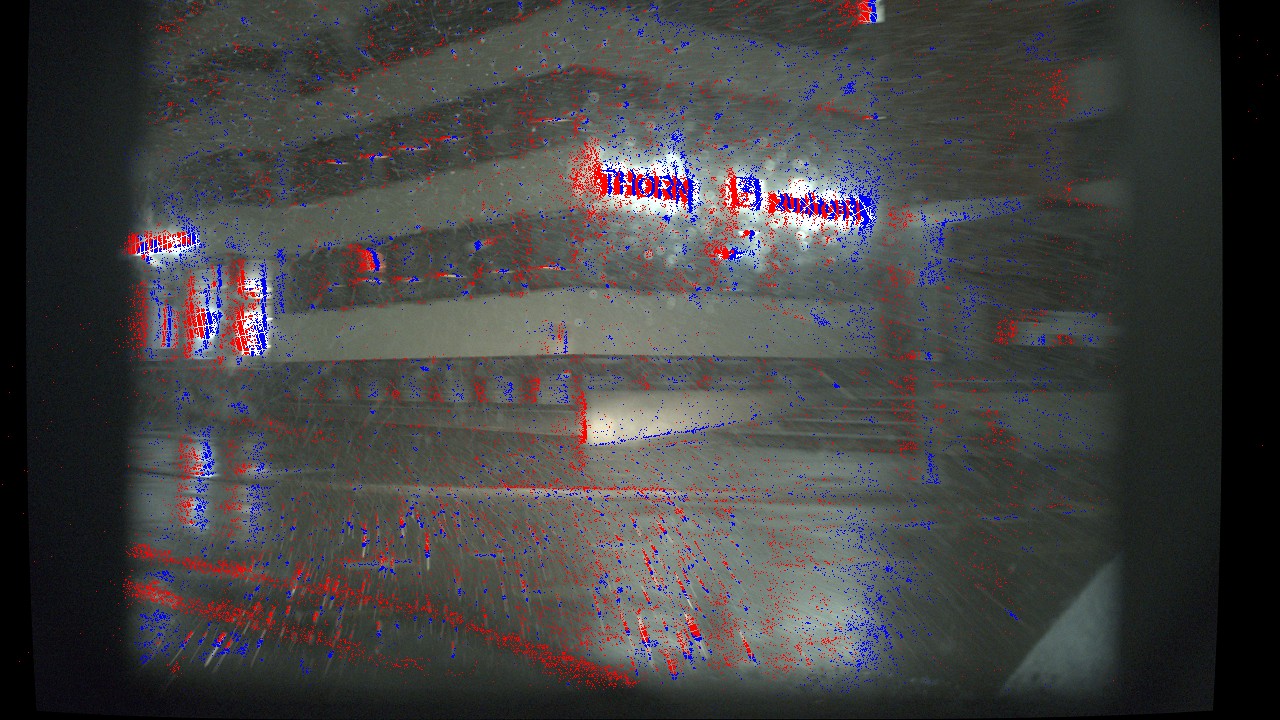}&
        \includegraphics[width=0.45\linewidth, trim=25cm 15cm 10cm 5cm, clip]{floats/images/overlay_result_realworld.jpg}\\
        \includegraphics[width=0.45\linewidth]{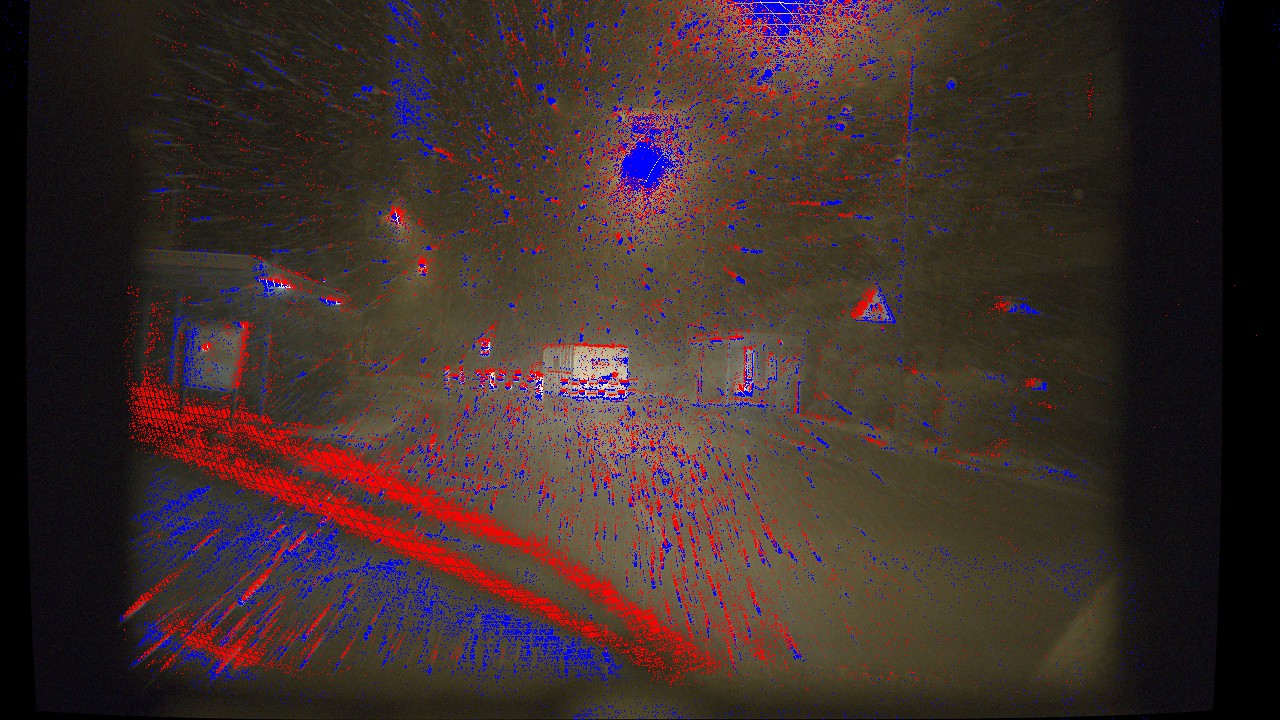}&
        \includegraphics[width=0.45\linewidth, trim=5cm 5cm 25cm 12cm, clip]{floats/images/overlay_result_realworld_2.jpg}\\
        \revtwo{(a) Overlay events on image} & \revtwo{(b) Zoomed in} \\
    \end{tabular}
    \caption{\revtwo{Checkerboard artifacts due to alignment:
    Overlaying events on groundtruth images from real-world dataset. 
    The alignment of the events to the RGB image results in a checkerboard pattern in the event stream, which is reflected in the reconstructed image.
    }}
    \label{fig:artifact_checkerboard}
\end{figure}

\begin{figure}[htp]
    \centering
    \begin{minipage}{0.45\textwidth}
        \centering
        \includegraphics[width=\linewidth]{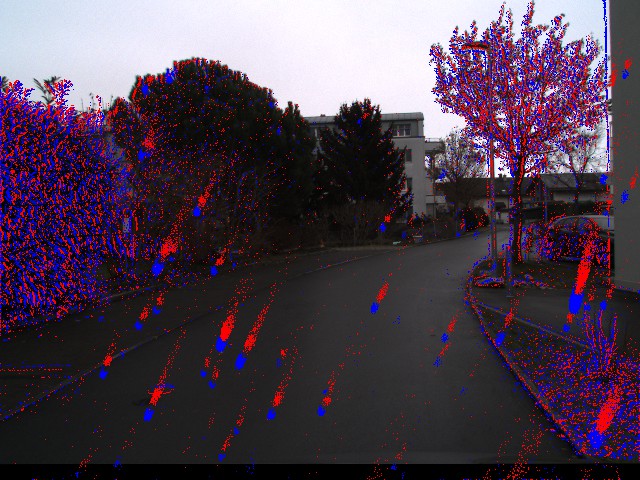}        \subcaption{\revtwo{Perfectly aligned background events and image}}
    \end{minipage}
    \hfill
    \begin{minipage}{0.45\textwidth}
        \centering
        \includegraphics[width=\linewidth]{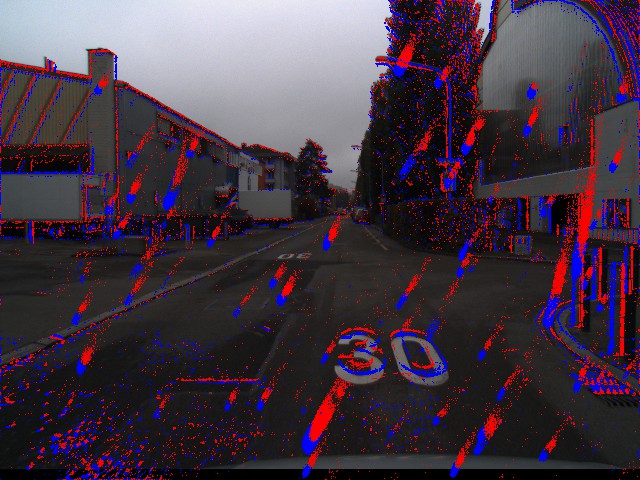}        \subcaption{\revtwo{Misaligned background events and image}}
    \end{minipage}
    \caption{\revtwo{Depth-dependent alignment limitations in DSEC dataset:
    Overlaying events on groundtruth images from \DSECSnow dataset. 
    The left image shows a scenario where the background events and image is perfectly aligned, while the right image shows a scenario where the road marking "30" is misaligned between events and images due to depth-dependent alignment limitations in the DSEC dataset.}}
    \label{fig:depth_artifact}
\end{figure}
}

\section{Additional Results}
\subsection{Generalization to other occlusions}
\label{sec:results:muses}
\input{floats/fig_muses.tex}
We also consider the all-weather driving dataset proposed in \cite{brodermann24eccv}, which contains synchronized and calibrated events, images, LiDAR, and RADAR measurements.
We only consider the sequences which have a mix of snow and rainfall.
Unfortunately, the snowfall is not dense enough in these sequences to hinder the view, unlike our dataset which was collected in heavy snowfall.
Nevertheless, we show the qualitative evaluation of our approach in \Fig \ref{fig:results_muses}.
By fusing temporal information from events, we are able to recover the bus in the background of the raindrop and building, which was occluded by a raindrop causing lens flare.

\noindent \textbf{Effect on downstream application}
We also evaluate the performance of our method on the downstream task of optical flow.
We show that our method is able to reconstruct the background scene with high accuracy, resulting in significant improvement in downstream applications, see \Fig \ref{fig:result_realsnow_downstream}.
We use the RAFT network \cite{teed20eccv} on the images reconstructed by all image restoration methods and compare the end-point-error (EPE) metric to evaluate the performance of optical flow.

\Tab \ref{tab:results:downstreamdepth} reports the End-Point Error (EPE) and accuracy metrics ($AE<1$, $AE<3$, $AE<5$) for optical flow computed on the \DAVISSnow dataset. 
Event-only (E2VID) and video-based (S2VD) baselines exhibit high EPE and low accuracy, reflecting their limited ability to recover motion information in the presence of severe occlusions. 
Image-based methods (Restormer, SnowFormer, RLP) provide improved performance, but still suffer from significant error, particularly under challenging conditions. 
Our approach, which fuses event and intensity modalities, achieves the lowest EPE.
The qualitative results in Fig.~\ref{fig:result_realsnow_downstream} further confirm these findings: optical flow maps generated using our method are visually closer to the ground truth, accurately capturing fine motion boundaries and overall scene structure, while baseline methods display artifacts and loss of detail. 
These results indicate that our event-image fusion approach not only improves desnowing quality, but also leads to substantial gains in the downstream tasks such as optical flow estimation.

\input{floats/fig_dsec_downstream.tex}

\input{floats/tables/tab_dsec_downstream.tex}

\begin{figure}[htp]
    \centering
    \vspace{-3mm}
    \begin{tabular}{cc}
        \includegraphics[trim={10cm, 2cm, 3cm, 10cm},clip,width=0.45\linewidth]{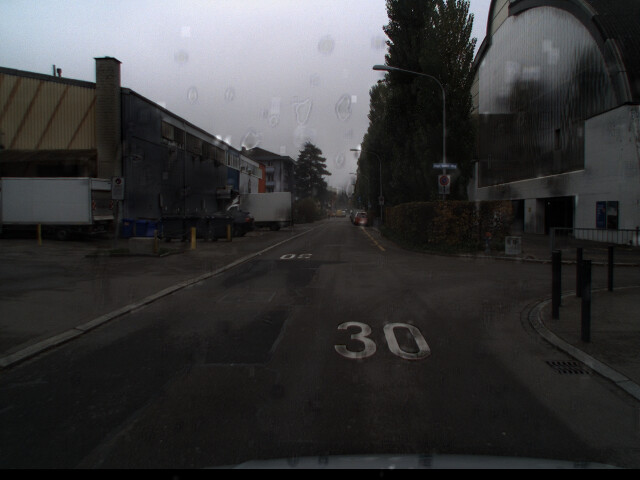} &
        \includegraphics[trim={10cm, 2cm, 3cm, 10cm},clip,width=0.45\linewidth]{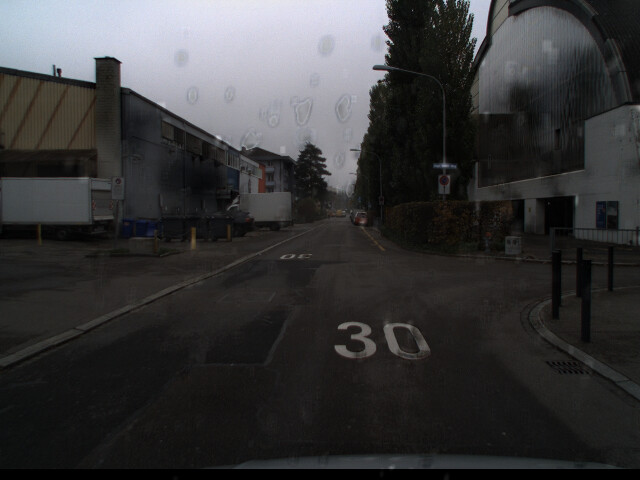} \\
        (a) Naive Fusion & (b) Ours \\
    \end{tabular}
    \caption{\revtwo{\textbf{Qualitative comparison of naive fusion and our proposed method}
    Our method effectively leverages the spatio-temporal features from events to reconstruct high-quality images, while the naive fusion approach results in several artifacts and loss of detail.}}
    \label{fig:ablation_network}
    \vspace{-6mm}
\end{figure}

\begin{figure}[htp]
    \centering
    \vspace{-6mm}
    \begin{tabular}{ccc}
        \includegraphics[trim={10cm, 2cm, 3cm, 10cm},clip,width=0.25\linewidth]{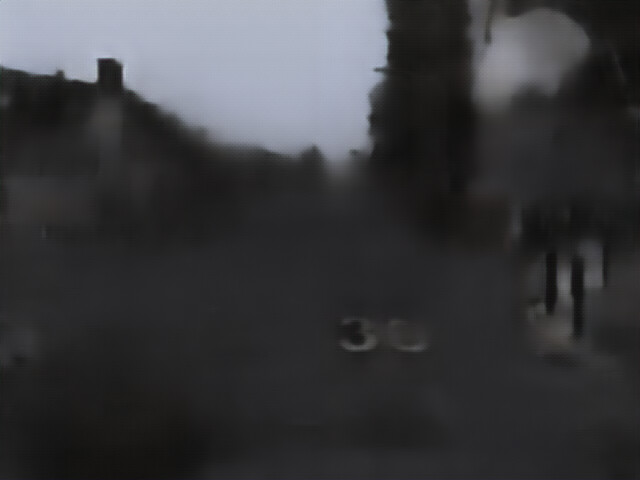} &
        \includegraphics[trim={10cm, 2cm, 3cm, 10cm},clip,width=0.25\linewidth]{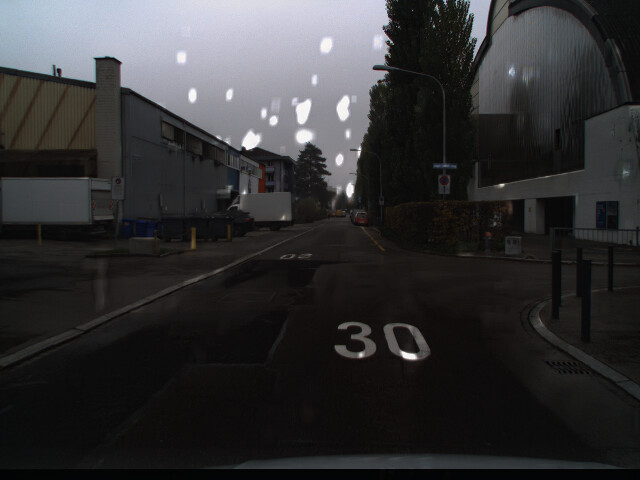} &
        \includegraphics[trim={10cm, 2cm, 3cm, 10cm},clip,width=0.25\linewidth]{floats/supplementary_figs/ablation_network/ours_000005.jpg} \\
        (a) Event Only & (b) Image Only & (c) Ours \\
    \end{tabular}
    \caption{\revtwo{\textbf{Qualitative comparison of event only, image only, and our proposed method}
    The event-only method lacks the low-frequency structural information present in image frames, while the image-only method struggles to recover motion cues in the presence of severe occlusions.
    Our method effectively fuses both modalities, resulting in better reconstruction quality.}}
    \label{fig:sensor_modality_ablation}
    \vspace{-3mm}
\end{figure}

\begin{table}[htp]
\centering
    \begin{tabular}{llll}
    \toprule
    \textbf{Model} &  & \textbf{PSNR (dB)} & \textbf{SSIM} \\
    \midrule
    Image-only & I    & 25.12          & 0.9240 \\
    Event-only & E    & 17.29          & 0.6864\\
    Naive Fusion & E + I    & 29.10        & 0.9519 \\
    Ours         & E + I   & \textbf{31.76}     & \textbf{0.9686}\\
    \bottomrule
    \end{tabular}
    \caption{\revtwo{Ablation study on effect of sensor modalities.}}
    \label{tab:ablation:network}
\end{table}

\begin{table}[htp]
    \centering
    \begin{tabular}{cc|ll}
    \toprule
    \multicolumn{2}{c|}{\textbf{Network Components}} & \textbf{PSNR (dB)} & \textbf{SSIM}  \\
    \cmidrule(lr){1-2}
    EventNet & Mask Prediction & & \\ 
    \midrule
    \texttimes & \texttimes & 29.10 & 0.9519 \\
    \texttimes & \checkmark & 29.45 & 0.9612 \\
    \checkmark & \texttimes & 31.09 & 0.9586 \\
    \checkmark & \checkmark & \textbf{31.76} & \textbf{0.9686} \\
    \bottomrule
    \end{tabular}

    \caption{\revtwo{Ablation study on network architecture.}}
    \label{tab:ablation:networkarchitecture}
    \vspace{-2em}
\end{table}

\rev{\noindent \textbf{Ablation on network architecture}
\label{sec:results:ablation}
We also compare the contribution of each sensor modality in \Tab \ref{tab:ablation:network} in first two rows.
The image-only method is same as Snowformer \cite{Chen22arxiv}, while the event-only method uses the same architecture as proposed method but discards the image input.
While events provide valuable motion cues, they lack the low-frequency structural information present in image frames, explaining the lower performance of the event-only pipeline compared to the image-only baseline.
\revtwo{\Fig \ref{fig:sensor_modality_ablation} shows a qualitative comparison of the three approaches, where the event-only method fails to reconstruct the overall structure of the scene, while the image-only method struggles to remove occlusions, resulting in artifacts and loss of detail.}
Fusion of both modalities leverages the strengths of each sensor, resulting in better reconstruction quality as evidenced by the improvement in PSNR and SSIM.
It can be seen that even with a naive fusion approach, we achieve higher PSNR and SSIM than image-only methods.
However, our proposed method outperforms the naive fusion approach by a significant margin, demonstrating the effectiveness of our architecture.

In addition, we show how different components of our network architecture contribute to the overall performance in \Tab \ref{tab:ablation:networkarchitecture}.
The naive fusion strategy performs a simple concatenation of the image and event inputs, which is then directly passed to the image reconstruction module. 
This approach lacks any form of temporal modeling, as it does not incorporate recurrence or sequential reasoning over the event stream.
Incorporating mask prediction from events allows the network to adaptively combine the input image and the network's prediction, leading to improved reconstruction quality. 
However, the most significant performance gain comes from EventNet, which extracts salient spatio-temporal features from the event stream before fusion.
\revtwo{Qualitatively, as shown in the \Fig \ref{fig:ablation_network} the naive fusion approach results in several artifacts and loss of detail, while our method effectively leverages the spatio-temporal features from events to reconstruct high-quality images.}

}

%% file: sections/supp/datasets.tex
\section{\DSECSnow Dataset}
We describe the setup used here to record our datasets.
In this section, we provide a detailed overview of the data generation process that was used to create the \DSECSnow dataset.
We used a black background to maximize the contrast between the foreground and background as shown in the experimental setup.
The underlying assumption in this setup is that snow particles tend to be brighter than most objects in the surroundings.
Therefore, to maximize the contrast between every snowflake movement and the background, we use a black screen.
These events can then be pruned to account for an arbitrary background.

The second assumption which we make is that snowflakes will always be in the foreground, i.e. they will never be occluded by the background, which for most common scenarios is a reasonable assumption.
Of course, this also means we do not model any depth perception in this fusion and treat foreground events as far away from the background.

Lastly, as of now, we do not model the motion of the snow particles according to ego-motion of the camera.
In typical driving scenarios, the ego-motion of the car makes the snowflake appear to come towards the camera rather than simply flying down.
This of course is a function of the speed of the camera and makes this matter of accurately simulating the motion of snowflakes quite challenging.
We therefore simplify this problem and only apply a homography transformation to the foreground events.
Overall, our dataset consists of around $200$ training and $50$ test sequences, with each sequence consisting of a short duration of driving with snowfall during both day and night.

\textbf{Augmentation Parameters}
We describe in detail the augmentation parameters used to generate the dataset.
Similar to the method proposed in \cite{Chen23ICCV}, we render different effects of snowfall on the image such as haze and illumination-dependent snow appearance.
To render haze, we follow the model proposed in \cite{li18tip} which uses the atmospheric scattering model for rendering hazy images.
To blend the snow foreground with the background image, we use the strategy proposed in \cite{Chen23ICCV}, by considering the ambient illumination and time of day to blend the snowflakes.
For example, during the day, snowflakes blend with the sky and are therefore not easily visible with the brighter sky background.
This is exactly the opposite during the night: snowflakes are more visible in brightly lit areas such as headlamps or streetlights \cite{Chen23ICCV}.
These hazy images are combined with the foreground snow events to produce the final image.
To simulate realistic snow occlusion, we apply different augmentations to the snow events.
As described in \Eq. 9, we apply different augmentations to the snow:\begin{itemize}
    \item Snow Speed: The speed of the snowflakes is artificially controlled by scaling the timestamp of the foreground events.
    \item Snow Density: The density of snowflakes is increased by overlaying multiple snow events by staggering the timestamps and applying a homography transformation to the foreground events.
    \item Motion Direction: Motion direction is only controlled by flipping the foreground events along the horizontal axis.
\end{itemize}

\textbf{Simulating Events}
 Background events correspond to driving sequences recorded using DSEC \cite{MGehrig21ral}.
It consists of an event camera and an RGB camera mounted next to each other on a car while driving in different cities in Switzerland.
The background events ($E_{haze}(x)$) are recorded with a Prophesee Gen3.1 event camera with a resolution of $640 \times 480$ pixels.
The background images ($J(x)$) are recorded at $20 Hz$ with a resolution of $1440 \times 1080$ pixels and are aligned with the events.
Since our approach relies on the temporal and spatial alignment of the events and images, we use the rectified and aligned events and images from the DSEC dataset.

\begin{figure}[!ht]
    \centering
    \includegraphics[trim={2cm, 0cm, 10cm, 2cm},clip,width=\linewidth]{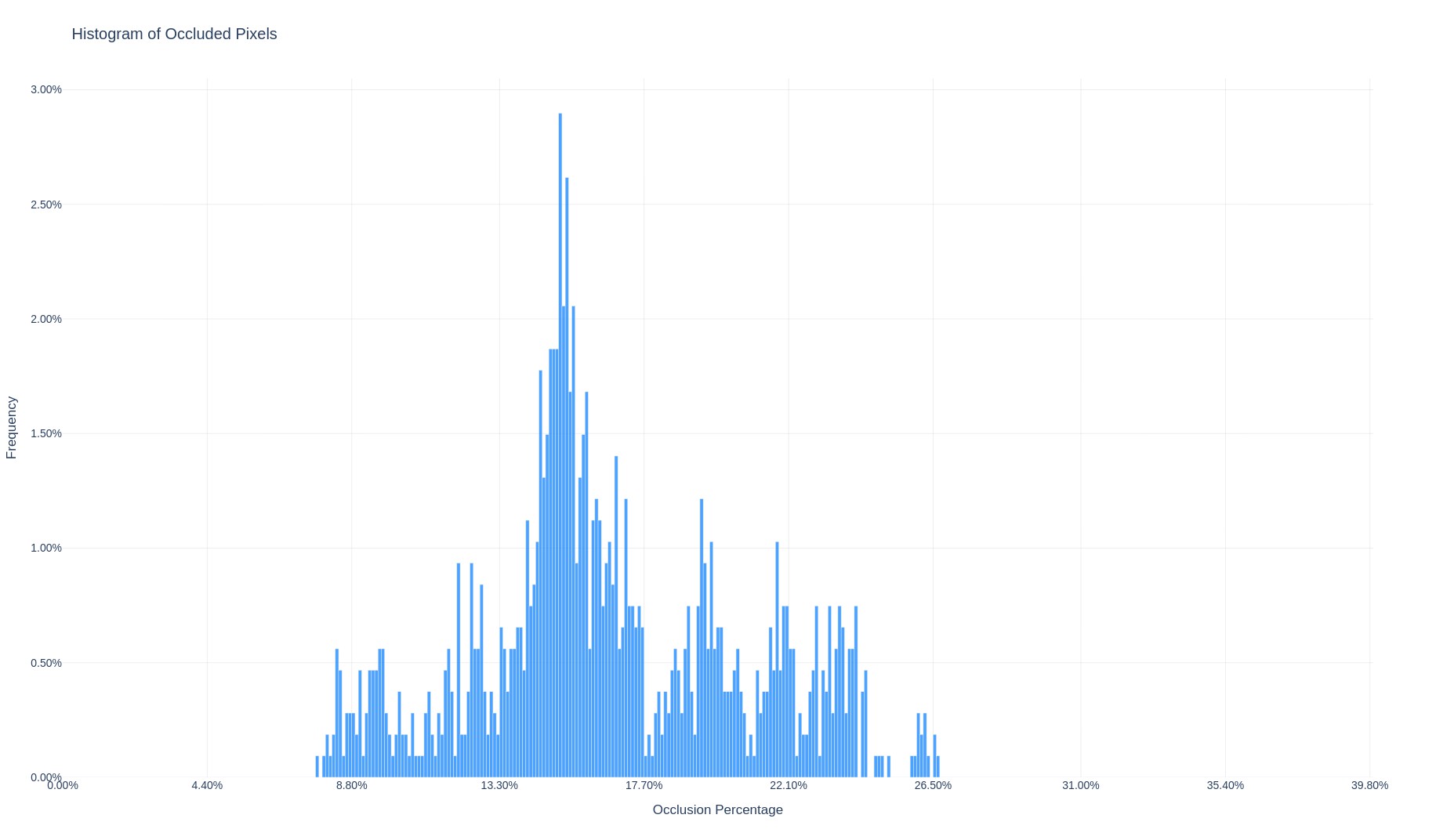}    \caption{Dataset statistics of \DSECSnow dataset. The histogram shows the percentage of occluded pixels in the dataset, illustrating the distribution of occlusion intensity across the dataset.}
    \label{fig:supp:dataset}
\end{figure}

The foreground events ($E_{snow}(x)$) are recorded using a Prophesee Gen4 event camera with a resolution of $1280 \times 720$ pixels.
The foreground events are recorded in front of a black background.
It was shown in \cite{Muglikar25PAMI}, that the signal-to-noise ratio (SNR) of an event camera depends on the illumination and contrast of the scene.
As illumination increases, the SNR of the events can drop significantly if the contrast is not high enough.
As we were restricted to outdoor recording, we could not control the illumination of the scene, so we use a black background to maximize the contrast between the foreground (snow, typically bright) and the background, ensuring sufficient SNR when recording the foreground events.
Dataset statistics are shown in \Fig~\ref{fig:supp:dataset}.
The dataset consists of $1000$ training and $470$ test pairs of images and events.
In addition, we provide ground truth images and events for both the training and test sequences. 
A histogram illustrating the percentage of occluded pixels reveals the distribution of occlusion intensity across the dataset.
Most images have occlusion levels concentrated between approximately 13\% and 22\%, indicating that moderate occlusion intensity is prevalent.
Fewer instances exhibit extreme occlusion levels, highlighting the realistic variability in weather conditions captured by the dataset.

%% file: sections/supp/all_comparison.tex
\subsection{Comparison with State-of-the-Art Methods}
\begin{figure*}
    \centering
    \begin{tabular}{ccc}
        \includegraphics[width=0.3\linewidth]{floats/supplementary_figs/dsec_snow/allexamples/dsec_snow_32854dcc-f592-45a7-bb79-7fcc57f66c7d/input_000003.jpg}&
        \includegraphics[width=0.3\linewidth]{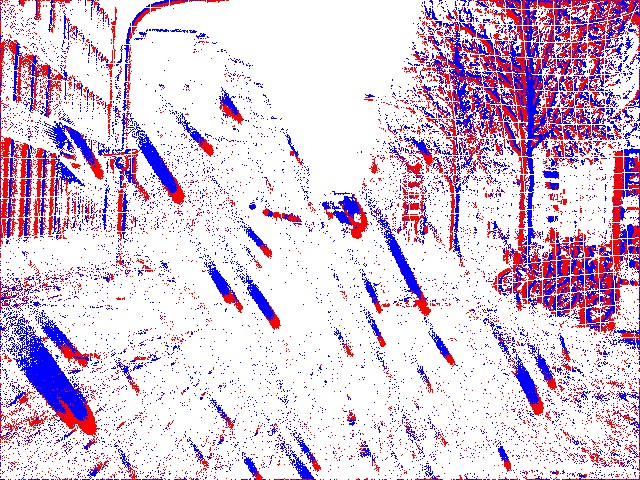}&
        \includegraphics[width=0.3\linewidth]{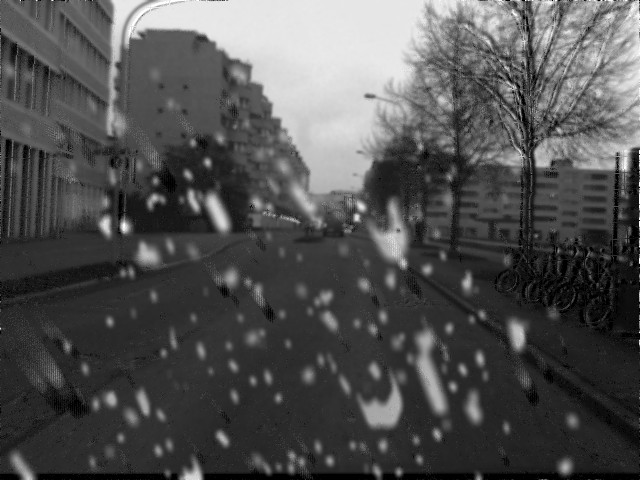}\\
        (a) Image & (b) Events  & (c) Model-based \\
        \includegraphics[width=0.3\linewidth]{floats/supplementary_figs/dsec_snow/allexamples/dsec_snow_32854dcc-f592-45a7-bb79-7fcc57f66c7d/restormer_000003.jpg}&
        \includegraphics[width=0.3\linewidth]{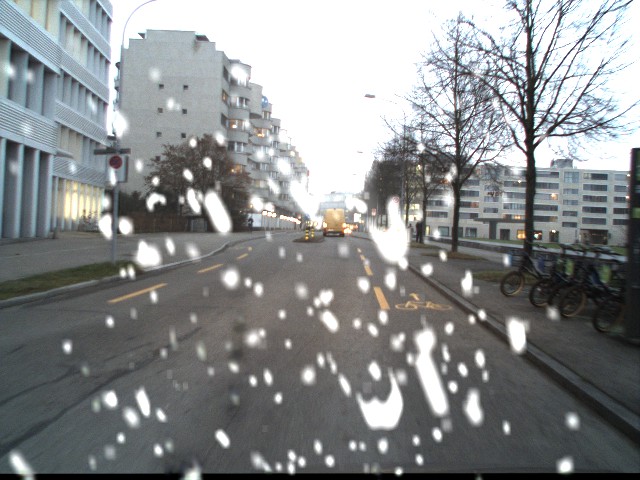}&
        \includegraphics[width=0.3\linewidth]{floats/supplementary_figs/dsec_snow/allexamples/dsec_snow_32854dcc-f592-45a7-bb79-7fcc57f66c7d/snowformer_000003.jpg}\\
        (d) Restormer \cite{Zamir21cvpr} & (e) RLP \cite{Zhang23ICCV} & (f) SnowFormer \cite{Chen22arxiv} \\
        \includegraphics[width=0.3\linewidth]{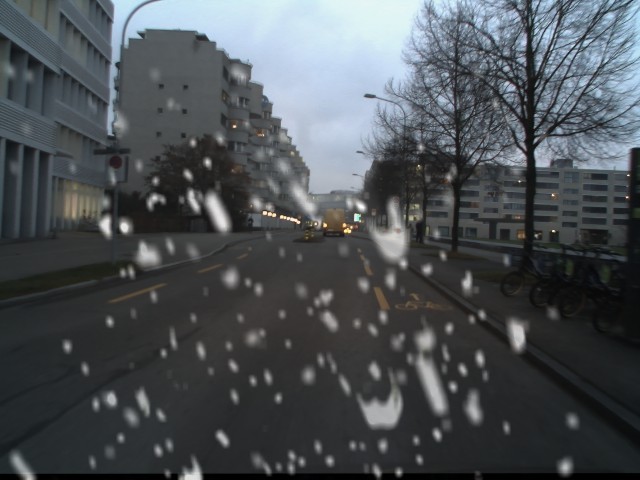}&
        \includegraphics[width=0.3\linewidth]{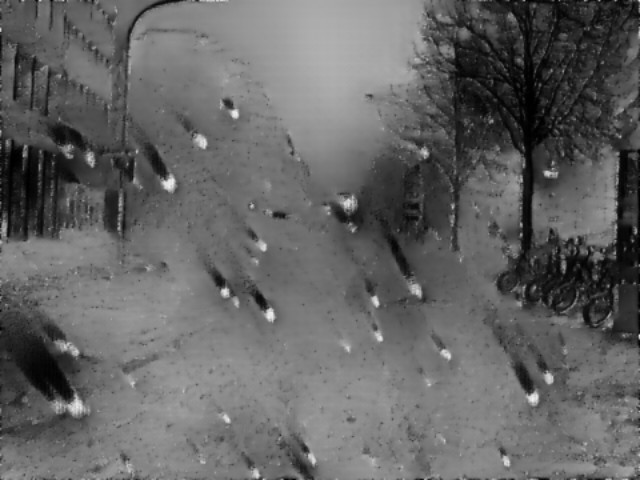}&
        \includegraphics[width=0.3\linewidth]{floats/supplementary_figs/dsec_snow/allexamples/dsec_snow_32854dcc-f592-45a7-bb79-7fcc57f66c7d/v11_nf_daylight_000003.jpg}\\
        (g) S2VD \cite{Yue21cvpr} & (h) E2VID \cite{Rebecq19pami} & (i) Ours \\
    \end{tabular}
    \caption{\textbf{Comparison of our method with state-of-the-art desnowing methods on \DSECSnow dataset.}
    }
    \label{fig:supp:full_dsec_snow}
\end{figure*}

We show samples from our method and compare it with state-of-the-art desnowing methods on the \DSECSnow dataset in Fig. \ref{fig:supp:full_dsec_snow}.

\begin{figure*}
    \centering
    \begin{tabular}{ccc}
        \includegraphics[trim=0.5cm 0.2cm 0.6cm 0cm, clip, width=0.3\linewidth]{floats/supplementary_figs/dsec_snow/allexamples/record9_138/image000138.jpg}&
        \includegraphics[trim=0.5cm 0.2cm 0.6cm 0cm, clip, width=0.3\linewidth]{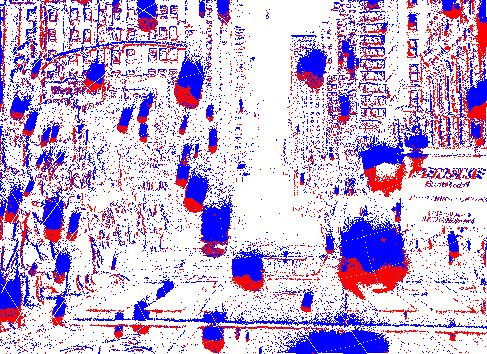}&
        \includegraphics[trim=0.5cm 0.2cm 0.6cm 0cm, clip, width=0.3\linewidth]{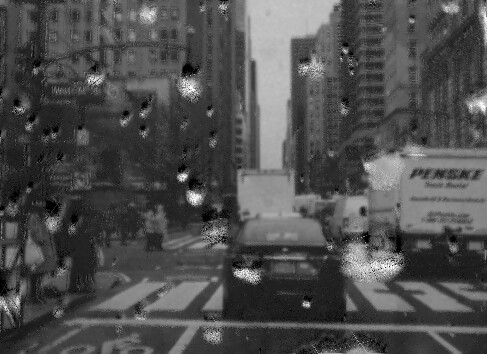}\\
        (a) Image & (b) Events  & (c) Model-based \\
        \includegraphics[trim=0.5cm 0.2cm 0.6cm 0cm, clip, width=0.3\linewidth]{floats/supplementary_figs/dsec_snow/allexamples/record9_138/restormer000138.jpg}&
        \includegraphics[trim=0.5cm 0.2cm 0.6cm 0cm, clip, width=0.3\linewidth]{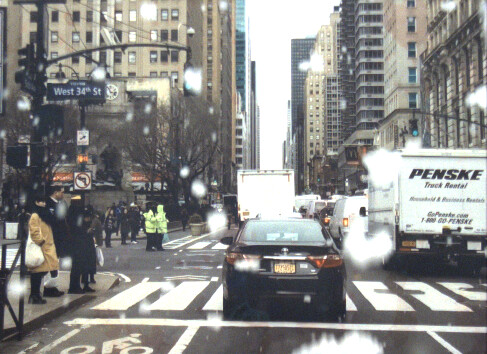}&
        \includegraphics[trim=0.5cm 0.2cm 0.6cm 0cm, clip, width=0.3\linewidth]{floats/supplementary_figs/dsec_snow/allexamples/record9_138/snowformer000138.jpg}\\
        (d) Restormer \cite{Zamir21cvpr} & (e) RLP \cite{Zhang23ICCV} & (f) SnowFormer \cite{Chen22arxiv} \\
        \includegraphics[trim=0.5cm 0.2cm 0.6cm 0cm, clip, width=0.3\linewidth]{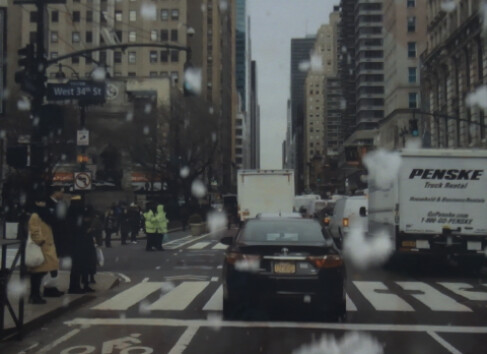}&
        \includegraphics[trim=0.5cm 0.2cm 0.6cm 0cm, clip, width=0.3\linewidth]{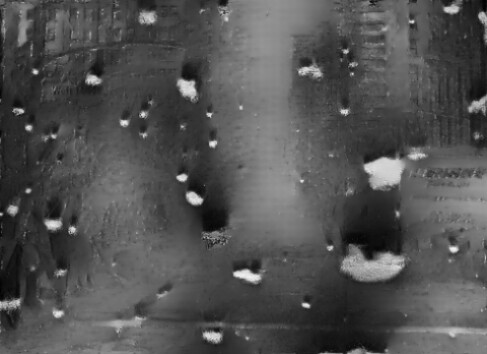}&
        \includegraphics[trim=0cm 0.2cm 0cm 0cm, clip,width=0.3\linewidth]{floats/supplementary_figs/dsec_snow/allexamples/record9_138/ours000138.jpg}\\
        (g) S2VD \cite{Yue21cvpr} & (h) E2VID \cite{Rebecq19pami} & (i) Ours \\
    \end{tabular}
    \caption{\textbf{Comparison of our method with state-of-the-art desnowing methods on \DAVISSnow dataset.}
    }
    \label{fig:supp:full_slider_snow}
    
\end{figure*}

\begin{figure*}
    \centering
    \begin{tabular}{ccc}
        \includegraphics[trim=0.5cm 0cm 0.4cm 0cm, clip, width=0.3\linewidth]{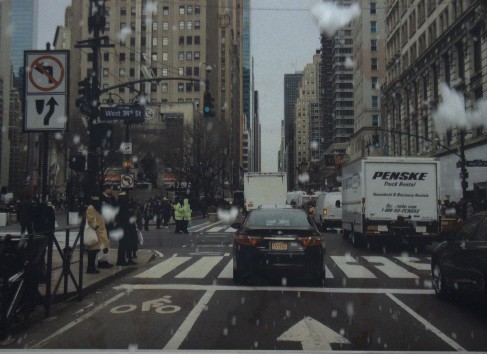}&
        \includegraphics[trim=0.5cm 0cm 0.4cm 0cm, clip, width=0.3\linewidth]{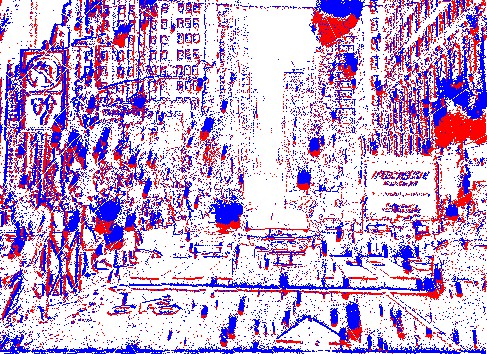}&
        \includegraphics[trim=0.5cm 0cm 0.4cm 0cm, clip, width=0.3\linewidth]{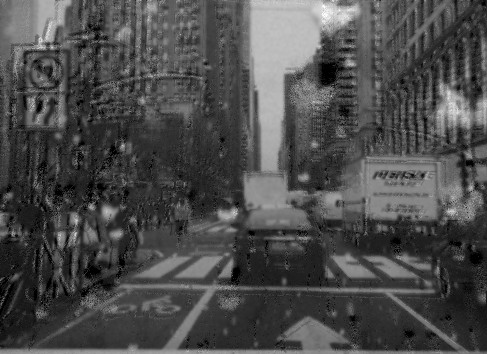}\\
        (a) Image & (b) Events  & (c) Model-based \\
        \includegraphics[trim=0.5cm 0cm 0.4cm 0cm, clip, width=0.3\linewidth]{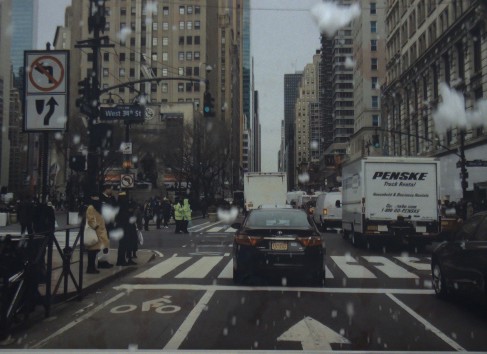}&
        \includegraphics[trim=0.5cm 0cm 0.4cm 0cm, clip, width=0.3\linewidth]{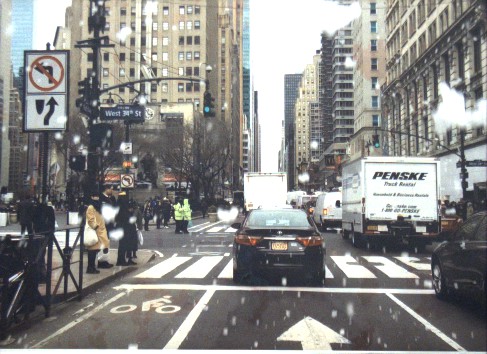}&
        \includegraphics[trim=0.5cm 0cm 0.4cm 0cm, clip, width=0.3\linewidth]{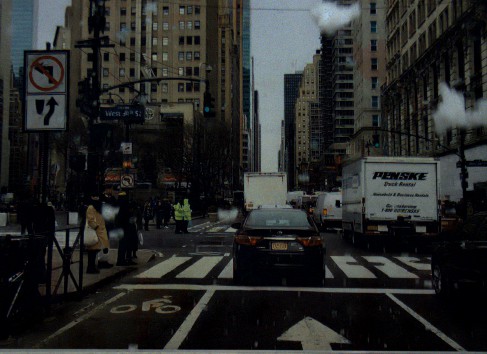}\\
        (d) Restormer \cite{Zamir21cvpr} & (e) RLP \cite{Zhang23ICCV} & (f) SnowFormer \cite{Chen22arxiv} \\
        \includegraphics[trim=0.5cm 0cm 0.4cm 0cm, clip, width=0.3\linewidth]{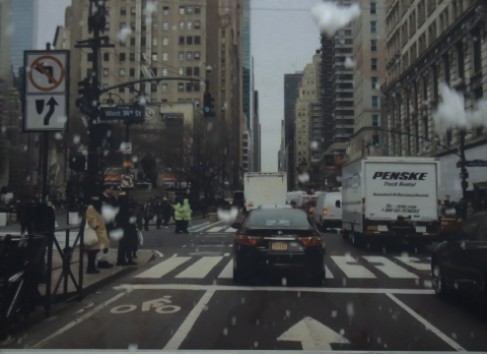}&
        \includegraphics[trim=0.5cm 0cm 0.4cm 0cm, clip, width=0.3\linewidth]{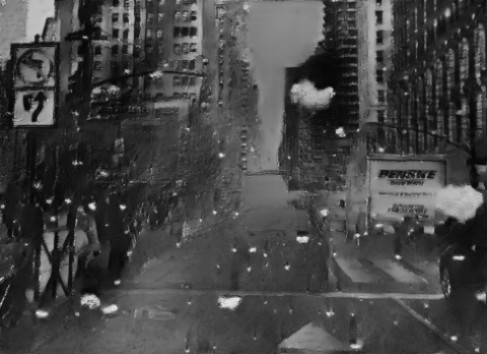}&
        \includegraphics[trim=0cm 0.2cm 0cm 0cm, clip,width=0.3\linewidth]{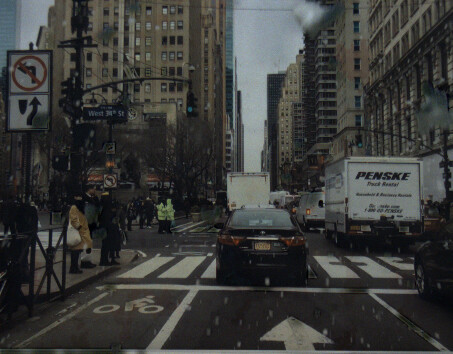}\\
        (g) S2VD \cite{Yue21cvpr} & (h) E2VID \cite{Rebecq19pami} & (i) Ours \\
    \end{tabular}
    \caption{\textbf{Comparison of our method with state-of-the-art desnowing methods on \DAVISSnow dataset.}
    }
    \label{fig:supp:full_slider_snow_2}
    
\end{figure*}

We show samples from our method and compare it with state-of-the-art desnowing methods on the \DAVISSnow dataset in \Fig ~\ref{fig:supp:full_slider_snow} and \Fig~\ref{fig:supp:full_slider_snow_2}.

\begin{figure*}
    \centering
    \begin{tabular}{cc}
        \includegraphics[width=0.45\linewidth]{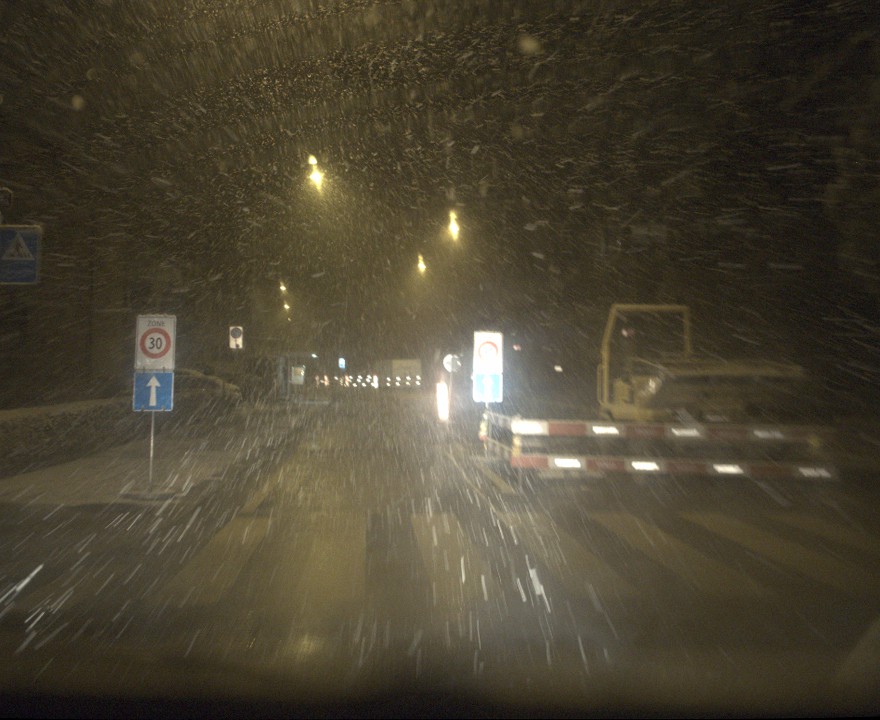}&
        \includegraphics[width=0.45\linewidth]{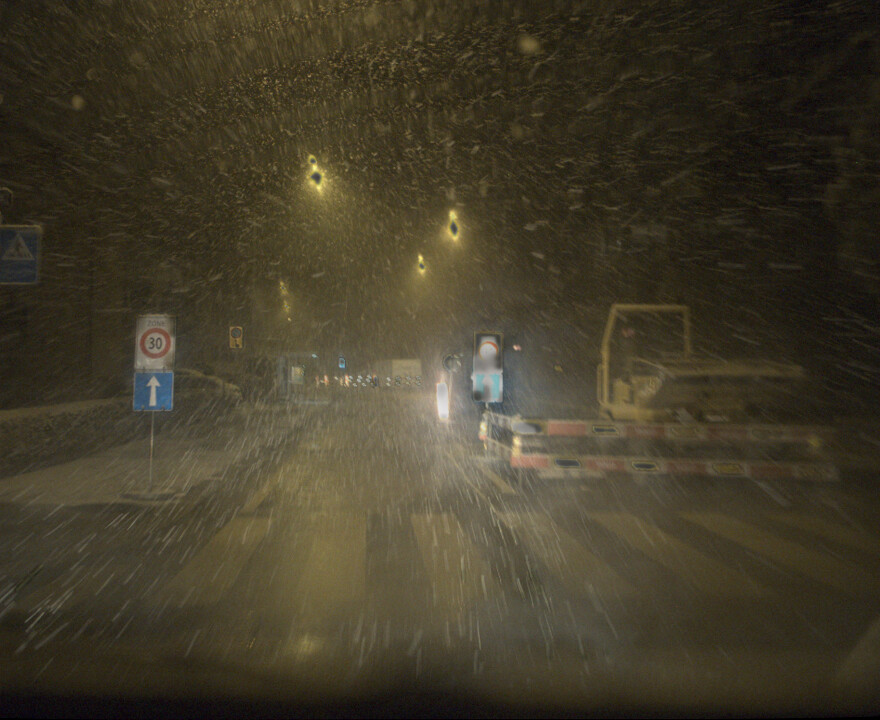}\\
        (a) Image & (b) Ours - Image \\
        \includegraphics[width=0.45\linewidth]{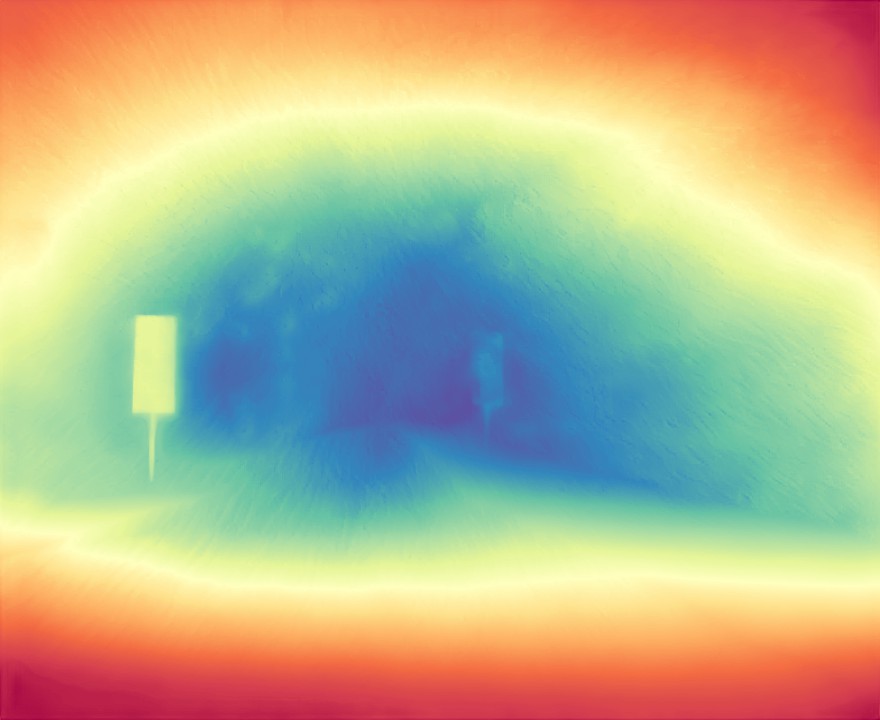} &
        \includegraphics[width=0.45\linewidth]{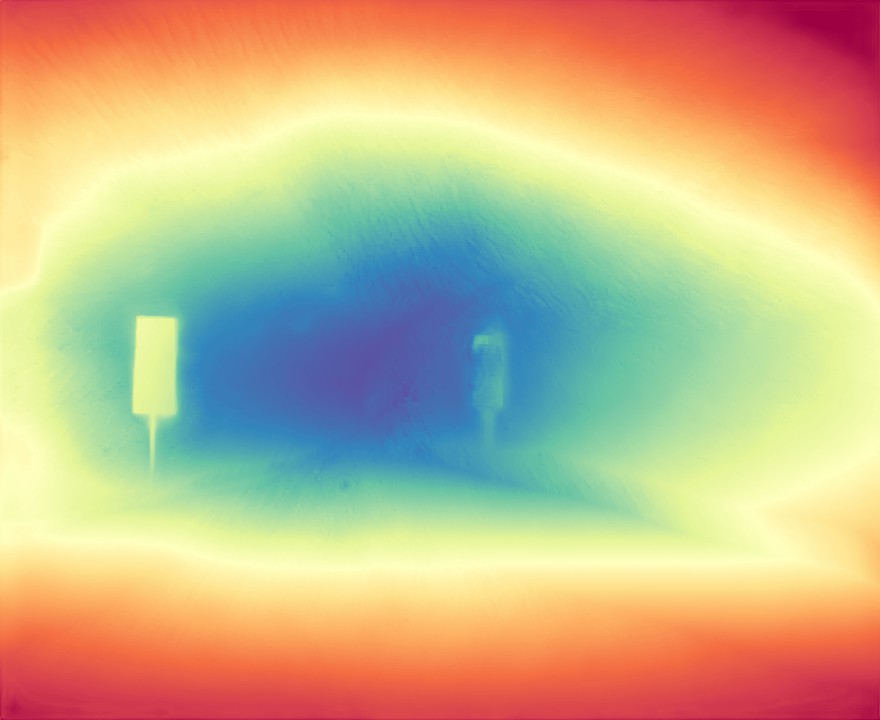} \\
        (c) Image-Depth & (d) Ours - Depth \\
    \end{tabular}
    \caption{\revtwo{\textbf{Qualitative results of our method on downstream task - depth estimation.}
    We show the input image and depth map from the DSEC-Snow dataset, along with the depth map generated by our method.
    The depth map is estimated using the input image and events, demonstrating the effectiveness of our approach in leveraging both modalities for improved depth estimation in snowy conditions.}}
    \label{fig:supp:downstream}
\end{figure*}

In addition to the qualitative results, we also evaluate our method on a downstream task, depth estimation, using the real driving dataset in \Fig \ref{fig:supp:downstream}.

%% file: floats/fig_muses.tex
\begin{figure*}[ht]
    \centering
    \newcommand{\thisfigWidth}{0.45\linewidth}
    \begin{tabular}{M{\thisfigWidth}M{\thisfigWidth}}
        \includegraphics[trim={12cm, 10cm, 35cm, 10cm},clip,width=\linewidth]{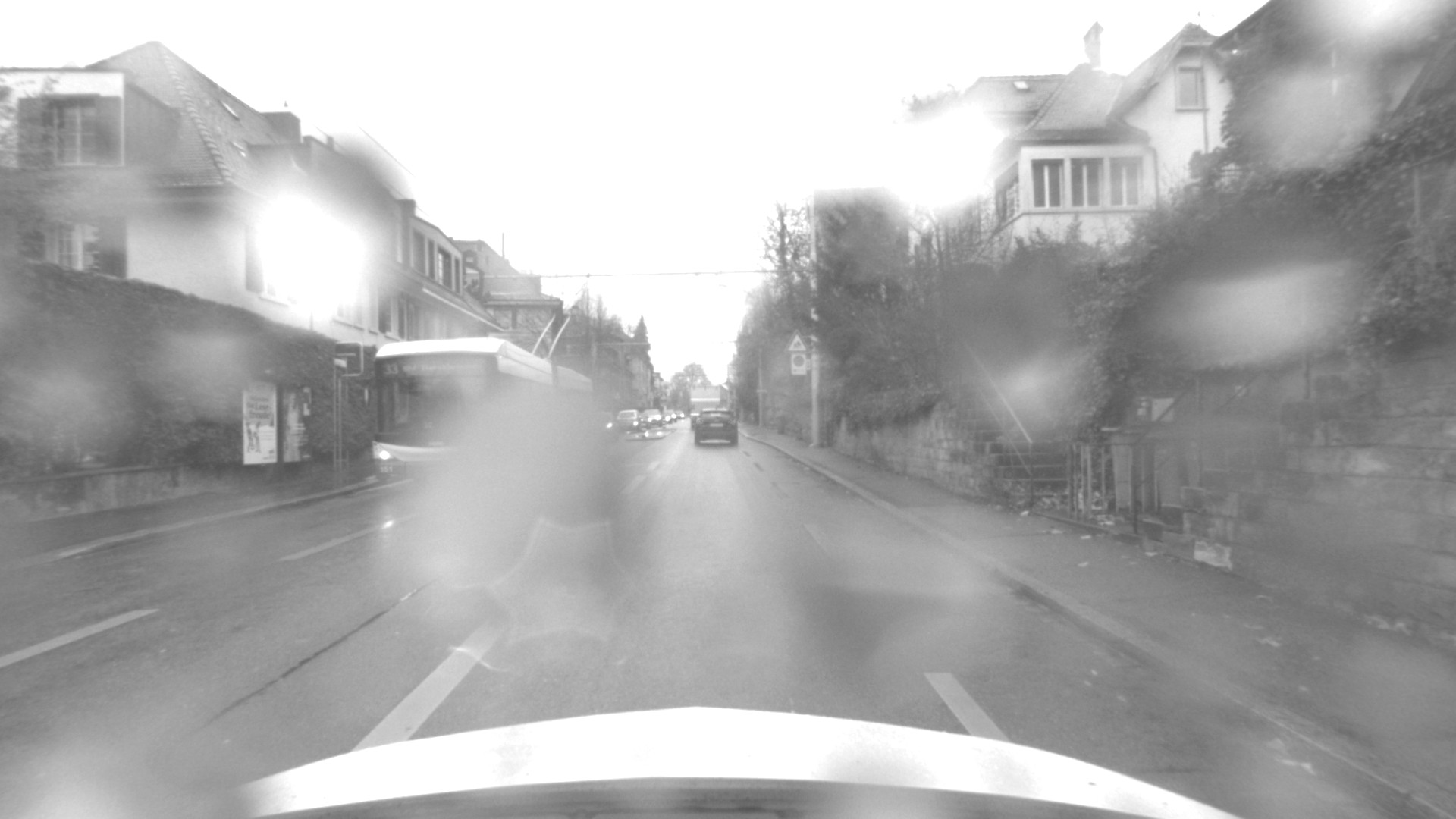}&
        \includegraphics[trim={12cm, 10cm, 35cm, 10cm},clip,width=\linewidth]{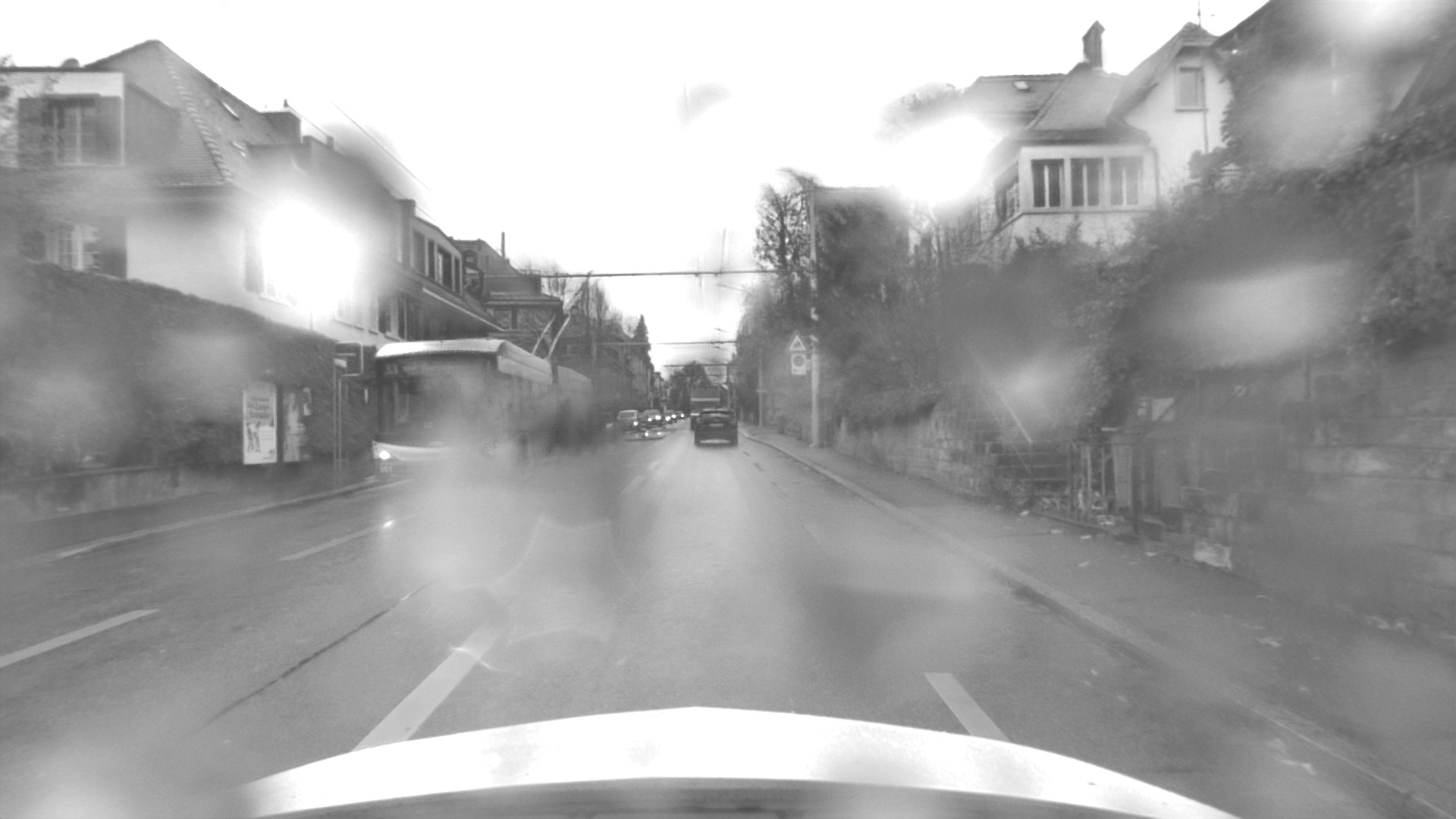}\\
        \includegraphics[trim={25cm, 15cm, 15cm, 5cm},clip,width=\linewidth]{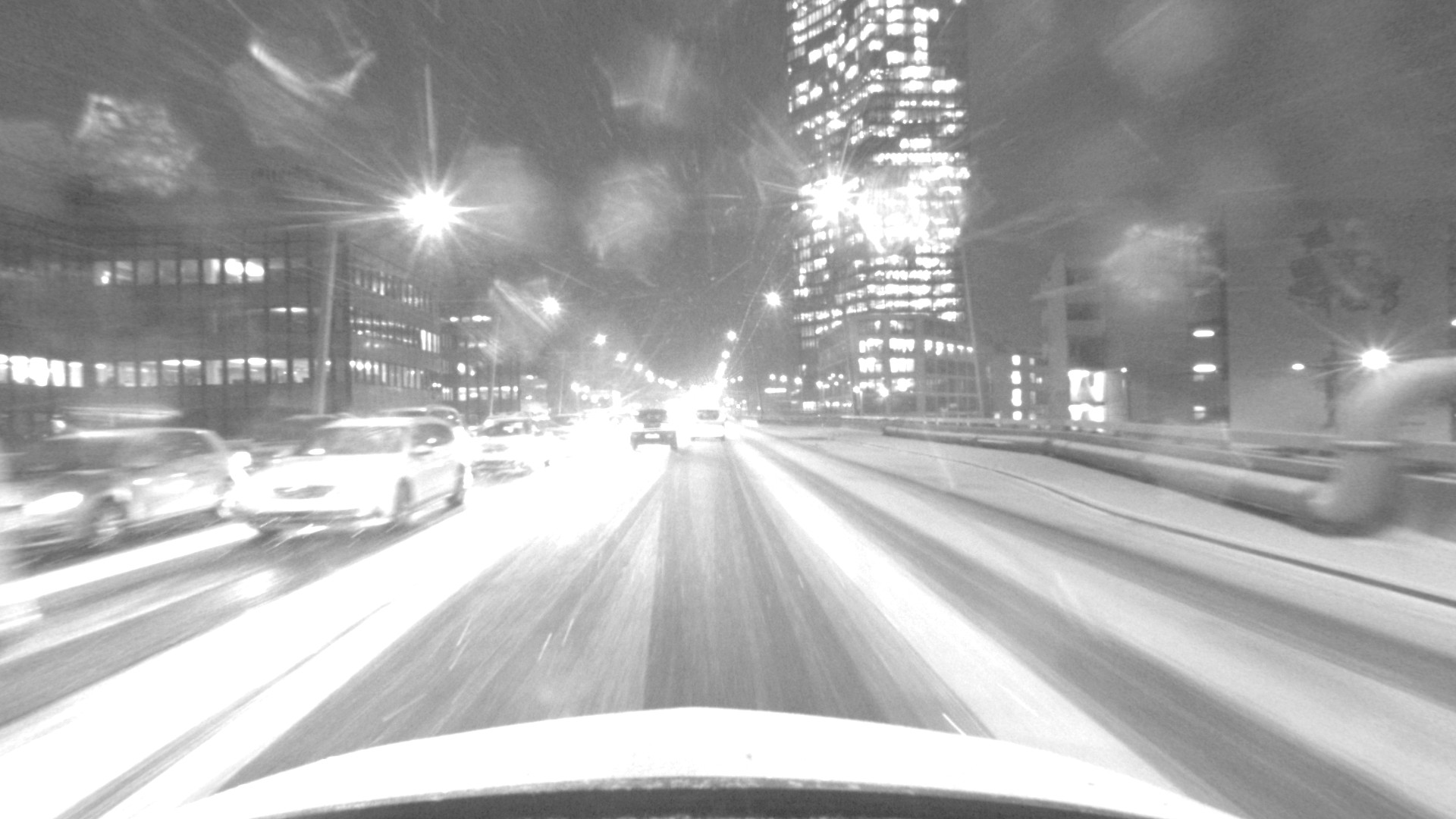}&
        \includegraphics[trim={25cm, 15cm, 15cm, 5cm},clip,width=\linewidth]{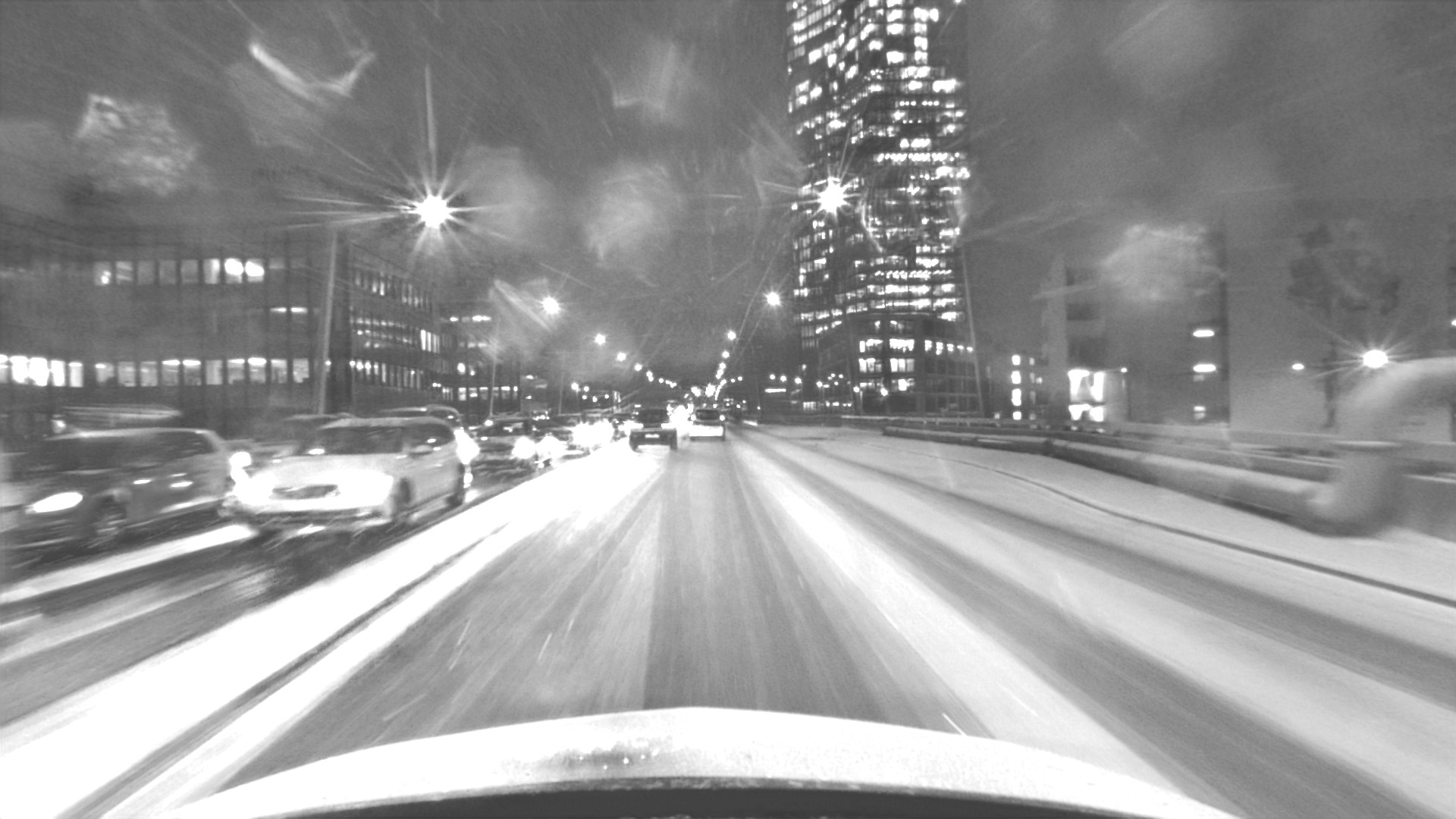}\\
        (a) Image & (b) Ours\\
    \end{tabular}
    \caption{\textbf{Qualitative comparison on the MUSES Dataset\cite{brodermann24eccv} with rain and snow occlusions.}
    (a) Input images affected by rain or snow occlusions, and (b) results from our method. 
    The examples show the effectiveness of our approach in handling adverse weather occlusions across different datasets.}
    \label{fig:results_muses}
\end{figure*}

%% file: floats/fig_dsec_downstream.tex
\begin{figure*}[!t]
    \centering
    \newcommand{\thisfigWidth}{0.18\linewidth}

    \begin{tabular}{
	M{\thisfigWidth}M{\thisfigWidth}M{\thisfigWidth}M{\thisfigWidth}M{\thisfigWidth}}
    \includegraphics[width=\linewidth]{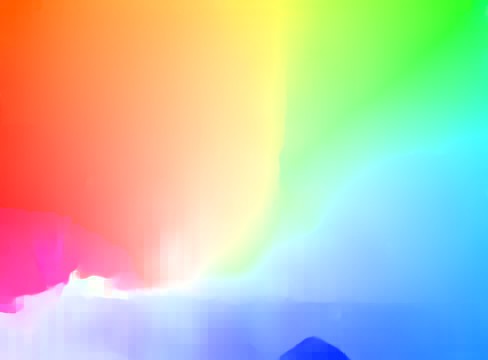}&
    \includegraphics[width=\linewidth]{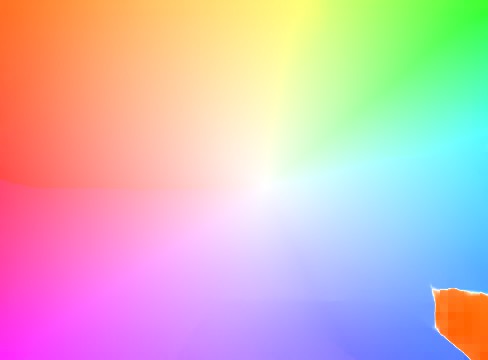}&
    \includegraphics[width=\linewidth]{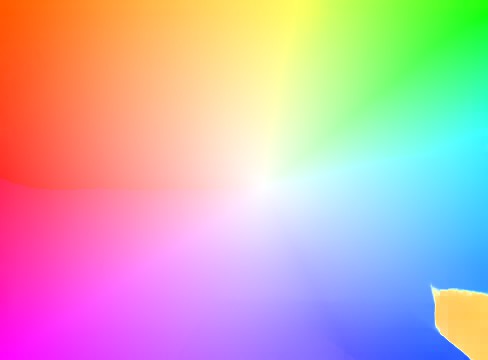}&
    \includegraphics[width=\linewidth]{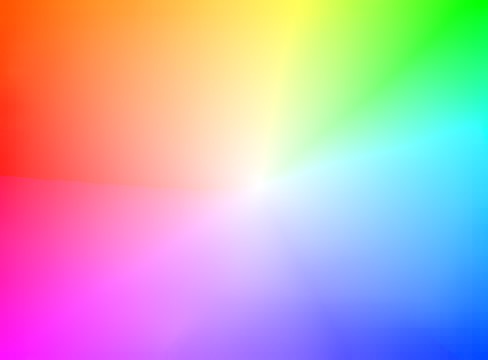}&
    \includegraphics[width=\linewidth]{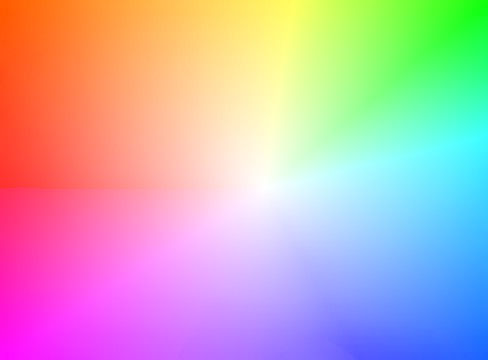}\\
            
    (a) E2VID \cite{Rebecq19pami} & (b) S2VD \cite{Yue21cvpr} & (c) Restormer \cite{Zamir21cvpr} & (d) Ours & (e) Groundtruth\\
    \end{tabular}
     \caption{\textbf{Comparing the image reconstruction, optical flow estimation} for event-only baseline E2VID, video baseline S2VD, and image baseline Restormer with our method on the \DAVISSnow dataset.}
    \label{fig:result_realsnow_downstream}
\end{figure*}

%% file: floats/tables/tab_dsec_downstream.tex
\begin{table}[ht]
\centering
\begin{adjustbox}{max width=\linewidth}
    \begin{tabular}{cc|cccc}
    \toprule
    Method & Input & \multicolumn{4}{c}{Optical Flow} \\
    
       &   & EPE $\downarrow$ & $ AE<1\uparrow$ & $AE<3 \uparrow$ & $AE<5 \uparrow$  \\
    \midrule
    Restormer \cite{Zamir21cvpr}  & I & \revtwo{19.36} & \revtwo{0.06} & \revtwo{0.36} & \revtwo{0.48}\\
    SnowFormer \cite{Chen22arxiv} & I & \revtwo{30.80} & \revtwo{0.02} & \revtwo{0.20} & \revtwo{0.30}\\
    RLP \cite{Zhang23ICCV}        & I & \revtwo{17.64} & \revtwo{0.06} & \revtwo{0.36} & \revtwo{0.47}\\
    S2VD \cite{Chen22arxiv}       & V & \revtwo{29.64} & \revtwo{0.03} & \revtwo{0.24} & \revtwo{0.35}\\
    E2VID \cite{Rebecq19pami}     & E & \revtwo{42.92} & \revtwo{0.00} & \revtwo{0.04} & \revtwo{0.08}\\ \hline

    Ours (Model-based)          & E+I & \revtwo{39.58} & \revtwo{0.00} & \revtwo{0.11} & \revtwo{0.19}\\
    Ours                        & E+I & \revtwo{\textbf{10.42}} & \revtwo{\textbf{0.11}} & \revtwo{\textbf{0.56}} & \revtwo{\textbf{0.67}}\\
    \bottomrule
    \end{tabular}
\end{adjustbox}
\caption{\revtwo{\textbf{Quantitative evaluation of downstream task (optical flow)}-on desnowed images using different restoration methods and input modalities. 
Our event-image fusion approach consistently achieves the best performance, as measured by EPE and accuracy for optical flow.}}
\label{tab:results:downstreamdepth}
\end{table}